\documentclass[11pt]{article}
\usepackage[margin=1in]{geometry}

\usepackage{amsmath,amsthm}
\usepackage[ruled,vlined]{algorithm2e}
\usepackage{xr-hyper}
\usepackage[affil-it]{authblk}
\usepackage{graphicx,subfigure,float}
\usepackage{amssymb,amsmath}
\usepackage{comment}
\usepackage{natbib}
\usepackage{setspace}
\newtheorem{prop}{Proposition}
\newtheorem{theorem}{Theorem}
\newtheorem{lem}{Lemma}
\newtheorem{remark}{Remark}
\newtheorem{cor}{Corollary}

\newtheorem{definition}{Definition}

\newcommand{\bE}{\mathbb{E}} 
\newcommand{\bR}{\mathbb{R}} 
\newcommand{\bC}{\mathbb{C}} 
\newcommand{\bP}{\mathbb{P}}
\newcommand{\bX}{\boldsymbol{X}} 
\newcommand{\bY}{\boldsymbol{Y}} 
\newcommand{\cP}{\mathcal{P}}
\newcommand{\cX}{\mathcal{X}} 
\newcommand{\cY}{\mathcal{Y}}
\newcommand{\cF}{\mathcal{F}} 
\newcommand{\cH}{\mathcal{H}} 
\newcommand{\cN}{\mathcal{N}} 
\newcommand{\err}{\mathrm{Error}}
\newcommand{\LOO}{\mathrm{LOO}}
\newcommand{\LOCO}{\mathrm{LOCO}}
\newcommand{\ind}{\mathbb{I}}
\newcommand{\del}{\backslash}
\newcommand{\stb}{\mathrm{stb}}
\newcommand{\RNum}[1]{\mathrm{\uppercase\expandafter{\romannumeral #1\relax}}}
\usepackage{enumitem} %FOR ALGORITHM SPACE
\usepackage{amsmath, amssymb}
\usepackage{xcolor}[name]

\usepackage[utf8]{inputenc} % allow utf-8 input
\usepackage[T1]{fontenc}  % use 8-bit T1 fonts
\usepackage{hyperref}    % hyperlinks
\usepackage{url}      % simple URL typesetting
\usepackage{booktabs}    % professional-quality tables
\usepackage{amsfonts}    % blackboard math symbols
\usepackage{nicefrac}    % compact symbols for 1/2, etc.
\usepackage{microtype}   % microtypography
\usepackage{xcolor}     % colors
\usepackage[symbol]{footmisc}

\newtheorem{assumption}{Assumption}

\providecommand{\keywords}[1]
{ 
	\small	
	\textbf{\textit{Keywords---}} #1
}
\DeclareMathOperator*{\argmin}{arg\,min}

\newcommand{\bbR}{\mathbb{R}}
\newcommand{\bbP}{\mathbb{P}}

\begin{document}

%%%%%%%%%%%%%%%%%%%%%%%%%%%%%%%%%%%%%%%%%%%%%%%%%%%%%%%%%%%%%%%%%%%%%%%%%%%%%%

\title{\bf LOCO Feature Importance Inference without Data Splitting via Minipatch Ensembles}
\author[1]{Luqin Gan$^\dagger$}
\author[2]{Lili Zheng$^\dagger$\thanks{Corresponding author: llzheng@illinois.edu}}
    \author[3]{Genevera I. Allen}

\affil[1]{Department of Statistics, Rice University}
\affil[2]{Department of Statistics, University of Illinois Urbana-Champaign}
\affil[3]{Department of Statistics, Columbia University}
  \date{}
  \maketitle
\bigskip

\begin{abstract}
Feature importance inference is critical for the interpretability and reliability of machine learning models. There has been increasing interest in developing model-agnostic approaches to interpret any predictive model, often in the form of feature occlusion or leave-one-covariate-out (LOCO) inference. Existing methods typically make limiting distributional assumptions, modeling assumptions, and require data splitting. In this work, we develop a novel, mostly model-agnostic, and distribution-free inference framework for feature importance in regression or classification tasks that does not require data splitting. Our approach leverages a form of random observation and feature subsampling called minipatch ensembles; it utilizes the trained ensembles for inference and requires no model-refitting or held-out test data after training. We show that our approach enjoys both computational and statistical efficiency as well as circumvents interpretational challenges with data splitting. Further, despite using the same data for training and inference, we show the asymptotic validity of our confidence intervals under mild assumptions. Additionally, we propose theory-supported solutions to critical practical issues including vanishing variance for null features and inference after data-driven tuning for hyperparameters. We demonstrate the advantages of our approach over existing methods on a series of synthetic and real data examples.
\end{abstract}        
\footnotetext{\hspace{-0.7em}$^\dagger$Equal co-authorship: order determined randomly.}
 \newpage

\section{Introduction}\label{sec:intro}
Reliability and interpretability are crucial ingredients to building trustworthy artificial intelligence (AI) and machine learning (ML) systems for deployment in high-stakes applications like autonomous vehicles, healthcare, criminal justice, and national security.  Feature importance, defined as how an input feature influences the output predictions of a ML model, is one of the most popular forms of ML interpretation. Consider, for example, utilizing ML models in banking to make automated decisions on mortgage applications.  It is important to understand which features significantly influence the predictions as this can help bankers better understand, develop, and deploy the model, help regulators assess the model for any violations of the Fair Housing Act, and help society gain trust in the model's decisions.  Furthermore, to ensure that the model diagnostics and deployment decisions are based on reliable feature importance scores, it is critical to quantify their associated uncertainty. Recently, there has been a surge of interest in machine learning uncertainty quantification; most existing work has focused on quantifying uncertainty in predictions (e.g. conformal inference) \citep{angelopoulos2023conformal} with much fewer studies focused on feature importance inference.

Various feature importance metrics have been considered in the literature, which we argue fall into two major categories. The first type, which we call {\em population feature importance}, characterizes properties of the underlying population model $f$ of the data, {\em e.g.,} the linear model coefficient and metrics of conditional dependency strength \citep{zhang2020floodgate}.  The second type, which we term as {\em machine learning feature importance}, focuses on the properties of the trained ML model $\hat{f}$ we have at hand, such as Layerwise Relevance Propagation \citep{bach2015pixel} in deep learning or random forest variable importance based on impurity or permutation \citep{breiman2001random}. 
These two types of feature importance metrics are also discussed in \cite{williamson2020efficient} and referred to as ``population vs. algorithmic variable importance". In the aforementioned example of an AI mortgage approval system, we primarily care about ML feature importance, as it explains features' relevance for the trained black box model $\hat{f}$ that will be deployed to make mortgage decisions.

Despite the critical role of ML feature importance in promoting model transparency and accountability, its uncertainty quantification is still an under-studied topic. The dominating statistics literature focuses on population feature importance, such as the post-selection inference \citep{berk2013valid,lee2016exact} and the de-biased Lasso \citep{zhang2014confidence,van2014asymptotically} for high-dimensional (generalized) linear models, and more recently, model-agnostic inference methods for population feature importance notions such as the floodgate \citep{zhang2020floodgate,wang2023total}, GCM and its extensions \citep{shah2020hardness,scheidegger2022weighted}, PCM \citep{lundborg2024projected}, VIMP \citep{williamson2021nonparametric,williamson2021general}, targeted-learning-based approach \citep{wang2024targeted}, Shapley-value-based feature importance inference \citep{williamson2020efficient}, as well as inference for importance metrics that are free from correlation distortions \citep{du2025disentangled,verdinelli2024decorrelated}. These population inference methods can often shed light on the underlying data-generating mechanism, but they can be less relevant when the primary interest is interpreting the current trained ML model. Further, \cite{shah2020hardness} have shown that population feature importance testing is fundamentally challenging as it is powerless without making limiting assumptions on the data-generating model or the consistency of the trained model; all the above mentioned approaches make such limiting assumptions. On the other hand, very few prior methods have been proposed to conduct ML feature importance inference. The conditional predictive impact (CPI) method \citep{watson2021testing} tests whether the trained model is more predictive with the original features than the knockoff counterpart, but its validity requires knowing the underlying data distribution. The model class reliance (MCR) approach \citep{fisher2019all} considers the feature importance amongst a collection of models with good predictions, instead of any single trained ML model.

To the best of our knowledge, the most relevant, general, and assumption-light approach for ML feature importance inference is the leave-one-covariate-out (LOCO) method \citep{lei2018distribution,rinaldo2019bootstrapping}, which targets the change in prediction error of any ML model after feature occlusion\footnote{Occlusion-based feature importance metrics are popular forms of ML feature importance with wide applicability and intuitive interpretation; we refer the readers to \cite{covert2021explaining} for an expansive discussion on this class of feature importance metrics.}. However, the LOCO method involves model refitting for each tested feature, costing intensive and extra computations for complex models and large-scale data sets. Perhaps more critically, it also requires data splitting, utilizing a training set and a held-out calibration or test set for inference. This means that the inference is only valid for the model built using the training set but not all available data, a major limitation for data-hungry ML models. Further, both the predictive model and inferential decision could change if a different data split was used. This is especially undesirable in the context of feature importance inference, whose goal is to aid interpretability and reliability. Additionally, data splitting is known to sacrifice both the accuracy of ML models and inference efficiency \citep{lei2018distribution}.

In this paper, we aim to develop a novel inference procedure for LOCO feature importance that inherits the advantages of the existing LOCO method: {\em model-agnostic, distribution-free, and assumption-lean}, while also addressing the aforementioned challenges by boosting {\em statistical and computational efficiency without utilizing data splitting}. Note that we use "model-agnostic" to mean that any ML predictive model (regression or classification) can be employed within our framework; that is, our inference approach can be employed for any $\hat{f}$. This is different from prior works on population feature importance inference, however, where a ``model-agnostic" method gives inference for a fixed population quantity regardless of the ML model utilized.

\paragraph{Contribution and Organization} Inspired by the Jacknife+-after-bootstrap approach \citep{kim2020predictive} in conformal inference, which uses observation subsampling to obtain fast leave-one-out predictions, we propose to leverage an ensemble learning framework which randomly subsamples both observations and features in tabular data, termed "minipatch ensembles", introduced by \cite{toghani2021mp,yao2020feature}. In Section \ref{sec:methods}, we introduce our inferential approach which utilizes the double-subsampling structure to avoid both model-refitting and data-splitting for LOCO inference, hence giving almost computationally-free inference for the current model trained on all available data.   
Despite the dependencies of our leave-one-out estimates created by avoiding data splitting, in Section \ref{sec:coverage_guarantee}, we still provide theoretical guarantees showing that our confidence intervals enjoy \emph{asymptotically correct coverage} for our feature importance score, under only minimal assumptions for any ML model and regression or classification task. We do so by proving an appealing stability property of minipatch ensembles. Exploiting the stability further, we address practical issues in feature importance inference, including vanishing variance of null features, and inference after hyperparameter tuning using the same data. Notably, to conduct valid inference with dependency, the idea of leveraging stability of random ensemble methods instead of having to split data, could be of independent interest in the ML post selection inference and uncertainty quantification literature. Finally, in Section \ref{sec:result}, we demonstrate the benefits of our approach via synthetic and real data experiments.

\section{LOCO Inference via Minipatch Ensembles}\label{sec:methods}
In this section, we propose a new inference procedure for LOCO feature importance scores associated with any predictive model that takes the form of minipatch ensembles. We will show that leveraging the structure of minipatch ensembles can bring a number of benefits in feature importance inference, both statistically and computationally. 

\paragraph{Background: LOCO feature importance.} The leave-one-covariate-out (LOCO) feature importance score was proposed and studied in \cite{lei2018distribution}. It captures how feature $j$ affects the prediction error in the trained model as follows: 
	\begin{align}\label{eq:loco_inference_target}
	\Delta_{j}(\boldsymbol{X},\boldsymbol{Y})=\mathbb{E} \left[ \mathrm{Error}\left(Y, \mu_{\backslash j}(X_{\backslash j};\boldsymbol{X}_{:,\backslash j},\boldsymbol{Y}) \right) - \mathrm{Error}\left(Y, \mu(X;\boldsymbol{X},\boldsymbol{Y}) \right) |\boldsymbol{X},\boldsymbol{Y}\right],
	\end{align}
	where $\mu(\cdot;\boldsymbol{X},\boldsymbol{Y})$ is the full predictive model fit on the training data set $(\boldsymbol{X},\boldsymbol{Y})$, using all features; $\mu_{\backslash j}(\cdot;\boldsymbol{X}_{:,\backslash j},\boldsymbol{Y})$ is the reduced predictive model trained on the same data set but without feature $j$; $\mathrm{Error}(\cdot)$ is some nonconformity function appropriate for the supervised learning task (e.g., absolute error, hinge loss, etc.), and the expectation is taken over a new test data point $(X, Y)$ sampled from the same distribution as the training data. One can understand this inference target $\Delta_{j}(\boldsymbol{X},\boldsymbol{Y})$ as the additional predictive power provided by feature $j$ given all other features, when the current predictive model $\mu(\cdot;\,\bX,\,\bY)$ fitted using the training data is applied on unseen test data. A large absolute value of $\Delta_{j}(\boldsymbol{X},\boldsymbol{Y})$ indicates that feature $j$ significantly affects the trained model's prediction, in a helpful (harmful) way if $\Delta_{j}(\boldsymbol{X},\boldsymbol{Y})>0 (<0)$. One may argue that $\Delta_{j}(\boldsymbol{X},\boldsymbol{Y})$ is not just a property of the original model $\mu(\cdot)$, as it also involves a reduced model $\mu_{\backslash j}(\cdot)$. In fact, model-agnostic feature importance metrics often require the construction of a comparison baseline, either through feature occlusion or permutation, or by creating a non-existent instance with designed feature values \cite[see][for a detailed discussion]{covert2021explaining}. Occlusion-based feature importance has a natural interpretation, and as we will see later, it also facilitates a connection between $\Delta_{j}(\boldsymbol{X},\boldsymbol{Y})$ and prior population feature importance notions. 
    
	\paragraph{Background: LOCO-Split.} To perform statistical inference for \eqref{eq:loco_inference_target}, \cite{lei2018distribution} propose to construct confidence interval for feature $j$ in regression problems via data splitting, which we refer to as ``LOCO-Split'' in order to distinguish it from our method. Specifically, the $N$ training samples are split into two sets: $D_1$, $D_2$. They then fit a full model $\mu(\cdot;\boldsymbol{X}_{D_1,:},\boldsymbol{Y}_{D_1})$ and a leave-j-covariate-out model $\mu_{\backslash j}(\cdot;\boldsymbol{X}_{D_1,\backslash j},\boldsymbol{Y}_{D_1})$ separately on the first part $D_1$ of the training data. Then they calculate the change of nonconformity on $D_2$ after removing feature $j$:
	\begin{equation}\label{eq:Delta}
	\hat{\Delta}_j(X_{i}, Y_{i}) = \mathrm{Error}(Y_{i},\mu_{\backslash j}(X_{i,\backslash j};\boldsymbol{X}_{D_1,\backslash j},\boldsymbol{Y}_{D_1})) - \mathrm{Error}(Y_{i},\mu(X_{i};\boldsymbol{X}_{D_1,:},\boldsymbol{Y}_{D_1})), i\in D_2,
	\end{equation}
	and construct confidence intervals using $D_2$ via an asymptotic Z-test or non-parametric sign test, which are valid under mild assumptions \citep{rinaldo2019bootstrapping}. However, due to data splitting, the target feature importance of this inference procedure is the property of $\mu(\cdot;\,\bX_{D_1,:},\,\bY_{D_1})$, the model trained using partial data $D_1$ instead of the full data at hand; it also suffers from a trade-off of training accuracy and inference power: a larger $|D_1|$ is desired for a better training model for deployment in the future, while this inevitably sacrifices the inference power. Furthermore, to perform inference for each feature $j$, LOCO-split requires \emph{intensive extra computation} due to re-fitting a model $\mu_{\backslash j}(\cdot;\bX_{D_1,\backslash j}, \bY_{D_1})$.
 
    \paragraph{Background: minipatch learning.}
	Before presenting our idea for tackling the challenges of LOCO inference, let us also introduce a recent ensemble method for machine learning prediction: "minipatch learning" \citep{toghani2021mp,yao2021minipatch}. Here, small subsets of both observations and features are randomly selected and used for model training, referred to as minipatches. Each minipatch is of a fixed size $(m,\,n)$, with $m<M$, $n<N$ being the number of subsampled features and observations, respectively. Given any machine learning algorithm, e.g., ridge regression, decision tree, neural networks, one can train a base model on each minipatch. The full predictor is the ensemble (average) of all the models trained on these random minipatches. As an ensemble method, minipatch learning is flexible and can be applied with any machine learning algorithm being the base learner on minipatches; it is also computationally efficient since model fitting on each tiny minipatch can be extremely fast, memory-efficient, and embarrassingly parallelizable. Prior works \citep{yao2021minipatch, lejeune2020implicit} also suggest that minipatch ensembles enjoy implicit ridge regularization properties, both empirically and theoretically. These properties make minipatch learning especially attractive for machine learning predictions in large-scale, correlated, and noisy settings. Interestingly, beyond prediction, the double subsampling of both observations and features also gives it a unique advantage for LOCO inference.
\subsection{Fast LOCO Inference for Minipatch Ensembles}\label{sec:LOCO-MP}
  Inspired by the LOCO-Split \cite{lei2018distribution} and the Jackknife+-after-bootstrap algorithm \cite{kim2020predictive}, we propose the LOCO-MP algorithm (Algorithm \ref{algo:loco}) to conduct statistical inference for LOCO feature importance associated with minipatch predictors. In particular, suppose that one has trained a minipatch ensemble of $K$ members for a machine learning prediction task; the full predictive model is
$\mu(\cdot;\boldsymbol{X},\boldsymbol{Y}) = \frac{1}{K}\sum_{k=1}^K\mu_k(\cdot)$,
 with $\mu_k(\cdot)$ defined in the step 1 of Algorithm \ref{algo:loco}. One then wants to perform LOCO inference on this trained model for diagnostic or auditing reasons, or formally, to {\em construct a confidence interval for $\Delta_j$ in \eqref{eq:loco_inference_target}, with $\mu_{\backslash j}(X_{\backslash j};\boldsymbol{X}_{:,\backslash j},\boldsymbol{Y})$ being a hypothetical reduced model, if one applies the same minipatch learning algorithm on $(\boldsymbol{X}_{:,\backslash j},\,\boldsymbol{Y})$. }
 
 Instead of splitting the data and refitting the model, here we utilize the unique double-subsampling scheme of minipatch learning to construct the test statistic. Specifically, we can approximate the effect of feature occlusion by only averaging over minipatches without feature $j$; the non-conformity scores can be computed in a leave-one-out fashion, as leaving one observation out is as simple as leaving one feature out.  
 As summarized in the step 2 and 3 of Algorithm \ref{algo:loco}, we compute the leave-one-observation-out and leave-one-covariate-out predictions for each observation $i=1,\dots,N$ and feature of interest $j$, and calculate the prediction error change due to feature occlusion for each sample in the training data set ($\{\hat{\Delta}_j(X_i,Y_i)\}_{i=1}^N$). Each $\hat{\Delta}_j(X_i,Y_i)$ illustrates the importance of feature $j$ when predicting sample $i$, if the predictive model is fitted without $i$. We then construct our confidence interval centered around their mean $\bar{\Delta}_j$, with the width calculated based on their sample variance. Note here that after the minipatch learning step, the computations for feature importance inference comes nearly for free since the remaining steps are simple averaging. Furthermore, all samples are used both for training and testing with no data-splitting, and hence we provide inference for the model trained using all the data instead of a sub-model that can change with different random splits. This added benefit, however, comes at a potential cost: now there exist strong dependencies amongst the different $\hat{\Delta}_j(X_i,Y_i)$'s, making theoretical analysis of our procedure challenging. We will revisit this and show how we address this challenge in Section \ref{sec:coverage_guarantee}.

 \paragraph{Distribution-free Predictive Inference} Aside from the confidence intervals for feature importance, our same procedure can also be leveraged to construct predictive intervals. Inspired by \citep{kim2020predictive} which provides fast and distribution-free predictive inference using Jackknife+ \citep{barber2021predictive} with bootstrap, we propose a Jackknife+ Minipatch conformal inference procedure (J+MP) that can additionally take advantage of our fitted leave-one-out (LOO) predictors to construct predictive confidence intervals, which also comes for free and can be obtained simultaneously as the feature importance interval. As far as we know, this is the first inference procedure that can perform feature importance inference and predictive inference at the same time. More details on the J+MP method are included in the Supplementary material.
	
	\begin{algorithm}[ht!]
    \singlespacing\noindent{\textbf{Input}}: Training data ($\boldsymbol{X}\in \bbR^{N\times M},\boldsymbol{Y}\in \bbR^N$), feature of interest $j$, minipatch sizes $n$, $m$; number of minipatches $K$; base learner $H$; confidence level $1-\alpha$.
		\begin{enumerate}
			\item Perform Minipatch Learning: For $k=1,...,K$
			\begin{enumerate}
				\item Randomly subsample $n$ observations, $I_k \subset [N]$, and $m$ features, $F_k \subset [M]$.
				\item Train prediction model: for any $X\in \bR^M$, $\mu_k(X) = H(\boldsymbol{X}_{I_k,F_k},\boldsymbol{Y}_{I_k})(X_{F_k})$.
			\end{enumerate}
			\item Obtain LOO and LOO + LOCO predictions:
			\begin{enumerate}
				\item Obtain the ensembled LOO prediction: $\mu_{-i}(X_i) = \frac{1}{\sum_{k=1}^{K} \mathbb{I}(i \notin I_k)} \sum_{k=1}^{K} \mathbb{I}(i \notin I_{k}) \mu_{k} (X_{i})$;
				
				\item Obtain the ensembled LOO + LOCO prediction: $\mu_{-i}^{-j}(X_i) = \frac{1}{\sum_{k=1}^{K} \mathbb{I}(i \notin I_k) \mathbb{I}(j \notin F_k)} \sum_{k=1}^K \mathbb{I}(i\notin I_k)\mathbb{I}(j\notin F_k) \mu_{k}(X_i)$;
			\end{enumerate}  
			\item Calculate LOO Feature Occlusion:
			$\hat{\Delta}_j(X_{i}, Y_{i}) = \mathrm{Error}(Y_{i},\mu_{-i}^{-j}(X_i)) - \mathrm{Error}(Y_{i},\mu_{-i}(X_i) )$; 
			\item Obtain a $1-\alpha$ interval for $\Delta_j$: $\hat{\mathbb{C}}_j=
			[\bar{\Delta}_j - \frac{z_{\alpha/2}\hat{\sigma}_j}{\sqrt{N}},\bar{\Delta}_j +\frac{z_{\alpha/2}\hat{\sigma}_j}{\sqrt{N}}]$, with $\bar{\Delta}_j=\frac{1}{N}\sum_{i=1}^N\hat{\Delta}_j(X_{i}, Y_{i})$ being the sample mean and $\hat{\sigma}_j= \sqrt{\frac{\sum_{i=1}^N(\hat{\Delta}_j(X_{i}, Y_{i})-\bar{\Delta}_j)^2}{N-1}}$ being the sample standard deviation. 
		\end{enumerate}
		\textbf{Output}: $ \hat{\mathbb{C}}_j$
		\caption{Minipatch LOCO Inference}
		\label{algo:loco}
	\end{algorithm}
 \subsection{On the Interpretation of LOCO-MP Target}\label{sec:target_discussion}
Our inference target $\Delta_j(\bX,\bY)$ in \eqref{eq:target_inference} characterizes the predictive accuracy difference between $\mu(\cdot;\bX,\bY)=\frac{1}{K}\sum_{k=1}^K\mu_k(\cdot)$ and $\mu_{\backslash j}(\cdot;\bX_{:,\backslash j},\bY)$. $\mu$ is the minipatch ensemble predictor obtained from the first step of Algorithm~\ref{algo:loco}, while $\mu_{\backslash j}$ is a hypothetical minipatch ensemble predictor with the same training process and training data, but without access to feature $j$. Both predictive models share the same pre-specified base model trained on each minipatch, e.g., decision trees, and the same minipatch size and number. 
Therefore, the target $\Delta_j(\bX,\bY)$ not only shows how much the current predictor $\mu(\cdot;\bX,\bY)$ uses feature $j$, but also whether it helps or hurts prediction within the same class of minipatch ensembles. This interpretation is similar to that of the target of LOCO-split, while some important distinctions also exist. First, our $\Delta_j$ is associated with the model trained with the full data $(\bX,\,\bY)$, while the inference target of LOCO-split is for the model trained with only a subset of the data, which may have degraded predictive performance and hence be less satisfying in model deployment. Second, since we operate within the minipatch framework, our inference procedure is almost but not entirely model-agnostic as LOCO-split; our inference is for the feature importance of any minipatch ensemble predictor, with any desired base model applied on each minipatch. 
Despite these differences, both the inference targets of ours and LOCO-split are close to a form of ML feature importance, as they care primarily about the trained ML model, although it was compared to a reduced model within the same model class. Interestingly, by introducing the reduced model into the target, our $\Delta_j$ can also be related to the population feature importance if willing to make some assumptions.

We first note that a recently proposed population feature importance score by \cite{williamson2021general} shares a similar occlusion-based form with \eqref{eq:target_inference}, while instead of evaluating the prediction performance of trained predictors ($\mu(\cdot;\bX,\bY)$ and $\mu_{\backslash j}(\cdot;\bX_{:,\backslash j}, \bY)$), they study the population risk minimizers. This connection hinted that when the trained predictors are close enough to their population counterpart, e.g., consistency assumptions often made in prior works, our ML feature importance score \eqref{eq:target_inference} may well approximate its population version. Here, we show that even under mild regularity assumptions without consistency, there is a connection between our target and a population feature importance score under the linear model, as an illustrative example. The informal statement is as follows, with the detailed version included in the Supplementary material.
 
Suppose that all data points $(X_i,y_i)$ are i.i.d. samples of a linear model: $y_i = X_i^\top\beta^*+\epsilon_i$, where $\beta^*\in \mathbb{R}^M$ is the linear regression parameter, and $\{\epsilon_i\}_{i=1}^N$ are independent sub-Gaussian noise of mean zero. 
Also, assume that the least squares estimator is our base learner for each minipatch, and the squared error $\err(Y,\hat{Y})=(Y-\hat{Y})^2$ is in use. For now, we focus on the setting with independent features: $X_i\sim\mathcal{N}(0,I_p)$, but we will revisit this linear model later with correlated features in Section \ref{sec:dependent_features}. 
\begin{theorem}[Informal]\label{prop:MP_target}  Under the described linear model setup, we have $\Delta_j = \Delta_j^* + \varepsilon$ with probability at least $1-N^{-c}$, where $\varepsilon$ is a vanishing approximation error, and 
		\begin{equation*}
		\Delta_j^*:=\left\{\gamma\left[(2-\gamma)\beta_j^{*2}-\left(2-\frac{2M-1}{M-1}\gamma\right)\frac{\|\beta^*_{\backslash j}\|_2^2}{M-1}\right]\right\}.
		\end{equation*} 
	\end{theorem}
	\noindent Here, $\gamma =\frac{m}{M}$ is the sampling ratio for features. The full details, including bounds for $\varepsilon$, and the proof of Theorem \ref{prop:MP_target} can be found in the Supplementary material. 
	When the minipatch size ratio $\gamma\rightarrow 0$, $\Delta_j^*\asymp 2\gamma\left(\beta_j^{*2}-\frac{\|\beta_{\backslash j}^*\|_2^2}{M-1}\right)$, which can be understood as the predictive power of feature $j$ ($\beta_j^{*2}$) compared to the average predictive power of the rest of the features ($\frac{\|\beta_{\backslash j}^*\|_2^2}{M-1}$). Under a sparse or weakly sparse setting, the latter would be a small quantity. We also note that:
	\begin{itemize}
 \vspace{-2mm}
		\item[(a)] When $\beta_j^*=0$, $\Delta_j^*= -\frac{\gamma[2(1-\gamma)M-2+\gamma]}{(M-1)^2}\|\beta^*_{\backslash j}\|_2^2\leq 0$ as long as $M\geq \frac{2-\gamma}{2-2\gamma}$;
  \vspace{-2mm}
		\item[(b)] When $\beta_j^* > \frac{2-2\gamma}{2-\gamma}\frac{\|\beta^*_{\backslash j}\|_2^2}{M-1}$, it is guaranteed that $\Delta_j^*>0$. 
	\end{itemize}
Without assuming any consistency properties of our minipatch ensembles, we prove Theorem \ref{prop:MP_target_full} by a careful analysis of the average behavior of the trained minipatch ensembles. We hope that this analysis would also inspire future analysis for more general models, and it might be of independent interest to the literature of random ensemble methods \citep{lejeune2020implicit}. Some additional comparisons with prior inference targets such as VIMP \citep{williamson2021general} and Floodgate \citep{zhang2020floodgate} are included in the Supplementary material.

\subsection{LOCO-MP Target for Correlated Features}\label{sec:dependent_features}
It is often challenging to disentangle individual feature importance under feature correlations \citep{verdinelli2024decorrelated}. Prior feature importance metrics/inference methods often fall into two categories: one focuses on the conditional importance of each feature given all others, e.g., the original LOCO method \citep{lei2018distribution}, the VIMP method \citep{williamson2021general}, and the Floodgate \citep{zhang2020floodgate}; the other type groups correlated features together (explicitly or implicitly via regularization) and assigns the same/similar importance to them \citep{buhlmann2013correlated,zou2005regularization,li2020graph}. Both types of methods have benefits and weaknesses: the first type can miss important signal features given strong correlations, while the second type may falsely select spurious noise features correlated with signal features. In practice, the choice over the two strategies depends on whether false positive or false negative is more tolerable. Other approaches have also been proposed to avoid the weakness of both types of methods \citep{verdinelli2024decorrelated,du2025disentangled}, while often requiring modeling and consistency assumptions.

Our LOCO-MP method falls into the aforementioned second category of approaches. The key idea lies that correlated features will appear in different minipatches due to random subsampling, so that the predictive power of each feature stands out in the potential absence of its correlated features. In the Supplementary material, we theoretically illustrate this phenomenon in the special case of linear models. We show that when the correlation between two features approaches $1$, their LOCO-MP feature importance scores converge to a function of the sum of their regression coefficients, thus grouping the two together. 
\subsection{Hypothesis Testing and Connections to Post-selection Inference}
The LOCO inference problem is broadly connected to the post-selection inference literature\citep{berk2013valid,lee2016exact}. As noted before, our inference target, the LOCO feature importance score, depends on the trained models $\hat{\mu}_{\backslash j},\,\hat{\mu}$ instead of being a fixed population quantity. Here, $\hat{\mu}_{\backslash j},\,\hat{\mu}$ are analogous to the selected features that will be tested for in post-selection inference \citep{lee2016exact}. Given such dependency between the inference procedure and the inference target itself, conventional inference methods often become invalid, leading to the rise of recent post-selection inference approaches. These approaches mainly fall into three categories: sample-splitting, conditional inference given the selection event, and simultaneous inference for all possible selections. However, all three types of approaches suffer from certain challenges, which are especially severe in our problem context. Sample-splitting suffers from both loss of data efficiency and interpretational challenges as discussed earlier. Conditional approaches often require assumptions on the data distribution and the selection/training procedure, not appropriate for our model-agnostic ML feature importance inference problem. Moreover, simultaneous inference approach in our context means coverage for the whole model class that $\hat{\mu},\hat{\mu}_{\backslash j}$ lie in, and hence can be highly conservative. We take a different route from these prior approaches, and directly establish a lower bound on the coverage probability $\bbP(\Delta_j\in \hat{\mathbb{C}}_j)$, where the probability is taken over both the $\Delta_j$ and $\hat{\mathbb{C}}_j$, instead of conditioning on the selection event as in \cite{lee2016exact}. More detailed discussion and comparisons are included in the Supplementary material.

In fact, the discussion above is also helpful for understanding another question: can we convert our confidence interval to hypothesis testing? In particular, suppose we would like to test whether feature $j$ affects model $\hat{\mu}$'s prediction performance, we may write the null hypothesis as $\mathcal{H}_0: \Delta_j = 0$. 
Given the confidence interval $\hat{\mathbb{C}}_j$, one natural idea is to simply reject $\mathcal{H}_0$ if $0\notin \hat{\mathbb{C}}_j$. However, due to the randomness of $\Delta_j$, $\cH_0$ is also a random event that may hold for some data sets but not for others even if they are sampled from the same distribution. This raises the question: {\em How should we define the Type I error when $\mathcal{H}_0$ is also a random event}? Is our test valid, and in what sense? As detailed in the Supplementary material, our test can control an extended notion of the conventional Type I error: $\bbP(\Delta_j=0 \ \& \ \mathcal{H}_0\text{ is rejected})\leq \alpha$. That is, the probability of falsely rejecting $\cH_0$ is bounded by $\alpha$. Here, the probability is marginalized over the random $\cH_0$, instead of conditioning on $\cH_0$ as in \citep{lee2016exact}. This notion of Type I error is similar to the strong post-selection error control in \cite{berk2013valid}, while the main difference lies that we only select one hypothesis instead of a set of hypotheses to test. More discussion on this can be found in the Supplementary material.

\section{Coverage of LOCO-MP Confidence Intervals}\label{sec:coverage_guarantee}
	In this section, we provide coverage guarantees for the confidence interval given by Algorithm \ref{algo:loco} (Section \ref{sec:coverage_guarantee}), as well as for two important variants: a slightly more conservative confidence interval with valid coverage under weaker assumptions (Section \ref{sec:variance_barrier}), and the confidence interval constructed after data-driven tuning for minipatch sizes (Section \ref{sec:datadriven_theory}). 
 Here we note that establishing coverage guarantees for LOCO-MP confidence intervals is non-trivial, due to the dependency between LOCO-LOO scores\footnote{Combing exchangeable but dependent test statistics for valid inference is still an area of active research \citep{guo2025rank}.}.
 Interestingly, by proving and exploiting a nice stability property of minipatch ensembles, as well as leveraging a recent central limit theorem for cross validation errors \citep{bayle2020cross}, we can address the challenge brought by dependency. We first define some notations.
 \paragraph{Notation} 
 For any interval $[a,b]\subset \bR$ with $a\leq b$ we use $|[a,b]|=b-a$ to denote its length. 
Let $\mu_{I,F}(X)=(H(\bX_{I,F},\bY_{I_k}))(X_{F})\in \mathbb{R}^d$ be the base model predictor trained on $(\bX_{I,F},\bY_{I})$. We define $\mu^*(\cdot;\bX,\,\bY)$ as the expected ensembled minipatch predictor, with expectation taken over the random subsampling of minipatches: $\mu^*(X;\bX,\bY) = \frac{1}{\binom{M}{m}\binom{N}{n}}\sum_{\overset{I:I\subset [N], |I|=n}{F:F\subset[M], |F|=m}}\mu_{I,F}(X).$ Let $h_j(X,Y;\boldsymbol{X},\boldsymbol{Y})$ denote the feature importance of feature $j$ evaluated at training data set $(\boldsymbol{X},\boldsymbol{Y})$ and test data point $(X,Y)$:
$h_j(X,Y;\boldsymbol{X},\boldsymbol{Y})=\err(Y,\mu^*_{\backslash j}(X_{\backslash j};\boldsymbol{X}_{:,\backslash j},\boldsymbol{Y}))-\err(Y,\mu^*(X;\boldsymbol{X},\boldsymbol{Y})),$ and let $h_j(X,Y)=\mathbb{E}_{\boldsymbol{X},\boldsymbol{Y}}[h_j(X,Y;\boldsymbol{X},\boldsymbol{Y})]$, with the expectation taken over the training data $(\boldsymbol{X},\boldsymbol{Y})$. Define $\sigma_j^2 = \mathrm{Var}(h_j(X,Y))$; we will see $\bar{\Delta}_j-\Delta_j$ will have asymptotic variance depending on $\sigma_j^2$. More details on some standard notations are also included in the web-based supporting materials.
	
\subsection{Guarantees for Feature Importance Inference}\label{sec:vanilla_coverage}
We first define a key stability quantity for the base learning algorithm $H: \cX\times \cY\rightarrow \cF$, which maps any training data set with arbitrary size to a predictor.
 \begin{definition}[Base model stability]\label{def:stb}
     Let $(X_0,Y_0),\,(X_1, Y_1),\,\cdots,\,(X_n, Y_n)$, and $(X_1',Y_1')$ be i.i.d. samples from $\cP$, with $X_i\in\bbR^M$ including $M$ features. For any feature subset $F$ of size $m$, let $\mu_{F}(\cdot)$ be the predictor trained using algorithm $H$ and training data $\{(X_{i,F}, Y_i)\}_{i=1}^n$; let $\mu_{F}'(\cdot)$ be trained with $H$ and data $\{(X_{1,F}', Y_1')\}\cup\{(X_{i,F}, Y_i)\}_{i=2}^n$. Then the stability of the base learning algorithm $H$ w.r.t. distribution $\cP$ is defined as $$\mathrm{stb}(m,n; H, \cP) = \frac{1}{\binom{M}{m}}\sum_{F\subset[M],|F|=m}\mathbb{E}\|\mu_F(X_0) - \mu_F'(X_0)\|_2^2,$$ where the expectation is taken over $(X_0,Y_0),\,\cdots,\,(X_n, Y_n)$.
 \end{definition}
 The stability notion defined above is similar to that in the prior literature \citep{bayle2020cross}, while the major difference lies that we consider the average stability across different feature subsets of a given size. In the following, we will abbreviate $\stb(m,n;H,\cP)$ to $\stb(m,n)$ since the base algorithm $H$ and data distribution $\cP$ are often clear from the context. 
	
\noindent We now state some assumptions imposed on the minipatches and the error function.
	\begin{assumption}[Lipschitz condition for Error function]\label{assump:Lip}
		The error function satisfies the Lipschitz condition with parameter $L$:
		$|\mathrm{Error}(Y,\hat{Y}_1)-\mathrm{Error}(Y,\hat{Y}_2)|\leq L\|\hat{Y}_1-\hat{Y}_2\|_2$ for any $Y\in \bR$ and predictors $\hat{Y}_1$, $\hat{Y}_2\in \bR^d$.
	\end{assumption}
	\noindent First note that the Lipschitz condition is not imposed on the loss function for training the predictive model. We only require the non-conformity score function $\err(\cdot,\cdot)$ to be Lipschitz, which defines the feature importance scores. Examples include the absolute error function for regression, and the hinge loss for classification; both are Lipstchitz with $L=1$. 
\begin{assumption}[Bounded average predictions]\label{assump:bnd_mu}
		Suppose that the average predictions of base minipatch predictors are bounded by some parameter $B>0$: 
  $$
  \frac{1}{N}\sum_{i=1}^N\mathbb{E}_{I,F}[\ind(i\notin I)\|\mu_{I,F}(X_i)\|_2^2]\leq B^2, \quad\bE_{X}\bE_{\substack{I,\,F}}\|\mu_{I,F}(X)\|_2^2\leq CB^2,
  $$ 
  where the first expectation is taken over randomly subsampled indices $I\subset[N],\,F\subset[M]$ with sizes $n$ and $m$, the second expectation is taken over the test data $X$ in addition to $I,\,F$.
	\end{assumption}
	\noindent Assumption \ref{assump:bnd_mu} requires the average/expected predictions given by models trained from random minipatches to be bounded, which is a mild assumption for standardized data sets. 
\begin{assumption}[Minipatch size and base model stability]\label{assump:mpsize}
		The minipatch sizes $(m, n)$ satisfy $\frac{n}{N},\frac{m}{M}\leq \gamma$ for some constant $0<\gamma<1$, and $n^2\stb(m,n)=o\left(\frac{\sigma_j^2}{L^2}N\right)$.
	\end{assumption}
\begin{remark}[Interplay between minipatch size and base model stability]
 Assumption \ref{assump:mpsize} can be satisfied {\em either by choosing a sufficiently small minipatch size $n$, or by deploying a sufficiently stable base model}. Bounded minipatch predictor (Assumption \ref{assump:bnd_mu}) immediately implies $\stb(m,n)\leq CB^2$ and hence Assumption \ref{assump:mpsize} holds as long as $n=o(\frac{\sigma_j}{LB}\sqrt{N})$ and $N\geq \frac{C\sigma_j^2}{L^2}$, $m\leq \gamma M$. While if the base model is known to be highly stable, e.g., $\stb(m,n;H,\cP) \leq \frac{C}{n^2}$, then the minipatch size $(m,n)$ can be larger: Assumption \ref{assump:mpsize} reduces to $\frac{n}{N}, \frac{m}{M}\leq \gamma$, $N\gg \frac{L^2}{\sigma_j^2}$. 
\end{remark}
We impose Assumption \ref{assump:mpsize} to ensure the desired stability (small squared norm difference in the predictor when swapping one training sample) for minipatch ensemble, which is a key for establishing coverage guarantee for LOCO-MP despite the dependency between the LOCO-LOO scores $\{\hat{\Delta}_j(X_i,Y_i)\}_{i=1}^N$. Leveraging algorithmic stability for distribution‐free inference has also been explored in the conformal inference literature \citep{liang2025algorithmic}. Moreover, the idea of using subsampling-based ensemble to achieve stability has also appeared in recent work \citep{soloff2024bagging}, although slightly different but related stability notions were considered.
We will also show how Assumption \ref{assump:mpsize} can be further relaxed for the coverage of a slightly more conservative confidence interval. 

	\begin{assumption}[Number of minipatches]\label{assump:mpnumber}
		The number of random minipatches $K$ satisfies
		$K\gg \left(\frac{L^2B^2N}{\sigma_j^{2}}+1\right)\log N.$
	\end{assumption}

	\noindent $K$ needs to be sufficiently large to control the level of subsampling randomness.
\begin{assumption}\label{assump:third_moment}
    The normalized feature importance r.v. satisfies the third moment condition: $\bE[h_j(X_i,\,Y_i) - \bE h_j(X_i,\,Y_i)]^3/\sigma_j^3\leq C$.
\end{assumption}
The third-moment condition on $h_j(X_i,\,Y_i)$ is to ensure the uniform integrability condition and helps us establish the central limit theorem. With these regularity conditions in place, we are ready to state our main theoretical result. 

\begin{theorem}[Coverage Guarantee]\label{cor:coverage-width}
		 Suppose that all training data $(X_i, Y_i)\overset{i.i.d}{\sim}\mathcal{P}$ and Assumptions \ref{assump:Lip}-\ref{assump:third_moment} hold. Then we have
		$\lim_{N\rightarrow \infty}\bP(\Delta_j\in \hat{\bC}_j)=1-\alpha$, where $\hat{\bC}_j$ is the output of Algorithm \ref{algo:loco}, and $\Delta_j$ is as defined in Section \ref{sec:LOCO-MP}.
\end{theorem}
 \vspace{-1mm}
 \noindent Theorem \ref{cor:coverage-width} shows that, under certain assumptions on the minipatch learning algorithm, our confidence interval $\hat{\mathbb{C}}_j$ constructed in Algorithm \ref{algo:loco} has asymptotically valid coverage for the feature importance score $\Delta_j$ associated with the current minipatch predictor trained with all available data. As explained earlier in Section \ref{sec:target_discussion}, $\Delta_j$ is the expected predictive improvement when including feature $j$ compared to excluding it, when we train the predictive model using the current training data and minipatch learning algorithm. 

The key of our proof lies in proving an attractive stability property for the ensembled minipatch predictor. In particular, the stability quantity (a similar notion to Definition \ref{def:stb}) of the ensembled MP predictor is $o_p(N^{-\frac{1}{2}})$ under Assumption \ref{assump:mpsize}. This is an essential prerequisite for us to utilize the central limit theorem in \cite{bayle2020cross}, originally proposed for cross-validation errors. More extensive theoretical results and their detailed proofs are included in the Supplementary material.

%%------------------------%%

 \subsection{Buffered CIs: Valid Coverage in Broader Scenarios}\label{sec:variance_barrier} 
 We have now provided a coverage guarantee (Corollary \ref{cor:coverage-width}) for our confidence interval $\hat{\mathbb{C}}_j$, with the help of Assumptions \ref{assump:Lip} - \ref{assump:mpnumber} on our minipatch learning algorithm. In particular, Assumption \ref{assump:mpsize} requires either the minipatch size $n$ to be sufficiently small or the base algorithm $H$ to be sufficiently stable; Assumption \ref{assump:mpnumber} also requires the number of minipatches to be much larger than $(\frac{L^2B^2N}{\sigma_j^2}+1)\log N$. However, small minipatch sizes lead to stronger regularization which may not give good predictions when the noise level is low; certain popular base learners such as decision trees are often not stable; and finally, the variance $\sigma_j^2$ of the test statistic can be close to zero when we test for a noise feature $j$ \citep[see e.g.,][]{rinaldo2019bootstrapping,verdinelli2024decorrelated, williamson2021general, dai2022significance}, which makes Assumptions \ref{assump:mpsize}-\ref{assump:mpnumber} more stringent. Although the first two challenges are tied to the minipatch learning algorithm, the last vanishing variance issue is typically seen in occlusion-based feature importance inference literature. Many prior works propose to manually inject noise, to inflate the variance estimate, or to further split the data \citep{rinaldo2019bootstrapping,verdinelli2024decorrelated, williamson2021general, dai2022significance}.

To address this challenging problem in our framework, we propose a theory-inspired strategy to tackle the coverage issue even when Assumptions \ref{assump:mpsize}-\ref{assump:mpnumber} are violated. Instead of splitting the data or manually injecting noise into the data, we consider a simple yet effective approach: adding a small barrier value $\epsilon(N)$ to the estimated standard deviation $\frac{\hat{\sigma}_j}{\sqrt{N}}$ for the test statistic, similar to \cite{verdinelli2024decorrelated}. Here we use the notation $\epsilon(N)$ to emphasize that the barrier value depends on $N$ instead of being a constant, and its scaling will be specified shortly. Now our new confidence interval becomes:
	\begin{equation}\label{eq:CI_barrier}
	\hat{\mathbb{C}}^{\mathrm{barrier}}_j=\Big[\bar{\Delta}_j - z_{\alpha/2}\max\Big\{\frac{\hat{\sigma}_j}{\sqrt{N}},\epsilon(N)\Big\}, \bar{\Delta}_j + z_{\alpha/2}\max\Big\{\frac{\hat{\sigma}_j}{\sqrt{N}},\epsilon(N)\Big\}\Big].
	\end{equation}
	This is slightly more conservative than our original confidence interval, and hence the coverage guarantee in Theorem \ref{cor:coverage-width} also holds for $\hat{\mathbb{C}}^{\mathrm{barrier}}_j$. Furthermore, we can also relax Assumptions \ref{assump:mpsize} and \ref{assump:mpnumber} to the following:
	
 \begin{assumption}[Variance barrier $\epsilon(N)$]\label{assump:buffer}
		The minipatch sizes satisfy $\frac{n}{N},\frac{m}{M}\leq \gamma$ for a constant $0<\gamma<1$, 
		\begin{equation}\label{eq:var_barrier}
        \epsilon(N)\geq \frac{cLn\sqrt{\stb(m,n)}}{N}\log N,
		\end{equation} for some constant $c>0$, and the number of random minipatches
		$K\gg \frac{B^2}{\stb(m,n)}\frac{N^2}{n^2\log N} + \log N.$
	\end{assumption}

\begin{theorem}\label{thm:coverage_buffer}
    Suppose that all training data $(X_i, Y_i)\overset{i.i.d}{\sim}\mathcal{P}$, Assumptions \ref{assump:Lip}, \ref{assump:bnd_mu}, \ref{assump:buffer}, and \ref{assump:third_moment} hold. Then we have $\liminf_{N\rightarrow \infty}\bP(\Delta_j\in \hat{\bC}_j^{\mathrm{barrier}})\geq 1-\alpha$, where $\hat{\bC}_j^{\mathrm{barrier}}$ is as defined in \eqref{eq:CI_barrier}, and $\Delta_j$ is as defined in Section \ref{sec:LOCO-MP}.
\end{theorem} 

\noindent Theorem \ref{thm:coverage_buffer} suggests that if setting $\epsilon(N)$ appropriately, the coverage of $\hat{\mathbb{C}}_j^{\mathrm{barrier}}$ only requires mild assumptions on the minipatch size $m,\,n$, and no assumption on the base model stability. Furthermore, a vanishing variance $\sigma_j^2$ does not affect the coverage guarantee. 
	\begin{remark}[Choice of $\epsilon(N)$]
		Such a choice of the variance barrier $\epsilon(N)$ as in \eqref{eq:var_barrier} is not heuristic, but has a theoretical foundation. One practical challenge is that the unknown value $\stb(m,n)$ (variability of minipatch predictors) may vary across data sets and training algorithms. For practical considerations, we propose to estimate $\stb(m,n)$ in a data-driven manner, denoted by $\hat{\delta}$, and then we plug it into \eqref{eq:CI_barrier} with $\epsilon(N) = \frac{c_0L\sqrt{\hat{\delta}}n}{N}\log N$ for a small constant $c_0$ (set as 0.005 throughout our empirical studies). The detailed estimation procedure for $\stb(m,n)$ is summarized in the Supplementary material. 
\end{remark}
\begin{remark}[Practical recommendation]
    A variance barrier in the confidence interval is only a partial solution to this universal vanishing variance challenge in occlusion-based inference, as it sacrifices statistical efficiency to ensure inference validity. Interestingly, as we will show in empirical studies, such tension between coverage and efficiency occurs only for noise features but not for signal features (the variance barrier often does not take effect for the latter). Therefore, in the case where only strong signal features are of primary interest, using the barrier and losing some efficiency for (near) noise features may be acceptable. If not using the barrier, the practitioner needs to avoid interpreting tiny confidence intervals for feature importance that are close to zero.
\end{remark}
\subsection{Minipatch LOCO Inference with Data-driven Tuning}\label{sec:datadriven_theory}
In practice, one needs select the minipatch sizes appropriately as they control the regularization strength and have non-negligible effect on the prediction performance of the ensembled predictor. Here we propose to select minipatch sizes based on the leave-one-observation-out prediction errors. In particular, we can train minipatch ensembles with a set of candidate minipatch sizes $\{(m_1,n_1),\,\dots,\,(m_s,n_s)\}$, and set $(\hat{m},\hat{n})$ as the minipatch size with the lowest LOO residuals. The detailed procedure is summarized in Algorithm \ref{algo:MPsize_tuning} in the Supplementary material. One important question is, with data-driven selection of the minipatch sizes, can we still perform valid LOCO inference using the original algorithm anymore? This is a non-trivial question as we use the same data for hyperparameter tuning, model training, and statistical inference (consider selective inference with data-driven hyperparameter tuning). Interestingly, we will show that our original LOCO confidence interval in Algorithm \ref{algo:loco} with data-driven selection of minipatch sizes can still have asymptotically valid coverage without adding significant assumptions.

In particular, the inference target $\Delta_j$ can still be written as in \eqref{eq:target_inference}, where the minipatch sizes associated with the predictors $\mu(\cdot;\bX,\bY)$ and $\mu_{\backslash j}(\cdot;\bX_{:,\backslash j},\bY)$ are $(\hat{m},\hat{n})$ selected by Algorithm \ref{algo:MPsize_tuning}. For theoretical purposes, we also define $(m^{\mathrm{oracle}}, n^{\mathrm{oracle}})\in\{(m_1,n_1),\,\dots,\,(m_s,n_s)\}$ as the minimizer of the population LOO residual (see Definition \ref{def:mn_oracle} in the Supplementary material); let $h_j(X,\,Y)$ and $\sigma_j^2$ be defined similarly as in Section \ref{sec:vanilla_coverage}, when the oracle minipatch size $(m^{\mathrm{oracle}}, n^{\mathrm{oracle}})$ are used in the ensembled predictor.
 
 \begin{theorem}[Coverage with data-driven minipatch sizes]\label{thm:datadriven_coverage}
    Suppose we run Algorithm \ref{algo:loco} with the data-driven selection of minipatch size $(\hat{m},\hat{n})$ by Algorithm \ref{algo:MPsize_tuning} to obtain $\widehat{\bC}_j$. Assume that the number $s$ of candidate minipatch sizes is bounded, Assumptions \ref{assump:Lip}, \ref{assump:mpnumber}, \ref{assump:third_moment}, \ref{assump:delta_LOO_bnd} hold, and with the new $\sigma_j^2$ defined above, Assumptions \ref{assump:bnd_mu}, \ref{assump:mpsize} hold for all candidate minipatch sizes $\{(m_l,\,n_l)\}_{l=1}^s$. Then we have $\lim_{N\rightarrow\infty}\bP(\Delta_j\in\hat{\mathbb{C}}_j) = 1-\alpha$, with $\Delta_j$ defined in Section \ref{sec:LOCO-MP} for minipatch predictors $\mu$ and $\mu_{\backslash j}$ with size $(\hat{m},\hat{n})$.
 \end{theorem}
 Theorem \ref{thm:datadriven_coverage} suggests that our confidence interval is still asymptotically valid with data-driven selection of the minipatch sizes, as long as all candidate minipatch sizes we consider are reasonably chosen. Assumption \ref{assump:delta_LOO_bnd} is included in the Supplementary material; it requires the prediction performance gap between the oracle and the sub-optimal minipatch sizes to be lower bounded. The proof of Theorem \ref{thm:datadriven_coverage} is non-trivial; it also reveals the inherent stability of the tuning procedure on minipatch ensembles. Similar to the scenario with a fixed minipatch size, we can also apply the variance barrier to the confidence interval to relax the assumptions on minipatch sizes or base model stability. More details and theoretical guarantees are included in the Supplementary material. 

\section{Empirical Studies}\label{sec:result} 
	We empirically validate our LOCO-MP approach with synthetic and real data sets, showing that it has several advantages over existing methods. A Python package for implementing LOCO-MP is available online: \href{https://github.com/DataSlingers/LOCOMP}{https://github.com/DataSlingers/LOCOMP}. Extra empirical details and results can be found in the Supplementary material.

	%%%%%%%%%%%%%%%%%%%%%%%%%%
	\subsection{Simulation Studies}
    \label{section:sim_framework}
	We propose two simulation models to assess our feature importance inference method: a sparse linear and non-linear model, for both regression and classification tasks. 
    For the linear simulation, we set $f(\mathbf{X}) = \boldsymbol{X}^{\top}_{1:5}\boldsymbol{\beta}$ where $\boldsymbol{\beta} = [ 3, 2.5, 2, 1.5, 1]$;
    for the nonlinear model, we set $f(\mathbf{X}) =3 \ind_{[-2,2]}(X_1)X_1+2.5\max(0,X_2)+2\min(0, X_3)+1.5\max(0,X_4)+\mathrm{sign}(X_5)$.  For regression tasks, we let $Y = f(\mathbf{X}) + \epsilon$ with $\epsilon \sim N(0, \mathbf{I})$; and for classification tasks, we generate $Y$ as: $Y \sim \mathrm{Bernoulli}( \frac{1}{1 + e^{-f(\mathbf{X})}})$. We set the number of features as $M = 50$, generating $\mathbf{X}$ from the standard normal distribution (a correlated setting is also considered later). The sample size $N$ ranges from $100$ to $2000$. The minipatch size and number are set as $m = 0.5M$, $n = N^{0.8}$, and $K= 10,000$. For the base estimators, we use (logistic) ridge regression, decision trees, and kernel ridge or SVMs; we set coverage level $1-\alpha = 0.9$. Throughout this section, we present the results for our CI with the variance barrier \eqref{eq:CI_barrier}; results without the barrier are included in the Supplementary material. The variance barrier is set as $\epsilon(N)=\frac{c_0Ln\sqrt{\hat{\stb}(m,n)}}{N}\log N$, where $c_0=0.005$, and $\hat{\stb}(m,n)$ is an estimated base model stability using Algo. \ref{algo:stability_est1} in the Supplementary material. Additionally, we use the error function $\err(Y,\mu(X))=|Y-\mu(X)|$ for regression and $\err(Y,\mu(X))=1-[\mu(X)]_Y$ for classification ($\mu(X)$ is a predicted probability vector).

\begin{figure}
\centering
    \includegraphics[width=.9\textwidth]{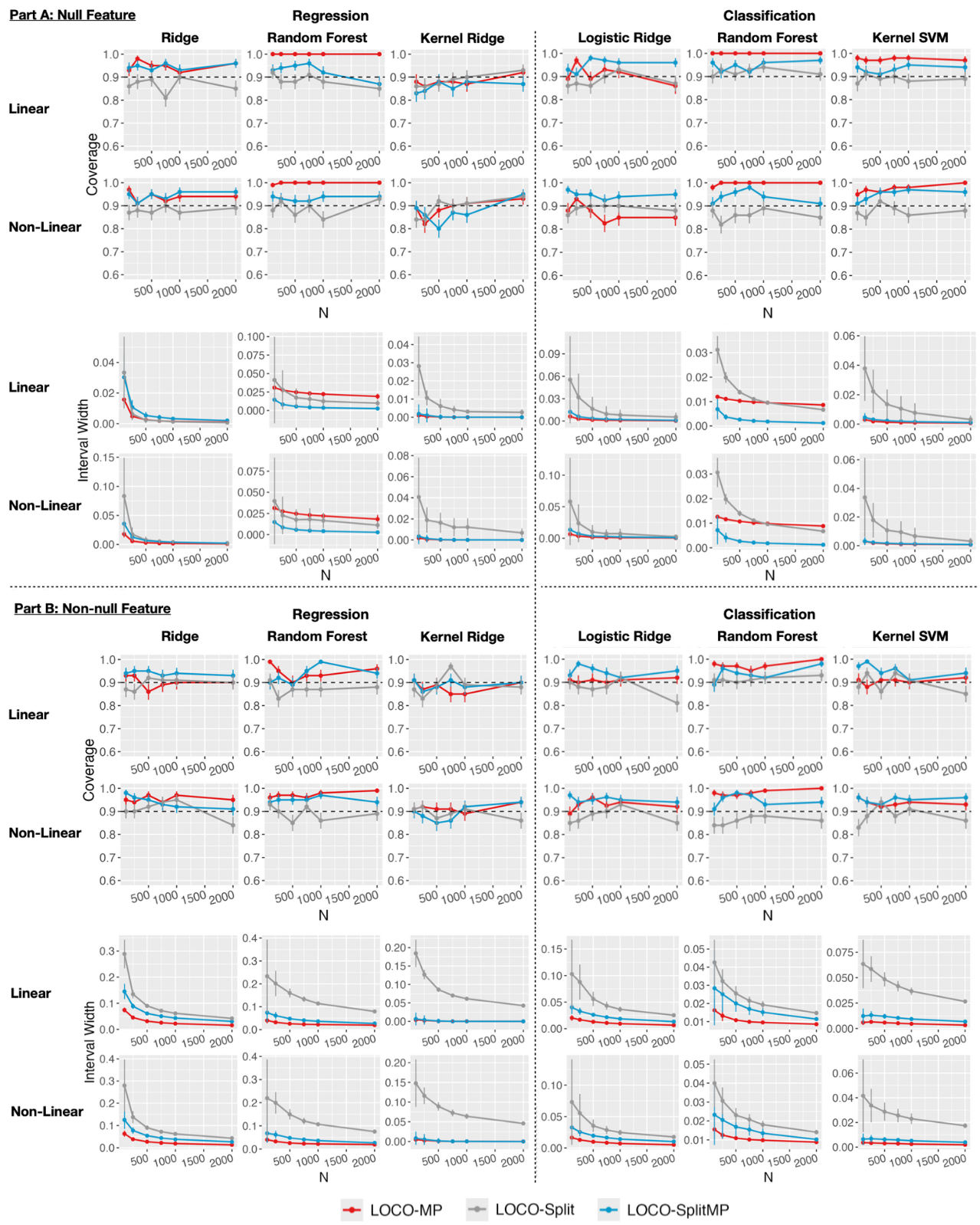}
\caption{\label{fig:validate1} \small Coverage of the inference target \eqref{eq:target_inference} and CI width for a null feature (Part A) and a non-null feature with SNR of 2 (Part B).  Left panels show results for regression tasks and the right panels for classification tasks under both linear and non-linear simulation designs as described in Section~\ref{section:sim_framework}.  Our LOCO-MP CIs employ a buffered interval defined in \eqref{eq:CI_barrier}. Results validate the asymptotically valid coverage of LOCO-MP intervals, as well as its shorter widths in most senarios.}
\end{figure}

 %%%%%%%%%%%%%%%%%%%%
	We demonstrate the coverage and width of our CIs for a null feature and a signal feature across different base models and simulation scenarios in Figure~\ref{fig:validate1}.  
    Since the exact value of the inference target $\Delta_j$ in \eqref{eq:target_inference} involves an expectation and cannot be exactly computed, we approximate it using the Monte Carlo method with $10,000$ test data points; more details are in the Supplementary materials. 
	We also implement two comparison methods: the LOCO-Split method with the same ML model as the base model for LOCO-MP, and a LOCO-SplitMP method,  which is LOCO-Split with our minipatch ensemble estimator as the prediction algorithm; we consider the LOCO-SplitMP method since it shares a very similar inference target with ours. Both methods use 75\% train and 25\% test split. (Recall from Section~\ref{sec:methods} that our inference target is not comparable to many prior population feature importance inference methods since our target is associated with the trained minipatch predictor instead of a population quantity.) 
    For LOCO-Split, we tune hyperparameters for the ML models via cross-validation whereas for minipatches, hyperparameters are set to a fixed small value.
    We evaluate the coverage and width of the confidence intervals constructed from $100$ replicates. Figure~\ref{fig:validate1} shows that LOCO-MP exhibits valid coverage rates in all scenarios and generates more efficient intervals with smaller widths decreasing as $N$ increases. Our intervals are only wider for the null feature and random forest model\footnote{For the random forest experiments, we set the minipatch base models for both LOCO-MP and LOCO-SplitMP as decision trees and train a random forest for LOCO-Split, as the ensembled MP trees are very similar to a random forest.}. This is due to the variance barrier: as discussed in Section \ref{sec:variance_barrier}, the barrier ensures the coverage for possibly unstable base models (like decision trees) while sometimes sacrificing efficiency for the null feature. Results without barrier are included in the Supplementary material, where we always have smaller width, but have a very slight loss of coverage for the null feature with logistic ridge under non-linear classification models. 
    Additional details of the empirical setup as well as coverage and width results with different minipatch sizes ($m = \sqrt{M}, n = \sqrt{N}$) can be found in the Supplementary material. 

\begin{figure}
    \centering
    \includegraphics[width=.9\textwidth]{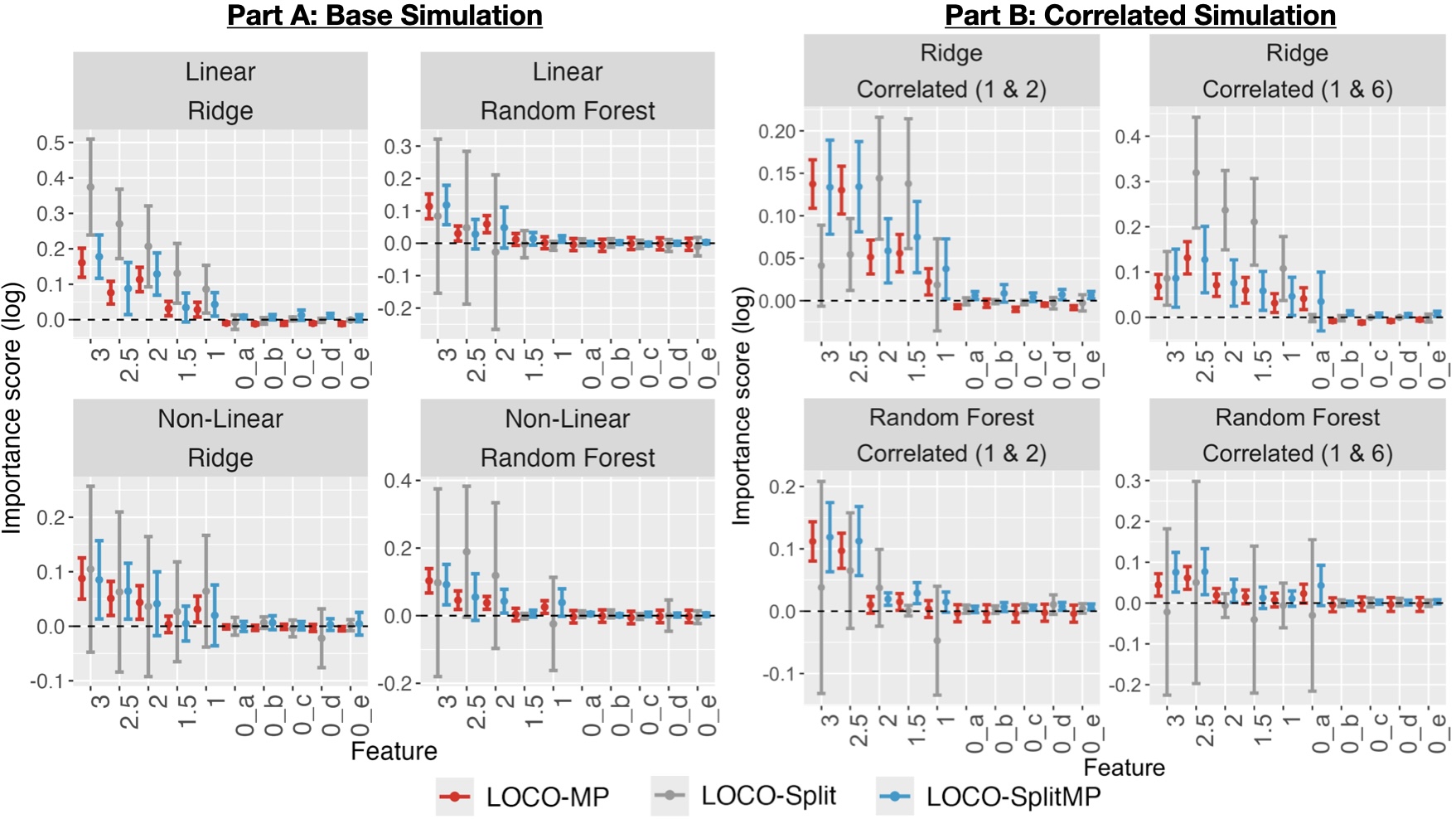}
    \caption{\small Comparative LOCO CIs (Bonferroni-corrected) in the base and correlated, linear and non-linear simulation settings, for ridge and random forest regression; the SNR of each feature is denoted on the x-axis. LOCO-MP CIs have smaller width and reveal more statistically significant signal features; they are especially powerful in identifying correlated signal features (1 \& 2). }
    \label{fig:loco_interval}
\end{figure}

Next, we visually compare the Bonferroni-corrected CIs of LOCO-MP to those of LOCO-Split \citep{lei2018distribution} and LOCO-SplitMP for a subset of our simulation scenarios with $N=200$ in Figure~\ref{fig:loco_interval}, part A.  (Recall that the targets for LOCO-SplitMP and LOCO-Split are slightly different than for our LOCO-MP, as they are associated with the models trained on a data split).  LOCO-MP provides more efficient intervals than both LOCO-Split and LOCO-SplitMP, while LOCO-Split fails to identify any significant features in the nonlinear regression data; additional interval comparisons are in the Supplementary material.  

Additionally as discussed in Section \ref{sec:dependent_features} and in contrast to other feature inference approaches including LOCO-Split, LOCO-MP has advantages in dealing with feature correlations by implicitly grouping correlated features. To demonstrate this, in Figure~\ref{fig:loco_interval} Part B, we consider a correlated linear simulation where all feature pairs are uncorrelated except for one pair with correlation $\rho=0.9$ as specified; the simulations are otherwise identical to those previously described.  When the two strongest signal features (1 \& 2) are correlated, notice that LOCO-Split fails to identify the most important feature as expected, but our method correctly identifies both features as statistically significant.

\subsection{Case Study: Genetic Biomarkers for Alzheimer's Disease}

We apply our method and several comparison methods to obtain feature importance CIs for transcriptomic biomarkers predictive of cognition in an Alzheimer's Disease (AD) study. Using the bulk RNA-sequencing data obtained from brain tissue in the Religious Orders Study Memory and Aging Project (ROSMAP) \citep{bennett2018religious}, we build regression models to predict the last global cognition score available for the subject. The data set has $n = 507$ samples and we aggressively filter genes down to $p=86$ features, using a high variance filter (variance $\geq 0.5$); note that we filtered features to this smaller number due to the computationally intensive nature of several comparison methods.  

We compare our LOCO-MP approach to LOCO-Split \citep{lei2018distribution} as well as other model-agnostic population feature importance inference methods: CPI \citep{watson2021testing}, VIMP \citep{williamson2021general}, GCM \citep{shah2020hardness}, and floodgate \citep{zhang2020floodgate}.  As previously noted, all of these methods have different targets of inference which are not directly comparable.  They all seek to identify important features, however, and hence we evaluate comparison methods for this task using a separate test set.  We compare all inference techniques for a random forest regressor with $200$ trees; for our minipatch ensemble we use decision trees as the base learner with $m = 0.5M$, $n = N^{0.8}$, and $K = 10,000$.  Using 70\% of the samples, we train models and conduct inference with $\alpha = 0.1$ and a Bonferroni correction for multiplicity.  

\begin{figure}
\includegraphics[width=\textwidth]{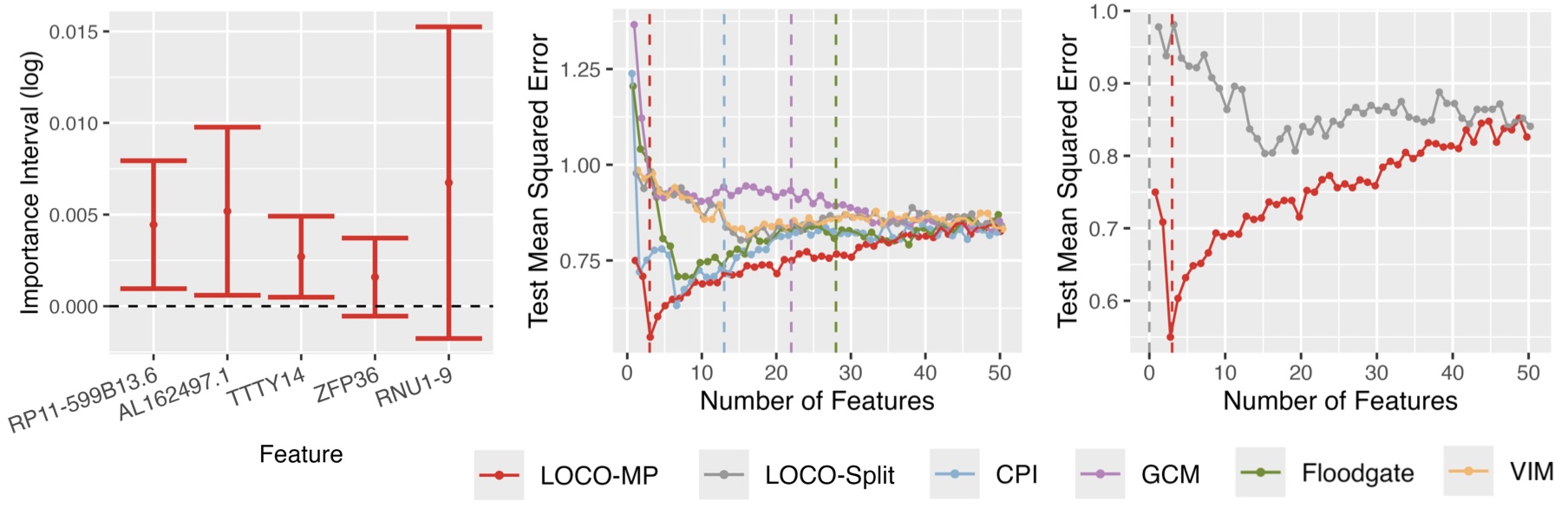}
\caption{\small Epigenetics of Alzheimer's Disease Case Study.  (Left) Top feature importance intervals (Bonferroni corrected) for LOCO-MP using decision tree base models to predict global cognition (regression task). (Middle (all methods) and Right (zoom in on LOCO-MP vs. LOCO-Split)) Test mean squared error for ablation experiments for LOCO-MP and comparison methods; vertical dashed lines indicate the number of statistically significant features selected by each method (Bonferroni corrected). LOCO-MP discovers three significant genes that also are highly predictive in the ablation experiment (lower test error), thus validating our approach.}
\label{fig:rosmap}
\end{figure}

In Figure~\ref{fig:rosmap}, we present the top five important features for LOCO-MP, three of which are statistically significant.  In comparison, LOCO-Split and VIMP identify no significant features; GCM finds 22, CPI finds 13, and Floodgate finds 28 important features, hinting that these methods may be mis-calibrated. The particular features identified by each method are given in the Supplemental material. Furthermore, we use the 30\% of samples set aside as a test set to sequentially evaluate features identified by each inference procedure.  We rank the top significant features for each method, and fit a random forest model with only the top $K$ ($K = 1,\dots, 50$) features on the training set.  We then apply these models to make predictions on the test set and report the test mean squared error (MSE) in Figure~\ref{fig:rosmap} (b) and (c) (showing a zoomed-in comparison of LOCO-MP and LOCO-Split); vertical dashed lines indicate the number of features that are identified as significant for the corresponding method.  LOCO-MP offers the smallest test MSE, indicating better predictability with the features identified as statistically significant. Additionally, the test MSE increases as we include insignificant features, which further indicates that LOCO-MP is well-calibrated, contrasting with other inference approaches. Scientifically, LOCO-MP also identifies interesting biomarkers associated with cognition in AD: TTTY14 is known to be upregulated in AD  microglia \cite{wang2024single}; RP11-599B13.6 is related to sleep disorder, further associated with cognition \citep{leng2017association}; and AL162497.1 is a long non-coding RNA (lncRNA) which regulates gene expression and have been implicated in neurodegenerative diseases such as AD \cite{huang2024elucidating}. 

\section{Discussion}\label{sec:disc}
In this paper, we propose a novel ensemble framework that seamlessly integrates predictive model training with uncertainty quantification for both model interpretations and predictions. This framework is almost model-agnostic as it can be applied with any base model, for regression or classification tasks. Once trained, no extra computation or held-out data is needed for generating confidence intervals for its interpretation, in the form of leave-one-covariate-out (LOCO) feature importance, as well as predictive intervals. Avoiding model-refitting and data-splitting, our framework is highly efficient both statistically and computationally. We achieve this by leveraging a new ensemble structure, termed minipatch ensembles, which involves double sub-sampling of both observations and features. 
Furthermore, our approach is distribution-free, assumption-light, and asymptotically valid despite the involved dependency between training and inference, brought by avoiding data-splitting. We also address a number of notoriously common but important challenges to broaden the applicability of our framework: we show that the statistical validity is preserved almost for free even after data-driven tuning for minipatch sizes; we show via partial theory and empirical studies the advantages of our approach for dealing with correlated features; we give a theory-grounded solution to the vanishing variance issue widely seen for occlusion-based feature importance inference. Furthermore, we illustrate intriguing connections between our problem and selective inference, as well as how our LOCO feature importance score relates to conventional population feature importance studied in prior works. Finally, while deriving our inference validity theory, we prove and leverage the inherent predictive stability of minipatch ensembles, which could be of independent interest. Various statistical advantages of our framework are also demonstrated via extensive empirical studies.

As a new framework, there remain many open research questions related to our work. For instance, one might wonder the impact of constraining the ML model to be minipatch ensembles.  We emphasize that minipatch ensembles are similar to double bagging, which is commonly used throughout statistics and machine learning. Further, we conjecture that minipatch ensembles can be viewed as implicitly regularizing the base model. This connection was made explicit for minipatch ensembles of linear models by \citep{lejeune2020implicit,yao2021minipatch,patil2023asymptotically}; \citep{mentch2020randomization} also showed that feature subsampling employed by random forests (similar to minipatch ensembles of trees) has an implicit regularization effect. Another open problem is adjusting for multiplicity when there are many features of interest. Throughout our empirical studies, Bonferroni correction was applied for simultaneous coverage, while it is future interest to develop methods that control false coverage rate. Furthermore, in high-dimensional settings, we may want to focus on the uncertainty quantification for top features, essentially a selective inference problem. Finally, our current minipatch training process utilizes uniformly subsampled minipatches, while some carefully designed adaptive sampling procedure may further improve its predictive accuracy and inference efficiency. Overall, LOCO-MP provides a powerful framework and opens up exciting new avenues for uncertainty quantification in ML interpretation.

\section*{Acknowledgments}
The authors acknowledge support from NSF DMS-1554821, NSF NeuroNex-1707400, and NIH 1R01GM140468. LG additionally acknowledges support from the Ken Kennedy Institute 2021/22 Shell Graduate Fellowship. The authors would also like to thank Daniel Lejeune, Ryan Tibshirani, Rina Foygel Barber, and Larry Wasserman for valuable discussions and helpful feedback on an earlier draft.
\newpage
\appendix
\section{Additional Details of the Methods}
The construction of the variance barrier requires the stability of the base model $\stb(m,\,n;\,\mathcal{H},\,\mathcal{P})$. We propose to estimate it as follows:

After minipatch training, we randomly subsample 20 minipatches that we already trained. For each minipatch, we randomly substitute one of its sample with another sample out of this minipatch, and then train a base model on this new minipatch. This leads to 20 pairs of minipatches where only one sample is different within each pair. For each pair of MP predictors, we apply them on all the out-of-patch samples for prediction and compare their prediction differences. We estimate $\stb(m,n)$ by the average difference across the out-of-patch samples and the 20 pairs of minipatches. The detailed algorithm is summarized in Alg. \ref{algo:stability_est1}.

 \begin{algorithm}[ht!]
		\noindent{\textbf{Input}}: Training samples ($\boldsymbol{X},\boldsymbol{Y}$), minipatch sizes $n$, $m$; number of minipatches for stability estimation $K'$; base learner $H$; trained $K$ minipatch predictors; small constant $c_0>0$.
		\begin{enumerate}
			\item Randomly select $K'$ minipatches from the set of trained minipatches: $\tilde{I}_{1},\,\tilde{I}_{2},\,\dots,\,\tilde{I}_{K'}$; $\tilde{F}_{1},\,\tilde{F}_{2},\,\dots,\,\tilde{F}_{K'}$.
            \item For each reselected minipatch ($\tilde{I}_{k}$ $\tilde{F}_k$):
            \begin{enumerate}
                \item Randomly choose $i\in \tilde{I}_k$ and $i'\notin \tilde{I}_{k}$; then let $\tilde{I}'_k = \tilde{I}_{k}\cup\{i'\}\backslash \{i\}$.
                \item Train a base model on $(\tilde{I}'_k,\,\tilde{F}_k)$: $\mu_{\tilde{I}'_k,\,\tilde{F}_k}(X)$
                \item For samples $l\notin \tilde{I}_k\cup\tilde{I}'_k$, compute $\|\mu_{\tilde{I}_k,\,\tilde{F}_k}(X_l)-\mu_{\tilde{I}'_k,\,\tilde{F}_k}(X_l)\|_2^2$ and take an average: $$\delta_k = \frac{1}{N-n-1}\sum_{l\notin \tilde{I}_k\cup\tilde{I}'_k} \|\mu_{\tilde{I}_k,\,\tilde{F}_k}(X_l)-\mu_{\tilde{I}'_k,\,\tilde{F}_k}(X_l)\|_2^2.$$
            \end{enumerate}
        \item Compute the average $\hat{\delta} = \frac{1}{K'}\delta_k$ as the final estimate of the stability score.
		\end{enumerate}
		\textbf{Output}: $\hat{\delta}$ as an estimate for $\stb(m,n)$. 
		\caption{Base Model Stability Estimation}
		\label{algo:stability_est1}
	\end{algorithm}

Here, we also present the detailed steps for tuning the minipatch size using LOO errors in Algorithm~\ref{algo:MPsize_tuning}.
\begin{algorithm}[ht!]
	\noindent{\textbf{Input}}: Training pairs ($\boldsymbol{X},\boldsymbol{Y}$), a set of candidate minipatch sizes $S=\{(m_1,\,n_1),\dots,(m_s,\,n_s)\}$; number of minipatches $K$; base learner $H$.
	\begin{enumerate}
        \item For $l=1,\dots,s$
        \begin{enumerate}
            \item Perform minipatch learning with minipatch size $(m_l,n_l)$: for $k=1,...,K$
		  \begin{enumerate}
				\item Randomly subsample $n_l$ observations, $I_k \subset [N]$, and $m_l$ features, $F_k \subset [M]$.
				\item Train prediction model $\hat{\mu}_k$ on $(\boldsymbol{X}_{I_k,F_k}, \boldsymbol{Y}_{I_k})$: for any $X\in \bR^M$, $\hat{\mu}_k(X) = H(\boldsymbol{X}_{I_k,F_k},\boldsymbol{Y}_{I_k})(X_{F_k})$.
			\end{enumerate}
            \item Obtain LOO predictions:\\
			$\hat{\mu}_{-i}(X_i) = \frac{1}{\sum_{k=1}^{K} \mathbb{I}(i \notin I_k)} \sum_{k=1}^{K} \mathbb{I}(i \notin I_{k}) \hat{\mu}_{k} (X_{i})$;
            \item Calculate the average LOO residual: \\
			$\mathrm{Err}_l=\frac{1}{N}\sum_{i=1}^N\mathrm{Error}(Y_{i},\hat{\mu}_{-i}(X_i) )$; 
        \end{enumerate}
		\item Find the minipatch size pair with the lowest average LOO error: $\hat{m}=m_{\hat{l}}$, $\hat{n} = n_{\hat{l}}$, $\hat{l}=\argmin_{1\leq l\leq s} \mathrm{Err}_l$.
		\end{enumerate}
		\textbf{Output}: Minipatch size pair $(\hat{m},\,\hat{n})$.
		\caption{Data-driven Tuning for Minipatch Sizes}
		\label{algo:MPsize_tuning}
	\end{algorithm}
    
\subsection{Discussion: Relationship to Selective Inference}\label{sec:selective_inference}
The LOCO inference problem is very different from conventional/textbook statistical inference. This is because the inference target, LOCO feature importance score, is a quantity that depends on the trained model and training data, instead of being a fixed population quantity determined before seeing the data, like the population mean or a linear model parameter. This characteristic of LOCO inference connects it to the recent selective/post-selection inference literature \citep{kuchibhotla2022post,berk2013valid,lee2016exact}. Post-selection inference often refers to the inference problem that targets at some parameter selected by exploiting the data \citep{kuchibhotla2022post}. Perhaps the most notable example of post-selection inference problem is testing the significance of a regressor after variable selection using the Lasso \citep{lee2016exact}. Although LOCO inference arises from a completely different motivation and problem context, it shares the essential idea and challenge with post-selection inference: the inference target is constructed after some model selection/training procedure.

In particular, recall that our inference target $\Delta_j$ takes the following form:
\begin{equation*}
    \Delta_j = \bE[\err(Y,\hat{\mu}_{\backslash j}(X_{\backslash j};\bX_{:,\backslash j}, \bY) - \err(Y,\hat{\mu}(X;\bX, \bY)|\bX,\bY],
\end{equation*}
where we use the notations $\hat{\mu}_{\backslash j},\,\hat{\mu}$ instead of $\mu_{\backslash j},\,\mu$ in the main paper to emphasize that they are trained from data. Given a pair of predictive models $\mu_1,\,\mu_2$, define the predictiveness gap between them as follows:
\begin{equation*}
    \beta_{(\mu_1,\mu_2)}=\bE[\err(Y,\mu_1(X) - \err(Y,\mu_2(X))],
\end{equation*}
where the expectation is taken over the test data $(X,\,Y)$. $\beta_{(\mu_1,\mu_2)}$ is a function that maps two predictive models to a scalar. Let $q=(\mu_1,\mu_2)$ denote a model pair to be tested for, and let $\hat{q} = (\hat{\mu}_{\backslash j},\hat{\mu})$ the trained model pair, then our inference target can be written as
\begin{equation}\label{eq:target_selective_inference}
    \Delta_j = \beta_{\hat{q}},
\end{equation}
the evaluation of the map $q\rightarrow \beta_{q}$ evaluated at the models $\hat{q}=(\hat{\mu}_{\backslash j},\,\hat{\mu})$ trained on data $(\bX,\,\bY)$. In a recent survey paper by \cite{kuchibhotla2022post}, the authors discuss post-selection inference as an example of a more general problem formulation termed as ``Valid Inference after Data Exploration (VIDE)", where they aim to conduct inference for parameter $\beta_{\hat{q}}$ that depends on the data implicitly through some selection event $\hat{q}$, and $\hat{q}$ is determined by exploiting data. By writing our target as in \eqref{eq:target_selective_inference}, we can more clearly see its connection to this VIDE problem. The goal of the VIDE problem is to construct a confidence interval $\widehat{\mathrm{CI}}_{\hat{q}}$, such that
\begin{equation}\label{eq:VIDE}
    \lim\inf_{n\rightarrow \infty}\bP(\beta_{\hat{q}}\in \widehat{\mathrm{CI}}_{\hat{q}})\geq 1-\alpha,
\end{equation}
where the probability is taken over the randomness of both $\hat{q}$ and $\mathrm{CI}_{\hat{q}}$.
We have established the above validity result for our LOCO-MP confidence intervals in Theorems \ref{cor:coverage-width} and \ref{thm:coverage_buffer} in the main paper.

However, our approach and theoretical guarantee is different from many prior works in post-selection inference. Most prior works consider three types of approaches to ensure the inference validity: (i) data-splitting \citep{rinaldo2019bootstrapping}, (ii) conditional selective inference \citep{lee2016exact,tibshirani2016exact}, (iii) simultaneous inference \citep{berk2013valid}. All three approaches achieve the validity \eqref{eq:VIDE} by aiming at a different but stronger coverage guarantee than \eqref{eq:VIDE}. 

Specifically, in the data-splitting strategy, a subset of the data is used to find $\hat{q}$ while the rest is used to construct the confidence interval. Under the independent data assumption, they can then obtain validity by conditioning on the training data that generates $\hat{q}$:
\begin{equation}\label{eq:conditional_validity}
    \lim\inf_{n\rightarrow \infty}\bP(\beta_{\hat{q}}\in \widehat{\mathrm{CI}}_{\hat{q}}|\hat{q} = q)\geq 1-\alpha,
\end{equation}
which implies the coverage guarantee \eqref{eq:VIDE} marginalized over the random $\hat{q}$.
Despite being flexible to the selction process, data-splitting sacrifices data efficiency and is subject to various interpretational challenges, as we also discussed in the main paper.

The conditional selective inference approach, on the other hand, directly characterizes the data subspace that leads to the event $\hat{q}=q$, and then builds the confidence interval or hypothesis test procedure on top of the conditional distribution of the test statistic given $\hat{q}=q$ \citep{tibshirani2016exact}. This approach also leads to the same conditional guarantee as in \eqref{eq:conditional_validity}. One major challenge of this approach lies in the characterization of the conditional distribution; it often requires various assumptions on both the selection algorithm that generates $\hat{q}$ and the data distribution. 

The simultaneous inference approach constructs a confidence interval that is sufficiently large, such that it is valid for all $q\in \mathcal{Q}$, the set of all possible selection events:
\begin{equation}\label{eq:simultaneous_validity}
    \lim\inf_{n\rightarrow \infty}\bP(\forall q\in \mathcal{Q}, \beta_{q}\in \widehat{\mathrm{CI}}_{q})\geq 1-\alpha,
\end{equation}
This approach can be highly conservative when the set $\mathcal{Q}$ of all selection events is huge, and especially when $\hat{q}$ is concentrated in a small subset of $\mathcal{Q}$. 

In summary, the three common strategies discussed above for post-selection inference all construct confidence intervals with stronger coverage guarantees (\eqref{eq:conditional_validity} and \eqref{eq:simultaneous_validity}). However, they either (i) lose data efficiency and suffer from interpretational challenge, or (ii) they only work for certain selection procedure and data distributions, or (iii) they can be highly conservative if the whole space $\mathcal{Q}$ of all possible selection events is huge, with many of the selection events of low probability. These challenges are especially severe for our problem context, where we want to make minimal assumptions on the model training process (i.e., the selection procedure) and data distributions, while maintaining statistical efficiency. Therefore, we choose to take a different route and directly establish the coverage guarantee as in \eqref{eq:VIDE}.

\subsection{Extension to Feature Importance Testing}\label{sec:testing}
Given our LOCO-MP confidence interval for the feature importance score $\Delta_j$, one natural question is whether and how we can convert it to a hypothesis testing procedure, so that we can determine whether feature $j$ is a significant predictor for the current model we trained. The brief answer is yes, but we need to interpret the test result with caution.

In particular, suppose we would like to test whether feature $j$ affects model $\hat{\mu}$'s prediction. Here we follow the notations in Section \ref{sec:selective_inference} and use $\hat{\mu}$ and $\hat{\mu}_{\backslash j}$ to denote the full trained model and the reduced model that excludes feature $j$, instead of using $\mu$ and $\mu_{\backslash j}$ as in the main paper. We may write the null hypothesis as
\begin{equation}\label{eq:null_twosided}
    \mathcal{H}_0: \Delta_j = 0.
\end{equation} 
We focus on a two-sided test here only for convenience of discussion, but all the ideas and conclusions can be directly extended to a one-sided test (e.g., $\cH_0:\Delta_j\leq 0$). 
As we discussed in Section \ref{sec:selective_inference}, the LOCO inference target 
\begin{equation*}
    \Delta_j = \bE[\err(Y,\hat{\mu}_{\backslash j}(X_{\backslash j};\bX_{:,\backslash j}, \bY) - \err(Y,\hat{\mu}(X;\bX, \bY)|\bX,\bY],
\end{equation*}
is a function of the training data $(\bX,\bY)$ instead of being a fixed population quantity in conventional hypothesis testing problems. For a given data distribution and training algorithm, $\Delta_j$ may be zero for some training data $(\bX_1,\bY_1)$ but non-zero for another training data set $(\bX_2,\bY_2)$. This raises the question: {\em How should we define the Type I error when $\mathcal{H}_0$ is also a random event?} Should we condition on $\cH_0$ or should we marginalize over the distribution of $\cH_0$? How do we interpret and use these extended notions of Type I error control?

In fact, when converting our confidence interval to testing \eqref{eq:null_twosided}, we will have asymptotic valid control for an extended notion of the Type I error based on marginalization over the random $\cH_0$\footnote{Similar results can also be shown for the confidence interval with variance barrier or with data-driven tuning, under the corresponding conditions in Theorems \ref{thm:coverage_buffer}, \ref{thm:datadriven_coverage}.}, as shown in the following Proposition.
\begin{prop}\label{prop:test}
   Suppose we reject $\cH_0$ defined in \eqref{eq:null_twosided} if $0\notin \widehat{\mathbb{C}}_j$, where $\widehat{\mathbb{C}}_j$ is given by Algorithm \ref{algo:loco}. Then under the same conditions as in Theorem \ref{thm:coverage_buffer}, we have
   \begin{equation}\label{eq:marginal_TypeI}
       \lim\sup_{N\rightarrow \infty}\bP(\text{reject }\cH_0 \& \cH_0\text{ holds})\leq \alpha.
   \end{equation}
\end{prop}
In the following, we will refer to $\bP(\text{reject }\cH_0 \& \cH_0\text{ holds})$ as the ``marginal Type I error", to emphasize the fact that it is marginalized over the random $\cH_0$. When $\cH_0$ is non-random and holds true, $\bP(\text{reject }\cH_0 \& \cH_0\text{ holds}) = \bP(\text{reject }\cH_0|\cH_0)$, reducing to the conventional type I error. Importantly, Proposition \ref{prop:test} suggests that with probability at least $1-\alpha$, we will only reject $\mathcal{H}_0$ if $\Delta_j\neq 0$ holds (no false rejection). Therefore, we note that this marginal Type I error could be a useful notion as it indeed controls the probability of falsely rejecting a null hypothesis. On the other hand, it also have important difference from the conventional Type I error: consider an extreme case where $\cH_0$ is unlikely to hold, i.e., $\bP(\cH_0\text{ holds})\leq \alpha$, then for an arbitrary test (even if it always reject $\cH_0$), \eqref{eq:marginal_TypeI} is satisfied. As far as we are aware, the notion of ``marginal Type I error" is not well-studied, only being related to the simultaneous inference literature \cite[see Corollary 4.2 in][]{berk2013valid}. We hope our discussion here also opens a gate to future research.

One may also wonder whether and how we can achieve a valid Type I error control through conditioning on $\cH_0$. This is related to the post-selection inference literature, where the selection event introduces randomness to the hypothesis to be tested. As we discussed in Section \ref{sec:selective_inference}, both data-splitting and the conditional selective inference approach obtain the coverage guarantee as in \eqref{eq:conditional_validity}, which conditions on the selection result $\hat{q}=q$. For these two types of approaches, if we directly convert their confidence intervals to hypothesis testing procedures \footnote{To convert it to a testing procedure for the null hypothesis $\cH_0: \beta_{\hat{q}}=0$, we can simply reject $\cH_0$ if $0\notin\widehat{\mathrm{CI}}_{\widehat{q}}$.}, they will satisfy a conditional type I error guarantee:
\begin{equation}\label{eq:conditional_TypeI}
    \lim\sup_{n\rightarrow \infty}\bP(\text{reject }\cH_0| \hat{q}=q, \cH_0\text{ holds})\leq \alpha,
\end{equation}
where $\mathcal{H}_0$ is no longer a random event conditioning on the selection $\hat{q}=q$. This Type I error control through conditioning is stronger than and implies \eqref{eq:marginal_TypeI}, while it requires characterizing the conditional distribution of the test statistic given the selection procedure, a challenging task that often requires assumptions on the selection procedure and data distributions. It is of future interest to investigate how one can relax these assumptions and extend the conditional post-selection inference ideas to the LOCO importance framework with minimal assumptions.

\begin{proof}[Proof of Proposition \ref{prop:test}]
    By the definition of the test in Proposition \ref{prop:test}, we note that the event that $\cH_0$ is rejected and $\cH_0$ holds immediately implies that $\Delta_j\notin \hat{\mathbb{C}}_j$. The conclusion follows directly by applying Theorem \ref{cor:coverage-width}.
\end{proof}
\subsection{Distribution-free Predictive Inference}\label{sec:J+MP}
	As can be seen from Algorithm \ref{algo:loco}, the minipatch learning framework makes the computation of leave-one-out and leave-one-covariate-out predictions nearly free given the fitted models from small minipatches. Inspired by \cite{kim2020predictive} which provides fast and distribution-free predictive inference using Jackknife+ \citep{barber2021predictive} with bootstrap, we propose a novel Jackknife+ Minipatch conformal inference procedure (J+MP) that can additionally take advantage of our fitted leave-one-out (LOO) predictors to construct predictive confidence intervals, which also comes for free and can be obtained simultaneously as the feature importance interval. As far as we are concerned, this is the first inference procedure that can perform feature importance inference and predictive inference at the same time.
	
	Specifically, given the LOO predictions computed in step 2 of Algorithm \ref{algo:loco}, we can further compute the non-conformity score for each data point $i$: $R^{(\LOO)}_{i}=\mathrm{Error}(Y_{i}, \mu_{-i}(X_i))$; Then for any new data point with feature $X_{N+1}$, we obtain its ensembled LOO prediction:
	$\mu_{-i}(X_{N+1}) = \frac{1}{\sum_{k=1}^{K} \mathbb{I}(i \notin I_k)} \sum_{k=1}^{K} \mathbb{I}(i \notin I_{k}) \mu_{k} (X_{N+1}),\quad i=1,\dots,N.$
	For a given confidence level $1-\alpha$, following the construction in \cite{barber2021predictive,kim2020predictive}, our Jackknife+ MP confidence set/interval for $Y_{N+1}$ is defined as follows:
	\begin{enumerate}
		\item In the regression setting, we focus on the absolute error loss, i.e., $\err(Y,\hat{Y})=|Y-\hat{Y}|$, and our confidence interval is
		\begin{equation}\label{eq:pred_interval_regress}
		\hat{C}_{\alpha}^{\mathrm{J+MP}}(X_{N+1})= [\hat{q}_{N,\alpha}^{-}\{\mu_{-i}(X_{N+1})-{R^{(LOO)}_{i} } \}, \hat{q}_{N,\alpha}^{+}\{\mu_{-i}(X_{N+1})+{R^{(LOO)}_{i}}\}],
		\end{equation}
		where we followed the notation in \cite{barber2021predictive}: for any values $\{v_i\}_{i=1}^N$, $\hat{q}_{N,\alpha}^{+}\{v_i\}$ is the $\lceil(1-\alpha)(N+1)\rceil$th smallest value in $\{v_i\}_{i=1}^N$, and $\hat{q}_{N,\alpha}^{-}\{v_i\}$ is the $\lfloor\alpha(N+1)\rfloor$th smallest value in $\{v_i\}_{i=1}^N$. 
		\item In the classification setting, 
		\begin{equation}\label{eq:pred_interval_class}
		\hat{C}_{\alpha}^{\mathrm{J+MP}}(X_{N+1}) =\left\{Y:\sum_{i=1}^{N}\mathbb{I}\left(\err(Y,\mu_{-i}(X_{N+1}))\geq R^{(LOO)}_i\right)\leq (1-\alpha)(N+1)\right\}.
		\end{equation}
	\end{enumerate}
	The full procedure for constructing predictive intervals is summarized in Algorithm~\ref{algo:j+mp}.
	
	\begin{algorithm}[H]\label{algo:j+mp}~
		\SetAlgoLined
		\footnotesize
		\DontPrintSemicolon
		
		\SetKwInOut{Input}{input}\SetKwInOut{Output}{output}
		
		\noindent{\textbf{Input}}: Training pairs ($\boldsymbol{X},\boldsymbol{Y}$), test point $X_{N+1}$, minipatch sizes $n$, $m$; number of minipatches $K$, base learner $H$, target confidence level $1-\alpha$;
		\begin{enumerate}
			\item Perform Minipatch Learning: For $k=1,...,K$:
			\begin{enumerate}
				\item Randomly subsample $n$ observations, $I_k \subset [N]$, and $m$ features, $F_k \subset [M]$.
				\item Train prediction model $\mu_k$ on $(\boldsymbol{X}_{I_k,F_k}, \boldsymbol{Y}_{I_k})$: for any $X\in \bR^M$, $\mu_k(X) = H(\boldsymbol{X}_{I_k,F_k},\boldsymbol{Y}_{I_k})(X_{F_k})$.
			\end{enumerate}
			\item Obtain LOO predictions : 
			\begin{enumerate}
				\item  Obtain the ensembled LOO prediction for $i = 1,...N$:\\ $\mu_{-i}(X_i) = \frac{1}{\sum_{k=1}^{K} \mathbb{I}(i \notin I_k)} \sum_{k=1}^{K} \mathbb{I}(i \notin I_{k}) \mu_{k} (X_{i})$;\\
				and ensembled LOO nonconformity scores: \\
				$R^{(LOO)}_{i} $ = $\mathrm{Error}(Y_{i}, \mu_{-i}(X_i))$; 
				\item  Obtain the ensembled LOO prediction for new observation $N+1$:\\ 
				$\mu_{-i}(X_{N+1}) = \frac{1}{\sum_{k=1}^{K} \mathbb{I}(i \notin I_k)} \sum_{k=1}^{K} \mathbb{I}(i \notin I_{k}) \mu_{k} (X_{N+1})$;
			\end{enumerate}
			
			\item Calculate Minipatch conformal interval $\hat{C}^{\mathrm{J+MP}}$ as in \eqref{eq:pred_interval_regress} for regression or \eqref{eq:pred_interval_class} for classification.
		\end{enumerate}
		
		\textbf{Output}: $ \hat{C}^{\mathrm{J+MP}}_{\alpha}(X_{N+1})$
		\caption{J+MP Minipatch Predictive Interval}
	\end{algorithm}

	Similar to \cite{kim2020predictive}, we guarantee the coverage of $\hat{C}^{\mathrm{J+MP}}$ with no distributional assumptions. 
	\begin{assumption}\label{assump:data_exchange}
		$(X_1,Y_1),...,(X_{N+1},Y_{N+1})$ are exchangeable. Formally, for any permutation $\sigma$ on $\{1,\dots,N+1\}$, $(X_1,Y_1,\dots,X_{N+1},Y_{N+1})\overset{\rm d.}{=}(X_{\sigma(1)},Y_{\sigma(1)},\dots,X_{\sigma(N+1)},Y_{\sigma(N+1)})$
	\end{assumption}
	\begin{assumption}\label{assump:alg_orderinvariant}
		The base prediction algorithm $H$ is invariant to the order of input. For $N>1$, and fixed N-tuple $((X_1,Y_1),...,(X_N,Y_N))$, and any permutation $\sigma$ on $\{1,...,N\}$:
		$H((X_1,y_1),...,(x_N,y_N)) = H((X_{\sigma(1)},Y_{\sigma(1)}),...,(X_{\sigma(N)},Y_{\sigma(N)}))$
	\end{assumption}
	\begin{assumption}\label{assump:binom_minipatch}
		There exists $\tilde{K}>0$, such that the number of minipatches in Algorithm 1 is generated from a Binomial distribution $K\sim (\tilde{K},1-\frac{n}{N+1})$.
	\end{assumption}
	These assumptions are standard in the literature of conformal inference \citep{barber2021predictive,kim2020predictive}. In particular, Assumption~\ref{assump:binom_minipatch} follows the Binomial assumption from Theorem 1 in \cite{kim2020predictive}: by generating the number of minipatches at random, it allows symmetrical treatment of samples in our proof.
	\begin{theorem}[Distribution-free Predictive Inference Guarantee]\label{thm:pred_coverage}
		Under Assumptions \ref{assump:data_exchange}-\ref{assump:binom_minipatch}, the Jackknife+ MP prediction interval satisfies 
		$\mathbb{P}\{Y_{N+1}\in \hat{C}_{\alpha}^{\mathrm{J+MP}} (X_{N+1})\}\ge 1-2\alpha.$
	\end{theorem}
	Our proof, included in Section~\ref{sec:proof_pred}, closely follows the proofs in \cite{barber2021predictive} and \cite{kim2020predictive}, while the main difference lies that we also show that features subsampling does not affect the exchangeability among samples.

\section{Additional Theoretical Results and Proofs}\label{sec:proof}
In this section, we present the proofs of all our theoretical statements on the valid coverage of our feature importance confidence intervals (Sections~\ref{sec:proof_main}-\ref{sec:proof_buffer}) and distribution-free predictive intervals (Section~\ref{sec:proof_pred}). We will also present some additional theoretical results we mentioned in the main paper together with their proofs in Sections~\ref{sec:theory_datadriven_supp}-\ref{sec:thm_disc_target} and Sections~\ref{sec:proof_datadriven}, \ref{sec:disc_proof}.

Before presenting the proofs, we first define and recall some notations and auxiliary functions that would be useful. 

\paragraph{Notations:} For any random variables $X$ and $Y$ where $X$ implicitly depends on the sample size $N$, we write $X\overset{p}{\rightarrow}Y$ if for any $\epsilon>0$, $\lim_{N\rightarrow \infty}\bP(|X(N)-Y|>\epsilon)=0$; $X\overset{L_1}{\rightarrow}Y$ if $\lim_{N\rightarrow \infty}\bE|X(N)-Y|=0$; 
 $X\overset{d}{\rightarrow}Y$ if for any $t\in \bR$, $\lim_{N\rightarrow \infty}\bP(X(N)\leq t)=\bP(Y\leq t)$. For any two random variables $X, Y$, we write $X=o_p(Y)$ if $\frac{X}{Y}\overset{p}\rightarrow 0$. For any two scalars $a,b\in \bR$ that may implicitly depend on sample size $N$, we write $a=o(b)$, or $b\gg a$, if $\lim_{N\rightarrow \infty}\frac{a}{b}=0$. 
 For any interval $[a,b]\subset \bR$ with $a\leq b$ we use $|[a,b]|=b-a$ to denote its length. For two random variables $X, Y$, we write $X\overset{\rm d.}{=}Y$ if the distribution of $X$ is the same as the distribution of $Y$. We use $[N]$ to represent the set $\{1,\dots,N\}$. For any $n, i$, we denote the $i$th canonical vector in $\bR^n$ by $e^n_i$, and we will omit the superscript when the dimension is clear from the context. 
	
	Let $h_j^{(K)}(X,Y;\boldsymbol{X},\boldsymbol{Y})=\err(Y,\mu_{\backslash j}(X_{\backslash j};\boldsymbol{X}_{:,\backslash j},\boldsymbol{Y}))-\err(Y,\mu(X;\boldsymbol{X},\boldsymbol{Y}))$
be the importance of the characteristic of the characteristic $j$ evaluated in the training data set $(\boldsymbol{X},\boldsymbol{Y})$ and test data point $(X,Y)$. Recall our definition of the inference target $\Delta_j$ in the main paper:
 \begin{equation}\label{eq:target_inference}
	\Delta_j(\boldsymbol{X},\boldsymbol{Y}) = \bE_{(X,Y)\sim \mathcal{P}}\left\{\err(Y,\mu_{\backslash j}(X_{\backslash j};\boldsymbol{X}_{:,\backslash j},\boldsymbol{Y})) - \err(Y,\mu(X;\boldsymbol{X},\boldsymbol{Y}))|\boldsymbol{X},\boldsymbol{Y}\right\}.
	\end{equation}
    One can see that $\Delta_j=\bE_{X,Y}(h_j^{(K)}(X,Y;\boldsymbol{X},\boldsymbol{Y})|\boldsymbol{X},\boldsymbol{Y})$
	is the expectation of $h_j^{(K)}(X,Y;\boldsymbol{X},\boldsymbol{Y})$ taken over the test data point. In addition, recall that 
   $h_j(X,Y;\boldsymbol{X},\boldsymbol{Y})=\err(Y,\mu^*_{\backslash j}(X_{\backslash j};\boldsymbol{X}_{:,\backslash j},\boldsymbol{Y}))-\err(Y,\mu^*(X;\boldsymbol{X},\boldsymbol{Y}))$,
 where the minipatch predictor $\mu^*$ and $\mu_{\backslash j}^*$ satisfy the following:
	\begin{align}\label{eq:minipatch_predictor}
	\mu^*(X;\boldsymbol{X},\boldsymbol{Y}) = &\frac{1}{\binom{N}{n}\binom{M}{m}}\sum_{\substack{I\subset [N], |I|=n \\ F\subset [M], |F|=m}}H(\boldsymbol{X}_{I,F},\boldsymbol{Y}_{I})(X_{F}),\\
        \mu^*_{\backslash j}(X_{\backslash j};\boldsymbol{X}_{:,\backslash j},\boldsymbol{Y})) = &\frac{1}{\binom{N}{n}\binom{M-1}{m}}\sum_{\substack{I\subset [N], |I|=n \\ F\subset [M]\backslash j, |F|=m}}H(\boldsymbol{X}_{I,F},\boldsymbol{Y}_{I})(X_{F}).
	\end{align}
 
We will see in our coverage guarantee that $\bar{\Delta}_j-\Delta_j$ will have asymptotic variance depending on the following function:  
	\begin{equation}\label{eq:h_j_abbrv_def}
	h_j(X,Y)=\mathbb{E}_{\boldsymbol{X},\boldsymbol{Y}}[h_j(X,Y;\boldsymbol{X},\boldsymbol{Y})],
	\end{equation} 
	the expectation of $h_j(X,Y;\boldsymbol{X},\boldsymbol{Y})$ over the training data set. Also, we define
$\hat{h}_j(X_i,Y_i;\boldsymbol{X}_{\backslash i,:},\boldsymbol{Y}_{\backslash i})=\hat{\Delta}_j(X_i,Y_i)$, the LOO feature occlusion score calculated in Algorithm 1; $\tilde{h}_j(X_i,Y_i;\boldsymbol{X}_{\backslash i,:},\boldsymbol{Y}_{\backslash i})= h_j(X_i,Y_i;\boldsymbol{X}_{\backslash i,:},\boldsymbol{Y}_{\backslash i})- h_j(X_i,Y_i)$; $\mu_{I,F}(X)=(H(\bX_{I,F},\bY_{I_k}))(X_{F})\in \mathbb{R}^d$, the prediction of the base learner trained on $(\bX_{I,F},\bY_{I})$. For convenience, in the following proofs, we denote $\mu^*(X_{\backslash j};\bX_{\backslash i,\backslash j},\bY_{\del i})$, $\mu^*(X;\bX_{\backslash i,:},\bY_{\del i})$, $\mu^*(X_{\backslash j};\bX_{:,\backslash j},\bY)$, and $\mu^*(X;\bX,\bY)$ by $\mu^{*-j}_{-i}(X)$, $\mu^*_{-i}(X)$, $\mu^{*-j}(X)$, and $\mu^*(X)$, respectively.
We also state extended versions of the theoretical results in the main paper here.
	\begin{theorem}[Asymptotic distribution of $\bar{\Delta}_j$]\label{thm:normal_approx}
	   Suppose that all training data $(X_i, Y_i)\overset{i.i.d}{\sim}\mathcal{P}$ and Assumptions \ref{assump:Lip}-\ref{assump:third_moment} hold. Then we have
		\begin{equation*}
		\sqrt{N}\sigma_j^{-1}(\bar{\Delta}_j-\Delta_j)\overset{d}{\rightarrow}\mathcal{N}(0,1),
		\end{equation*}
		where $\sigma_j^2 = \mathrm{Var}_{(X,Y)\sim \cP}(h_j(X,Y))$ with $h_j(\cdot,\cdot)$ being defined in \eqref{eq:h_j_abbrv_def}.
	\end{theorem}
    \begin{theorem}[Consistent variance estimate]\label{thm:var_est}
		Consider the sample variance $\hat{\sigma}_j^2$ defined in Algorithm 1. Under the same assumptions as in Theorem \ref{thm:normal_approx}, we have $\frac{\hat{\sigma}_j^2}{\sigma_j^2}\overset{p}{\rightarrow}1$.
	\end{theorem}
	\vspace{-1mm}
	\noindent By combining Theorems \ref{thm:normal_approx} and \ref{thm:var_est}, we immediately have the asymptotic correct coverage of our confidence interval $\hat{\bC}_j$ for $\Delta_j$ as stated in Theorem \ref{cor:coverage-width} in the main paper. The width of the confidence interval scales as $\sigma_j/\sqrt{N}$, 

\begin{table}[!htb]
    \centering
    \begin{tabular}{c|c}
    \hline
        $\mu_k(X)$, $\mu_{I_k,F_k}(X)$ & $H(\bX_{I_k,F_k},\bY_{I_k})(X_{F_k})$\\        $\mu(X;\bX,\bY)$ &  $\frac{1}{K}\sum_{k=1}^K\mu_k(X)$\\
        $\mu_{\backslash j}(X_{\backslash j};\bX_{:,\backslash j},\bY)$ & $\frac{1}{K}\sum_{k=1}^K\mu_{\tilde{I}_k,\tilde{F}_k}(X)$\\
        $\mu_{-i}(X)$ & $\frac{1}{\sum_{k=1}^K \ind(i\notin I_k)}\sum_{k=1}^K\ind(i\notin I_k)\mu_k(X)$\\
        $\mu_{-i}^{-j}(X)$ & $\frac{1}{\sum_{k=1}^K \ind(i\notin I_k)\ind(j\notin F_k)}\sum_{k=1}^K\ind(i\notin I_k)\ind(j\notin F_k)\mu_k(X)$\\
        $\mu^*(X)$ & $\frac{1}{\binom{N}{n}\binom{M}{m}}\sum_{\substack{I\subset[N], |I|=n\\F\subset[M], |F|=m}}\mu_{I,F}(X)$\\
        $\mu^*_{\backslash j}(X)$, $\mu^{*-j}(X)$ & $\frac{1}{\binom{N}{n}\binom{M-1}{m}}\sum_{\substack{I\subset[N], |I|=n\\F\subset[M], |F|=m}}\ind(j\notin F)\mu_{I,F}(X)$\\
        $\mu^*_{-i}(X)$ & $\frac{1}{\binom{N-1}{n}\binom{M}{m}}\sum_{\substack{I\subset[N], |I|=n\\F\subset[M], |F|=m}}\ind(i\notin I)\mu_{I,F}(X)$\\
        $\mu^{*-j}_{-i}(X)$ & $\frac{1}{\binom{N-1}{n}\binom{M-1}{m}}\sum_{\substack{I\subset[N], |I|=n\\F\subset[M], |F|=m}}\ind(i\notin I)\ind(j\notin F)\mu_{I,F}(X)$\\
        $\tilde{\mu}_{-i}(X)$ & $\frac{N}{N-n}\frac{1}{K}\sum_{k=1}^K\ind(i\notin I_k)\mu_{I_k,F_k}(X)$\\
        $\tilde{\mu}_{-i}^{-j}(X)$ & $\frac{N}{N-n}\frac{M}{M-m}\frac{1}{K}\sum_{k=1}^K\ind(i\notin I_k)\ind(j\notin F_k)\mu_{I_k,F_k}(X)$\\
    \hline
    \end{tabular}
    \caption{List of notations for predictors. In the third row, $(\tilde{I_k},\tilde{F_k})$'s are i.i.d uniformly sampled indices from $[N]$ and $[M]\backslash j$, with size $n,\,m$. When the minipatch size is unclear from the context, we add superscript $(m,n)$ to the corresponding predictors.}
    \label{tab:notations_mu}
\end{table}

 \subsection{Theoretical Details for Section 3.3}\label{sec:theory_datadriven_supp}
 \paragraph{Additional notations:} To establish theory for the data-driven tuning version of LOCO-MP, we need to define some additional notations and redefine some previous notations. Since different minipatch sizes are being considered in this context, we will add superscript $(m,\,n)$ to certain quantities that depend on minipatch sizes. In particular, we let 
 \begin{equation}\label{eq:mu_mn}
     \mu^{*(m,n)}(X;\bX,\bY) = \frac{1}{\binom{N}{n}\binom{M}{m}}\sum_{\substack{I\subset [N], |I|=n \\ F\subset [M], |F|=m}}H(\boldsymbol{X}_{I,F},\boldsymbol{Y}_{I})(X_{F})
 \end{equation} 
 be the minipatch predictor with infinite sampling with size $(m,\,n)$. Also let $$h_j^{(m,n)}(X,Y;\boldsymbol{X},\boldsymbol{Y})=\err(Y,\mu^{*(m,n)}_{\backslash j}(X_{\backslash j};\boldsymbol{X}_{:,\backslash j},\boldsymbol{Y}))-\err(Y,\mu^{*(m,n)}(X;\boldsymbol{X},\boldsymbol{Y})),$$ $h_j^{(m,n)}(X,Y) = \bE_{\bX,\bY}[h_j^{(m,n)}(X,Y;\bX,\bY)]$, $\tilde{h}_j^{(m,n)}(X_i,Y_i;\boldsymbol{X}_{\backslash i,:},\boldsymbol{Y}_{\backslash i})= h_j^{(m,n)}(X_i,Y_i;\boldsymbol{X}_{\backslash i,:},\boldsymbol{Y}_{\backslash i})- h_j^{(m,n)}(X_i,Y_i)$. 

\subsection{LOCO-MP with Data-driven Selection of Minipatch Sizes: Valid Coverage with Variance Barrier}
Here, we present the detailed method and theory for LOCO-MP with data-driven selection of minipatch sizes and the variance barrier. In particular, we still consider the confidence interval in \eqref{eq:CI_barrier}, but with a different choice of $\epsilon(N)$.
\begin{assumption}[Minipatch size and number with variance barrier]\label{assump:buffer_datadriven}
    All candidate minipatch sizes satisfy $\frac{n_l}{N},\,\frac{m_l}{M}\leq \gamma$ for some constant $0<\gamma<1$, $$\epsilon(N)\geq c\frac{L\log N}{N}\sqrt{\sum_{l=1}^s n_l^2\stb(m_l,n_l)},$$
    for some constant $c>0$. In addition, $K\gg \big(\frac{L^2B^2}{\epsilon^2(N)}+1\big)\log N$.
\end{assumption}
 
 In addition, we define the oracle minipatch sizes $(m^{\mathrm{oracle}},\,n^{\mathrm{oracle}})$ as the minimizer for the population LOO residual:
 \begin{definition}\label{def:mn_oracle}
     \begin{equation*}
        (m^{\mathrm{oracle}},\,n^{\mathrm{oracle}}) = \argmin_{(m,\,n)\in S}\bE_{\bX,\bY}\LOO(m,\,n),
    \end{equation*}
    where $\LOO(m,\,n) = \frac{1}{N}\sum_i \err(Y_i,\,\mu^{*(m,n)}(X_i;\bX_{\backslash i,:},\bY_{\backslash i})$ is the leave-one-out residual on the training data $(\bX,\,\bY)$ when the minipatch sizes $(m,\,n)$ are in use; the expectation here is taken over the training data. 
 \end{definition}
 Let $(\hat{m},\,\hat{n})$ be the selected minipatch size pair by Algorithm \ref{algo:MPsize_tuning} based on random sampling of minipatches. Let $(m^*,\,n^*)$ be the best minipatch sizes that if one has access to the combinatorial average of all minipatches, defined formally in Definition \ref{def:delta_LOO}.
 \begin{definition}[Performance gap of sub-optimal minipatch sizes]\label{def:delta_LOO}
    Define the leave-one-observation-out residual for training data $(\bX,\bY)$ and a given minipatch size pair $(m,n)$ as $\LOO(m,n) = \frac{1}{N}\sum_i\err(Y_i,\mu^{*(m,n)}(X_i;\bX_{\backslash i,:}, \bY_{\backslash i}))$, where $\mu^{*(m,n)}(\cdot;\bX_{\backslash i,:}, \bY_{\backslash i})$ is as defined in \eqref{eq:mu_mn}, except that the training data $(\bX_{\backslash i,:}, \bY_{\backslash i})$ excludes sample $i$. Let $(m^*,n^*)=\argmin_{(m,n) \in S}\LOO(m,n)$, and $$\delta_{\LOO}(S) = \min_{(m,n)\neq (m^*,n^*)}\LOO(m,n) - \LOO(m^*,n^*)$$ be the performance gap between the best and the second best minipatch sizes.
\end{definition}
 Similarly, $m^*_{-i},\,n^*_{-i}$ are defined as the best minipatch sizes on training data $(\bX_{-i,:},\,\bY_{-i})$ which excludes sample $i$:
$$
(m^*_{-i},\,n^*_{-i}) = \argmin_{(m,n) \in S}\LOO_{-i}(m,\,n),\quad \LOO_{-i}(m,n) = \frac{1}{N}\sum_{l\neq i}\err(Y_l,\mu^{*(m,n)}(X_l;\bX_{\backslash \{l,i\},:}, \bY_{\backslash \{l,i\}})).
$$
Here, $m^*,\,n^*$ are functions of $(\bX,\bY)$, and hence we can also write them as $m^*(\bX,\bY)$, $n^*(\bX,\bY)$; $m^*_{-i},\,n^*_{-i}$ can also be written as $m^*(\bX_{\backslash i,:},\bY_{\backslash i}),  n^*(\bX_{\backslash i,:},\bY_{\backslash i})$. We use the abbreviated version $m^*,\,n^*$ only when the training data is the full data set $(\bX,\bY)$ we have at hand. We then let 
\begin{equation}\label{eq:h_j_full_datadriven}
    h_j(X,Y;\bX,\bY) = h_j^{m^*(\bX,\bY),n^*(\bX,\bY)}(X,Y;\bX,\bY)
\end{equation} 
be $j$'s feature importance score when the minipatch ensemble is both trained and tuned using $(\bX,\bY)$ and tested on $(X,Y)$. Similarly, $$h_j(X,Y;\bX_{\backslash i,:},\bY_{\backslash i,:}) = h_j^{m^*(\bX_{\backslash i,:},\bY_{\backslash i,:}),n^*(\bX_{\backslash i,:},\bY_{\backslash i,:})}(X,Y;\bX_{\backslash i,:},\bY_{\backslash i,:}) = h_j^{m^*_{-i},n^*_{-i}}(X,Y;\bX_{\backslash i,:},\bY_{\backslash i,:}).$$ 

\begin{definition}[Sensitivity of feature importance w.r.t. minipatch sizes]\label{def:delta_LOCO}
 Let $$\delta_{\LOCO}(j, S; X,Y,\bX,\bY)=\max_{1\leq l,l'\leq s}|h_j^{(m_l,n_l)}(X,Y;\bX,\bY) - h_j^{(m_{l'},n_{l'})}(X,Y;\bX,\bY)|$$ be the maximum difference in $j$'s feature importance when considering two different pairs of minipatch sizes, indicating the sensitivity of feature importance w.r.t. minipatch sizes. We then define the quantity $\delta_{\LOCO}^2(j,S)$ as the upper bound for the following three average notions of the feature importance sensitivity:
    \begin{equation*}
        \begin{split}
            \delta_{\LOCO}(j,S) = \max\{\delta_{\LOCO}^{(1)}(j,S),\,\delta_{\LOCO}^{(2)}(j,S),\,\delta_{\LOCO}^{(3)}(j,S),\,\delta_{\LOCO}^{(4)}(j,S)\}\\
            \delta_{\LOCO}^{(1)}(j,S) = \sqrt{\frac{1}{N}\sum_{i=1}^N\delta^2_{\LOCO}(j, S; X_i,Y_i, \bX_{\backslash i,:},\bY_{\backslash i})},\\
            \delta_{\LOCO}^{(2)}(j,S) = \sqrt{\frac{1}{N}\sum_{i=1}^N\bE_{X_i,Y_i}\left[\delta^2_{\LOCO}(j, S; X_i,Y_i, \bX_{\backslash i,:},\bY_{\backslash i})|\bX_{\backslash i,:},\bY_{\backslash i}\right]},\\
            \delta_{\LOCO}^{(3)}(j,S) = \sqrt{\bE\left[\delta^2_{\LOCO}(j, S; X_i,Y_i, \bX_{\backslash i,:},\bY_{\backslash i})\right]},\\
            \delta_{\LOCO}^{(4)}(j,S) = \left\{\bE\left[\delta^4_{\LOCO}(j, S; X_i,Y_i, \bX_{\backslash i,:},\bY_{\backslash i})\right]\right\}^{1/4}.
        \end{split}
    \end{equation*}
\end{definition}
\begin{definition}[Variance of LOO residuals]\label{def:sigma_LOO}
Define the maximum variance of the LOO residual across different minipatch sizes as:
    \begin{equation*}
        \sigma_{\LOO}^2(S)=\max_{(m,n)\in S}\mathrm{Var}[\err(Y_i,\mu^{*(m,n)}(X_i;\bX_{\backslash i,:},\bY_{\backslash i})].
    \end{equation*}
Also let the maximum kurtosis be defined as 
    \begin{equation*}
        \kappa_{\LOO}(S)=\max_{(m,n)\in S}\frac{\bE[\err(Y_i,\mu^{*(m,n)}(X_i;\bX_{\backslash i,:},\bY_{\backslash i}) - \bE(\err(Y_i,\mu^{*(m,n)}(X_i;\bX_{\backslash i,:},\bY_{\backslash i}))]^4}{\mathrm{Var}[\err(Y_i,\mu^{*(m,n)}(X_i;\bX_{\backslash i,:},\bY_{\backslash i})]}.
    \end{equation*}
\end{definition}
In the following, we present two regularity assumptions on the quantities we just defined.
\begin{assumption}[Sufficient sub-optimality gap in LOO residuals]\label{assump:delta_LOO_bnd}
    We assume the sub-optimality gap $\delta_{\LOO}(S)$ in the average LOO residuals is lower bounded by a constant proportion of the highest average LOO residual:
    $$\delta_{\LOO}(S)\geq c\max_{(m,n)\in S}\LOO(m,n)\geq c'LB,$$
    where $L$ and $B$ are defined in Assumptions \ref{assump:Lip}-\ref{assump:bnd_mu}.
    In addition, we assume the average LOO residuals are lower bounded by a constant factor of their standard deviations: for all $(m,n)\in S$,
    $$\LOO(m,n) = \frac{1}{N}\sum_i\err(Y_i,\mu^{*(m,n)}(X_i;\bX_{\backslash i,:}, \bY_{\backslash i})) \geq c\sqrt{\mathrm{Var}[\err(Y_i,\mu^{*(m,n)}(X_i;\bX_{\backslash i,:},\bY_{\backslash i})]}.$$
    Furthermore, the maximum kurtosis for the LOO residual is bounded: $\kappa_{\LOO}(S)\leq C$.
\end{assumption}
Assumption \ref{assump:delta_LOO_bnd} immediately implies $\sigma_{\LOO}(S)\leq C\delta_{\LOO}(S)$.
\begin{assumption}[Moment bound for feature importance score with data-driven selected minipatch sizes]\label{assump:datadriven_h_third_moment}
    Assumption \ref{assump:third_moment} still holds when $h_j(X,\,Y) = \bE_{\bX,\bY}[h_j^{(m^{\mathrm{oracle}},n^{\mathrm{oracle}})}(X,Y;\bX,\bY)]$ is substituted by a slightly different feature importance function: $$h_j'(X,\,Y) = \bE_{\bX,\bY}[h_j^{(m*(\bX,\bY),n^*(\bX,\bY)}(X,Y;\bX,\bY)].$$ In addition, the variance of $h_j'(X,\,Y)$ is not too much larger than the variance of $h_j(X,\,Y)$: $\mathrm{Var}(h_j'(X,\,Y))\leq C\mathrm{Var}(h_j(X,\,Y))$, where the latter was defined as $\sigma_j^2$ in Section \ref{sec:datadriven_theory}.
\end{assumption}
\begin{assumption}[Bounded sensitivity of feature importance score]\label{assump:delta_LOCO_bnd}
We assume the sensitivity of the feature importance score  w.r.t. the minipatch size is upper bounded by a constant factor of its variance:
    $\delta_{\LOCO}^2(j,S)\leq C\mathrm{Var}(h_j'(X,\,Y))$, where $h_j'(X,\,Y)$ is as defined in Assumption \ref{assump:datadriven_h_third_moment}.
\end{assumption}
Assumption \ref{assump:delta_LOCO_bnd} assumes that the variance of the feature importance score is not negligible compared to the sensitivity parameter $\delta_{\LOCO}^2(j,S)$. This is a mild assumption: although $\mathrm{Var}(h_j'(X,\,Y))$ can be close to zero when the feature $j$ is a noise feature which is not helpful for the prediction task, $h_j^{(m,n)}(X,Y;\bX,\bY)$ is also likely small in this case for different minipatch sizes, leading to small $\delta_{\LOCO}(j,S)$. 
\begin{assumption}\label{assump:stb_kurtosis}
    With the same notations as in Definition \ref{def:stb}, define the fourth-order stability as
    \begin{equation*}
        \stb^{(4)}(m,n;H,\cP) = \frac{1}{\binom{M}{m}}\sum_{F\subset[M],|F|=m}\bE\|\mu_F(X_0) - \mu_F'(X_0)\|_2^4.
    \end{equation*}
    We assume that the fourth-order stability is not much larger than the squared of the original second-order stability: $\stb^{(4)}(m,n;H,\cP)\leq C\stb^2(m,n;H,\cP)$.
\end{assumption}
\begin{theorem}[Coverage with data-driven MP sizes and variance barrier]\label{thm:datadriven_coverage_barrier}
    Suppose Assumptions \ref{assump:Lip}, \ref{assump:buffer_datadriven}, \ref{assump:delta_LOO_bnd}-\ref{assump:stb_kurtosis} hold, the number of candidate minipatch sizes $s$ is bounded, and Assumption \ref{assump:bnd_mu} holds for all candidate minipatch sizes $\{(m_l,\,n_l)\}_{l=1}^s$. Then given the data-driven selection of the minipatch sizes in Algorithm \ref{algo:MPsize_tuning}, the confidence interval \eqref{eq:CI_barrier} has asymptotically valid coverage: $\lim\inf_{N\rightarrow\infty}\bP(\Delta_j\in \hat{\mathbb{C}}_j^{\mathrm{barrier}})\geq 1-\alpha$.
\end{theorem}
When the variance barrier is appropriately chosen as in Assumption \ref{assump:buffer_datadriven}, Theorem \ref{thm:datadriven_coverage_barrier} guarantees the asymptotically valid coverage of $\hat{\mathbb{C}}_j^{\mathrm{barrier}}$, without Assumption \ref{assump:mpsize} and \ref{assump:mpnumber}. 
\subsection{Detailed Theoretical Results and Expanded Discussion for Section 2}\label{sec:thm_disc_target}
In this section, we present the full details of the theoretical results we presented or mentioned in Section 2 of the main paper. The proofs are included in Section \ref{sec:disc_proof}. 
\subsubsection{Special Example: the Linear Model}\label{sec:proof_disc_target}
First, we present the characterization of target under the linear model with independent features, the setting described in Section 2.2 in the main paper. Specifically, we assume that all data points $(X_i,y_i)$ are i.i.d. samples of a linear model: $y_i = X_i^\top\beta^*+\epsilon_i$, where $\beta^*\in \mathbb{R}^M$ is the linear regression parameter, and $\{\epsilon_i\}_{i=1}^N$ are independent sub-Gaussian noise of mean zero, variance $\sigma_{\epsilon}^2$, and sub-Gaussian parameter bounded by $C\sigma_{\epsilon}^2$. Also assume that the least squares estimator is our base learner for each minipatch, and the squared error $\err(Y,\hat{Y})=(Y-\hat{Y})^2$ is in use. We also assume independent features for now: $X_i\sim\mathcal{N}(0,I_p)$. We first define some key technical quantities that are useful for our theory. For a given minipatch $(I,F)$, let $\hat{\Theta}_{I,F} = \left(\frac{1}{n}\boldsymbol{X}_{I,F}^\top\boldsymbol{X}_{I,F}\right)^{-1}$ be the corresponding sample precision matrix. When the minipatch sizes are appropriately chosen ($n>m$), we may assume $\frac{1}{n}\boldsymbol{X}_{I,F}^\top\boldsymbol{X}_{I,F}$ to be of full-rank and hence $\hat{\Theta}_{I,F}$ is well defined. Let
\begin{align*}
    \lambda_{n,m}(\boldsymbol{X})=&\frac{1}{\binom{N}{n}\binom{M}{m}}\sum_{I\subset[N],|I|=n}\sum_{F\subset[M],|F|=m}\lambda_{\max}\left(\hat{\Theta}_{I,F}\right)
\end{align*} be the average maximum eigenvalue of the minipatch precision matrices, and 
\begin{align*}
    \lambda_{n,m}(\boldsymbol{X}_{:,\backslash j})=&\frac{1}{\binom{N}{n}\binom{M-1}{m}}\sum_{I\subset[N],|I|=n}\sum_{j\notin F,|F|=m}\lambda_{\max}\left(\hat{\Theta}_{I,F}\right),\\
    \lambda_{n,m}^{(j)}(\boldsymbol{X})=&\frac{1}{\binom{N}{n}\binom{M-1}{m-1}}\sum_{I\subset[N],|I|=n}\sum_{j\in F,|F|=m}\lambda_{\max}\left(\hat{\Theta}_{I,F}\right),
\end{align*} 
be the ones excluding feature $j$ or including feature $j$. Also define $\overline{\lambda}_{m,n}$ as the maximum over these three average eigenvalue quantities: $\overline{\lambda}_{m,n}=\max\{\lambda_{n,m}(\boldsymbol{X}), \lambda_{n,m}(\boldsymbol{X}_{:,\backslash j}), \lambda_{n,m}^{(j)}(\boldsymbol{X})\}$. Based on existing theory for linear regression, we know that a smaller value of $\overline{\lambda}_{m,n}$ implies that the average linear models fitted on the minipatches are more accurate.
Here we also use a mild regularity condition before stating a theorem on that characterizes our target $\Delta_j$.
\begin{assumption}\label{assump:FiniteExpectationErrMP}
    For any minipatch $I\subset[N],\,F\subset[M]$, the minipatch prediction has finite expectation:
    $\bE_X(\|\mu_{I,F}(X)\|_2|\bX,\bY)< \infty$; the prediction error of zero predictor also has finite expectation: $\bE_Y(\err(Y,0))< \infty$.
\end{assumption}
The following theorem is a complete version of Theorem \ref{prop:MP_target} in the main paper.
\begin{theorem}\label{prop:MP_target_full}
Consider the linear model described above, with the least squares estimator applied as the base learner in minipatch learning. Suppose that Assumption \ref{assump:FiniteExpectationErrMP} holds. Then for a given feature $j$, our inference target $\Delta_j$ satisfies 
$|\lim_{K\rightarrow\infty}\Delta_j-\Delta_j^{*}|\leq \varepsilon$ with probability at least $1-N^{-c}$ for some constant $c>0$,
where 
\begin{equation}\label{eq:tilde_Delta_linear}
    \Delta_j^{*}:=\left\{\gamma\left[(2-\gamma)\beta_j^{*2}-\left(2-\frac{2M-1}{M-1}\gamma\right)\frac{\|\beta^*_{\backslash j}\|_2^2}{M-1}\right]\right\},
\end{equation} 
$\gamma =\frac{m}{M}$, and 
\begin{equation}\label{eq:linar_MP_target_epsilon2}
\begin{split}
    |\varepsilon| \leq &C\sqrt{\overline{\lambda}_{m,n}}(\|\beta^*\|_2+\sigma_{\epsilon})(\sqrt{\gamma}\|\beta^*\|_2+|\beta_j^*|)\frac{m(m+\sqrt{m}\log N)^{\frac{1}{2}}}{M\sqrt{N}}\\
    &+C\overline{\lambda}_{m,n}(\|\beta^*\|_2^2+\sigma_{\epsilon}^2)\frac{m(m+\sqrt{m}\log N)}{MN}.
\end{split}
\end{equation}
\end{theorem}
As a consequence of Theorem \ref{prop:MP_target_full}, we can show that our confidence interval also has valid coverage for the population feature importance $\Delta_j^{*}$:
\begin{cor}\label{prop:coverage_tilde_Delta_full}
Consider the linear model described above, and suppose that the conditions in Theorem \ref{thm:coverage_buffer} all hold. If the minipatch size $(m,n)$ satisfies 
\begin{equation}\label{eq:gamma_cond_coverage}
\begin{split}
    \frac{m^2}{n^2}\left(\frac{m^2}{\log^2 N}+m\right)\ll& \frac{L^2B^2M^2}{\max\{(\mathbb{E}\overline{\lambda}_{m,n})^2,\overline{\lambda}^2_{m,n}\}},\\
    \frac{m^2}{n^2}\left(\frac{m}{\log N}+\sqrt{m}\right)\ll& \frac{L^2B^2M^2\log N}{\max\{\mathbb{E}\overline{\lambda}_{m,n},\overline{\lambda}_{m,n}\}N},
\end{split}
\end{equation}
then the confidence interval $\hat{\mathbb{C}}_j^{\mathrm{barrier}}$ defined in \eqref{eq:CI_barrier} also has asymptotically $1-\alpha$ coverage for both $\Delta_j^*$ defined in \eqref{eq:tilde_Delta_linear}.
\end{cor}

\paragraph{Comparison with inference targets in prior works: } Under the linear model setting described in this section, we can also derive the closed forms of the inference target of some prior works on model-agnostic feature importance inference.
	\vspace{-2mm}
	\begin{enumerate}[leftmargin=*]
		\item VIMP \citep{williamson2021nonparametric}: The target of VIMP is the predictive power using all features subtracting the predictive power excluding feature $j$. One can show that under the linear model described earlier in this section, their target $\Psi_j(P)=\left[1-\frac{\mathbb{E}[y-\mathbb{E}(y|X)]^2}{\mathrm{Var}(y)}\right]-\left[1-\frac{\mathbb{E}[y-\mathbb{E}(y|X_{\backslash j})]^2}{\mathrm{Var}(y)}\right]=\frac{\beta_j^{*2}}{\|\beta^*\|_2^2+\sigma_{\epsilon}^2}$, which also reflects the relative magnitude of $\beta_j^{*2}$ compared to the rest of the regression coefficients $\|\beta^*_{\backslash j}\|_2^2$, similar to $\Delta_j^{*}$, the population quantity that our inference target is close to. The main difference lies that $\Delta_j^{*}$ takes the difference between $\beta_j^{*2}$ and $\frac{\|\beta^*_{\backslash j}\|_2^2}{M-1}$, while the target of VIMP looks at the ratio. 
		\item Floodgate \citep{zhang2020floodgate}: The MSE gap studied by Floodgate takes a similar form. They aim to provide a lower confidence bound for $\mathcal{I}=\mathbb{E}(y-\mathbb{E}(y|X_{\backslash j}))^2-\mathbb{E}(y-\mathbb{E}(y|X))^2=\beta_j^{*2}$, also reflecting the magnitude of $\beta_j^*$.
	\end{enumerate}

\subsubsection{Correlated Features}
As discussed in Section 4.4 of the main paper, the dependence among features can be a challenge for feature-occlusion-based feature importance inference. Interestingly, \cite{verdinelli2021decorrelated} proposes a couple of decorrelated variable importance quantities to address this challenge and also develop corresponding inference methods, but with certain modeling and consistency assumptions. The Shapley value has also been proposed as a potential solution to the dependent feature problem \citep{owen2017shapley,williamson2020efficient}, but with a different interpretation from our LOCO feature importance measure. In addition, in the literature of causal inference, one also faces the problem of correlated features when making inference for a causal estimand. An idea of balancing \citep{imai2014covariate,fong2018covariate} was also proposed to decorrelate the features, but it requires knowing or estimating well the conditional distribution of one feature given the others.

 For our procedure based on minipatch ensembles, we suspect that the issues brought by feature dependence are less problematic. The key idea is that our minipatch framework involves random subsampling of small subsets of features and observations, and hence strongly correlated features may appear in different minipatches so that the predictive power of each feature can stand out in the absence of its correlated feature. As discussed in the main paper, when the base learner for each minipatch is the least squares estimator, the ensemble is close to a ridge estimator \citep{lejeune2020implicit}, and when the base learner is a decision tree, the ensemble is similar to random forest \citep{louppe2012ensembles}. Both ridge regression and random forest have been observed in prior works to group together and to assign higher importance to correlated features \citep{gromping2009variable,nicodemus2010behaviour}. 
 We also formally verify this argument in theory for linear models, where we show that our inference targets for correlated features are functions of the average regression coefficients for these features, and hence resembles the idea of grouping correlated features together in the literature, e.g., the group Lasso \citep{yuan2006model}, fused Lasso \citep{tibshirani2005sparsity}, and elastic net \citep{zou2005regularization}. 
 
Here we present the detailed theoretical results for the linear model with correlated features. Let $$\beta^{(m)*} = \frac{m}{M}\beta^*+\frac{1}{\binom{M}{m}}\sum_{F\subset [M]}R_F^\top \boldsymbol{\Sigma}_{F,F}^{-1}\boldsymbol{\Sigma}_{F,F^c}\beta_{F^c}^*,
$$
$$
\beta^{(m,-j)*} = \frac{m}{M-1}\beta^{*\backslash j}+\frac{1}{\binom{M-1}{m}}\sum_{F\subset [M],j\notin F}R_F^\top \boldsymbol{\Sigma}_{F,F}^{-1}\boldsymbol{\Sigma}_{F,F^c}\beta_{F^c}^*,
$$ 
with $\beta^{*\backslash j}\in \mathbb{R}^M$ satisfying $\beta^{*\backslash j}_j=0$ and $\beta^{*\backslash j}_{\backslash j}=\beta^*_{\backslash j}$; Also define the norm $\|\cdot\|_{\boldsymbol{\Sigma}}$ for $M$-dimensional vectors as follows: $\|\beta\|_{\boldsymbol{\Sigma}}=(\beta^\top \boldsymbol{\Sigma}\beta)^{\frac{1}{2}}$ for any $\beta\in \mathbb{R}^M$. 
\begin{prop}[$\Delta_j^*$ under linear model with correlated features]\label{prop:MP_target_corr}
Suppose that each row of $\boldsymbol{X}$ independently follows $\mathcal{N}(0,\boldsymbol{\Sigma})$. Then there exists a population quantity $\Delta_j^{*(\rm L)}$ such that the inference target $\Delta_j$ satisfies
$$
|\bE\lim_{K\rightarrow \infty}\Delta_j-\Delta_j^{*(\mathrm{L})}|\leq 2\lambda_{\max}(\boldsymbol{\Sigma})(\lambda_{\max}(\boldsymbol{\Sigma})\|\beta^*\|_2^2 + \sigma_{\epsilon}^2)\frac{m}{N}\left[\frac{m}{M}\mathbb{E}\lambda_{n,m}(\boldsymbol{X})+\frac{m}{M-1}\mathbb{E}\lambda_{n,m}(\boldsymbol{X}_{:,\backslash j})\right],
$$
where $\Delta_j^{*(\mathrm{L})} = \left\|\beta^* - \beta^{(m,-j)*}\right\|_{\boldsymbol{\Sigma}}^2-\left\|\beta^* - \beta^{(m)*}\right\|_{\boldsymbol{\Sigma}}^2$. In particular, when $\boldsymbol{\Sigma}=\begin{pmatrix}
1&\rho&0&\cdots&0\\
\rho&1&0&\cdots&0\\
0&0&1&\cdots&0\\
\vdots&\ddots&\ddots&\ddots&\vdots\\
0&\cdots&\cdots&0&1\\
\end{pmatrix}$, we have 
\begin{align}\label{eq:tilde_Delta_corr1}
\lim_{\rho\rightarrow 0}\Delta_1^{*(\mathrm{L})} =& \gamma(2-\gamma)\beta_1^{*2}-\gamma\left(\frac{2}{M-1}-\frac{m(2M-1)}{(M-1)^2M}\right)\|\beta^*_{\backslash 1}\|_2^2,\\
\lim_{\rho\rightarrow 1}\Delta_1^{*(\mathrm{L})} =& \gamma(2-\gamma)\frac{(M-m-1)^2}{(M-1)^2}(\beta_1^*+\beta_2^*)^2-\gamma\left(\frac{2}{M-1}-\frac{m(2M-1)}{(M-1)^2M}\right)\|\beta^*_{\backslash (1,2)}\|_2^2.   
\end{align}
\end{prop}
	\begin{remark}
		Since \eqref{eq:tilde_Delta_corr1} takes a complicated form, here we discuss the situation when $\gamma=\frac{m}{M}\rightarrow 0$. Then $\lim_{\rho\rightarrow 1}\Delta_j^{*(L)}$ for $j=1,\,2$ scales roughly as $2\gamma\left[(\beta_1^*+\beta_2^*)^2-\frac{1}{M-1}\|\beta^*_{\backslash (1,2)}\|_2^2\right]$,
		where $(\beta_1^*+\beta_2^*)^2$ is the main effect term, compared with the average predictive power of the rest of the features $\frac{1}{M-1}\|\beta^*_{\backslash (1,2)}\|_2^2$. Therefore, when feature 1 and 2 become fully correlated, our inference target groups their coefficients together when performing inference for one of them. Either feature would have strong feature importance unless both have no predictive power for the response $Y$.
	\end{remark}
	It turns out that our inference target resembles the idea of grouping correlated features together in the literature on correlated variables, e.g., the group Lasso \citep{yuan2006model}, fused Lasso \citep{tibshirani2005sparsity}, and elastic net \citep{zou2005regularization}. 
To understand whether this is a desirable property or not, here we consider two different cases.
\begin{itemize}
    \item[(a)] \textbf{Correlated signal features}. When some signal features are highly correlated, the feature importance considered by prior methods such as LOCO-Split \citep{lei2018distribution} or VIMP \citep{williamson2021general} would be small for all these features. On the contrary, our feature importance target would be large for all these features, as long as the average signal in these correlated features is strong.
    \item[(b)] \textbf{Noise feature correlated with signal feature}. Suppose we would like to make inference for a noise feature, which is strongly correlated with some signal features. 
    Our feature importance target can be large for this noise feature, as long as the signal features correlated with it have strong signals. On one hand, this could lead to an inflated Type I error if the goal is to perform conditional independence test; on the other hand, this is justifiable if the goal is to find features useful for prediction. Similar phenomenon also occurs for variable importance measures based on the Shapley value \cite[see, e.g., Theorem 4.1 in][]{owen2017shapley}, where the variable importance of one feature can still be positive even if its corresponding regression coefficient is zero. 
\end{itemize}
\subsection{Proof of Theorem \ref{thm:normal_approx}}\label{sec:proof_main}
Our proof utilizes some key results in \cite{bayle2020cross}, the central limit theorem for cross-validation errors when the predictive algorithm satisfies a certain stability notion. Inspired by this, our main proof is devoted to showing the stability and accounting for the randomness of our minipatch algorithm (Proof of Lemma \ref{lem:bnd_err_terms}). We start by decomposing the deviation of each feature occlusion score to our inference target; note that for any $1\leq i\leq N$, we have:
\begin{align*}
    &\hat{\Delta}_j(X_i,Y_i)-\Delta_j \\
    =&\hat{h}_j(X_i,Y_i;\boldsymbol{X}_{\backslash i,:},\bY_{\backslash i}) - \bE[h_j^{(K)}(X,Y;\boldsymbol{X},\boldsymbol{Y})|\boldsymbol{X},\boldsymbol{Y}]\\
    =&\hat{h}_j(X_i,Y_i;\boldsymbol{X}_{\backslash i,:},\boldsymbol{Y}_{\backslash i})-h_j(X_i,Y_i;\boldsymbol{X}_{\backslash i,:},\boldsymbol{Y}_{\backslash i})\\
    &+h_j(X_i,Y_i;\boldsymbol{X}_{\backslash i,:},\boldsymbol{Y}_{\backslash i}) - \bE[h_j(X_i,Y_i;\boldsymbol{X}_{\backslash i,:},\boldsymbol{Y}_{\backslash i})|\boldsymbol{X}_{\backslash i,:},\boldsymbol{Y}_{\backslash i})]\\
    &+\bE[h_j(X_i,Y_i;\boldsymbol{X}_{\backslash i,:},\boldsymbol{Y}_{\backslash i})|\boldsymbol{X}_{\backslash i,:},\boldsymbol{Y}_{\backslash i})]-\bE[h_j(X,Y;\bX,\bY)|\bX,\bY]\\
    &+\bE[h_j(X,Y;\bX,\bY)|\bX,\bY]-\Delta_j,
\end{align*}
where the first term characterizes the deviation of randomly subsampled minipatch algorithm to its population counterpart (limit as $K\rightarrow \infty$); the second term controls how well the LOO residuals approximate the generalization errors on unseen data; the third term considers how our target changes when $N-1$ instead of $N$ training data is in use; the last term examines how the target changes with randomly subsampled minipatches, compared to the combinatorial average $\mu^*,\,\mu^*_{\backslash j}$. Recall the definition of $h_j(X,Y)$, $\tilde{h}_j(X,Y;\boldsymbol{X},\boldsymbol{Y})$ in the beginning of Section \ref{sec:proof}, we can then further decompose the second term as follows:
\begin{align*}   
    &h_j(X_i,Y_i;\boldsymbol{X}_{\backslash i,:},\boldsymbol{Y}_{\backslash i}) - \bE[h_j(X_i,Y_i;\boldsymbol{X}_{\backslash i,:},\boldsymbol{Y}_{\backslash i})|\boldsymbol{X}_{\backslash i,:},\boldsymbol{Y}_{\backslash i})]\\
    =&h_j(X_i,Y_i)-\bE(h_j(X_i,Y_i)) +\tilde{h}_j(X_i,Y_i;\boldsymbol{X}_{\backslash i,:},\boldsymbol{Y}_{\backslash i})-\bE[\tilde{h}_j(X_i,Y_i;\boldsymbol{X}_{\backslash i,:},\boldsymbol{Y}_{\backslash i})|\boldsymbol{X}_{\backslash i,:},\boldsymbol{Y}_{\backslash i})].
\end{align*}
Let
\begin{align*}
    \varepsilon^{(1)}_{i,j}=&\hat{h}_j(X_i,Y_i;\boldsymbol{X}_{\backslash i,:},\boldsymbol{Y}_{\backslash i})-h_j(X_i,Y_i;\boldsymbol{X}_{\backslash i,:},\boldsymbol{Y}_{\backslash i}),\\
    \varepsilon^{(2)}_{i,j}=&\bE[h_j(X_i,Y_i;\boldsymbol{X}_{\backslash i,:},\boldsymbol{Y}_{\backslash i})|\boldsymbol{X}_{\backslash i,:},\boldsymbol{Y}_{\backslash i})]-\bE[h_j(X,Y;\bX,\bY)|\bX,\bY],\\
    \varepsilon^{(3)}_{i,j}=&\tilde{h}_j(X_i,Y_i;\boldsymbol{X}_{\backslash i,:},\boldsymbol{Y}_{\backslash i})-\bE[\tilde{h}_j(X_i,Y_i;\boldsymbol{X}_{\backslash i,:},\boldsymbol{Y}_{\backslash i})|\boldsymbol{X}_{\backslash i,:},\boldsymbol{Y}_{\backslash i})],
\end{align*}
and define $\varepsilon^{(k)}_j=\frac{1}{\sigma_j\sqrt{N}}\sum_{i=1}^N\varepsilon^{(k)}_{i,j}$, \begin{equation*}
    \begin{split}
        \varepsilon_j^{(4)} = &\frac{\sqrt{N}}{\sigma_j}\left(\bE[h_j(X,Y;\bX,\bY)|\bX,\bY]-\Delta_j\right)\\
        = &\frac{\sqrt{N}}{\sigma_j}\left(\bE[h_j(X,Y;\bX,\bY)|\bX,\bY]-\bE[h_j^{(K)}(X,Y;\bX,\bY)|\bX,\bY]\right).
    \end{split}
\end{equation*}
Our goal is to show that $$\frac{1}{\sigma_j\sqrt{N}}\sum_{i=1}^N(\hat{\Delta}_j(X_i,Y_i)-\Delta_j)=\frac{1}{\sigma_j\sqrt{N}}\sum_{i=1}^N[h_j(X_i,Y_i)-\bE(h_j(X_i,Y_i))]+\sum_{k=1}^4\varepsilon_j^{(k)}$$ converges to standard Gaussian distribution. For the error terms $\varepsilon^{(k)}_j$, $k=1,\dots,4$, the following lemma suggests that they all converge to zero in probability. 
\begin{lem}\label{lem:bnd_err_terms}
Under the same conditions as in Theorem \ref{thm:normal_approx}, $\varepsilon_j^{(k)}\overset{p}{\rightarrow} 0$, $k=1,2,3$.
\end{lem}
While for $\frac{1}{\sigma_j\sqrt{N}}\sum_{i=1}^N[h_j(X_i,Y_i)-\bE(h_j(X_i,Y_i))]$, we note that $\bE[h_j(X_i,\,Y_i) - \bE h_j(X_i,\,Y_i)]^3/\sigma_j^3 \leq C$, and hence 
$$
\frac{1}{(\sigma_j\sqrt{N})^3}\sum_{i=1}^N\bE[h_j(X_i,\,Y_i) - \bE h_j(X_i,\,Y_i)]^3 \leq C/\sqrt{N}\rightarrow 0.
$$
Therefore, Lyapunov's condition holds, implying 
$\frac{1}{\sigma_j\sqrt{N}}\sum_{i=1}^N[h_j(X_i,Y_i)-\bE(h_j(X_i,Y_i))]\overset{d}{\rightarrow}\cN(0,1)$. 
Finally, applying Slutsky's theorem finishes the proof of Theorem \ref{thm:normal_approx}.
\subsection{Proof of Theorem \ref{thm:var_est} and Corollary \ref{cor:coverage-width}}
\begin{proof}[Proof of Theorem \ref{thm:var_est}]
We would like to apply the variance consistency result (Theorem 5 in \cite{bayle2020cross}) in our setting, with the $h_n(Z_i,Z_{B_j})$ under their notation being substituted by $h_j(X_i,Y_i;\bX_{\del i,:},\bY_{\del i})$, denoted by $\tilde{\Delta}_j(X_i,Y_i)$ in this proof. The only difference between $\tilde{\Delta}_j(X_i,Y_i)$ and $\hat{\Delta}_j(X_i,Y_i)$ is that the former is computed using the combinatorial minipatch ensembles (the deterministic minipatch algorithm), while the latter is based on random sampling of minipatches. Also define 
\begin{align}\label{eq:bar_h_j}
    \bar{h}_j(X_i,Y_i)=\bE_{\bX_{\del i, :},\bY_{\del i}}[h_j(X_i,Y_i;\bX_{\del i,:},\bY_{\del i})|X_i,Y_i].
\end{align}
Let $\bar{\tilde{\Delta}}_j=\frac{1}{N}\sum_{i=1}^N\tilde{\Delta}_j(X_i,Y_i)$, and 
\begin{align*}
    \tilde{\sigma}_j^2&=\mathrm{Var}_{X,Y}(\bar{h}_j(X,Y)),\\
    \hat{\tilde{\sigma}}_j^2&=\frac{1}{N}\sum_{i=1}^N\left(\tilde{\Delta}_j(X_i,Y_i)-\bar{\tilde{\Delta}}_j(X_i,Y_i)\right)^2.
\end{align*}

Here, we first show that the moment condition in Assumption \ref{assump:third_moment} immediately implies the uniform integrability of $[h_j(X_i,Y_i)-\bE(h_j(X_i,Y_i))]^2/\sigma_j^2$. Let $\xi_{N,i} = [h_j(X_i,Y_i)-\bE(h_j(X_i,Y_i))]^2/\sigma_j^2$, we can then write $\sup_N\bE[|\xi_{N,i}|\ind(|\xi_{N,i}|>t)]\leq \sup_N(\bE|\xi_{N,i}|^{3/2})^{2/3}[\bP(|\xi_{N,i}|>t)]^{1/3}\leq C(\frac{\bE|\xi_{N,i}|}{t})^{1/3} = Ct^{-\frac{1}{3}}$, which converges to zero as $t\rightarrow \infty$. Then Theorem 5 in \cite{bayle2020cross} suggests that, if $\gamma_{loss}(h_j)=o(\tilde{\sigma}_j^2/N)$ and $\gamma_{ms}(h_j)=o(\tilde{\sigma}_j^2)$ hold, we have $\frac{\hat{\tilde{\sigma}}_j^2}{\tilde{\sigma}_j^2}\overset{p}{\rightarrow}1$. In the following, we will show (i) $\lim_{N\rightarrow \infty}\frac{\sigma_j^2}{\tilde{\sigma}_j^2}=1$; (ii) the stability quantities associated with $h_j$ satisfy $\gamma_{loss}(h_j)=o(\sigma_j^2/N)$, $\gamma_{ms}(h_j)=o(\sigma_j^2)$; (iii) $\frac{\hat{\sigma}_j^2}{\hat{\tilde{\sigma}}_j^2}\overset{p}{\rightarrow}1$. Combining these three results and Theorem 5 in \cite{bayle2020cross}, we will then have $\frac{\hat{\sigma}_j^2}{\sigma_j^2}\overset{p}{\rightarrow}1$, and our proof will be complete.
\begin{itemize}
    \item[(i)]To show the closeness between the variances of $\bar{h}_j(X,Y)$ and $h_j(X,Y)$, first we can write out
    \begin{align*}
        \left|\frac{\tilde{\sigma}_j^2}{\sigma_j^2}-1\right|&=\sigma_j^{-2}\bE([\bar{h}_j(X,Y)-\bE(\bar{h}_j(X,Y))]^2-[h_j(X,Y)-\bE(h_j(X,Y))]^2).
    \end{align*}
    Since for any random variables $\xi_1,\xi_2$, $|\bE(\xi_1^2-\xi_2^2)|=|\bE(\xi_1-\xi_2)^2+2\xi_2(\xi_1-\xi_2)|\leq \bE(\xi_1-\xi_2)^2+2\sqrt{\bE(\xi_2^2)}\sqrt{\bE(\xi_1-\xi_2)^2}$, we can further bound $\left|\frac{\tilde{\sigma}_j^2}{\sigma_j^2}-1\right|$ by 
    \begin{align*}
        \left|\frac{\tilde{\sigma}_j^2}{\sigma_j^2}-1\right|
        &\leq \sigma_j^{-2}\bE[\bar{h}_j(X,Y)-h_j(X,Y)-\bE(\bar{h}_j(X,Y)-h_j(X,Y))]^2\\
        &\hspace{3mm}+2\sigma_j^{-1}\left(\bE[\bar{h}_j(X,Y)-h_j(X,Y)-\bE(\bar{h}_j(X,Y)-h_j(X,Y))]^2\right)^{1/2}\\
        &\leq\sigma_j^{-2}\bE[\bar{h}_j(X,Y)-h_j(X,Y)]^2+2\sigma_j^{-1}\left(\bE[\bar{h}_j(X,Y)-h_j(X,Y)]^2\right)^{1/2},
    \end{align*}
    where we have applied the fact that for any random variable $\xi$, $\mathrm{Var}(\xi)\leq \bE(\xi^2)$. As has been shown in the last part of the proof of Lemma \ref{lem:bnd_err_terms}, $\bE[\bar{h}_j(X,Y)-h_j(X,Y)]^2\leq \frac{4L^2n^2\stb(m,n)}{N^2}$. Hence $\left|\frac{\tilde{\sigma}_j^2}{\sigma_j^2}-1\right|\leq \frac{4L^2n^2\stb(m,n)}{\sigma_j^2N^2}+\frac{4Ln\sqrt{\stb(m,n)}}{\sigma_jN}=o(1)$ by Assumption \ref{assump:mpsize}. 
    \item[(ii)] As shown when bounding $\varepsilon_j^{(3)}$ in the proof of Lemma \ref{lem:bnd_err_terms}, $\gamma_{loss}(h_j)\leq \frac{4L^2n^2\stb(m,n)}{(1-\gamma)^2(N-1)^2}=o(\frac{\sigma_j^2}{N})$. Similar to that proof, here we let $(X_{N+1},Y_{N+1})$ be a sample from $\cP$ which is independent from $(\bX,\bY)$, and denote by $(\bX^{\del l}_{\del i,:},\bY^{\del l}_{\del i})$ the $N-1$ training set with sample $i$ excluded, and sample $l$ replaced by $(X_{N+1},Y_{N+1})$. By the definition of the mean-squared stability in \cite{bayle2020cross}, 
    \begin{align*}
        \gamma_{ms}(h_j)&=\frac{1}{N-1}\sum_{l\neq i}\bE[(h_j(X_i,Y_i;\bX_{\del i,},\bY_{\del i})-h_j(X_i,Y_i;\bX^{\del l}_{\del i,},\bY^{\del l}_{\del i})^2].
    \end{align*}
    Then by following similar arguments as in the proof of Lemma \ref{lem:bnd_err_terms}, we have $\gamma_{ms}(h_j)\leq \frac{4L^2n^2\stb(m,n)}{(1-\gamma)^2(N-1)^2}=o(\frac{\sigma_j^2}{N})$. Thus Applying Theorem 5 in \cite{bayle2020cross} leads to $\frac{\hat{\tilde{\sigma}}_j^2}{\tilde{\sigma}_j^2}\overset{p}{\rightarrow}1$.
    \item[(iii)] Now we show the closeness between our own variance estimate $\hat{\sigma}_j^2=\frac{1}{N}\sum_{i=1}^N(\hat{\Delta}_j(X_i,Y_i)-\bar{\Delta}_j)^2$ and $\hat{\tilde{\sigma}}_j^2=\frac{1}{N}\sum_{i=1}^N(\tilde{\Delta}_j(X_i,Y_i)-\bar{\tilde{\Delta}}_j)^2$. Similar to the proof in (i), we can first write
    \begin{align*}
        &|\hat{\sigma}_j^2-\hat{\tilde{\sigma}}_j^2|\\
        \leq &\frac{1}{N}\sum_{i=1}^N(\hat{\Delta}_j(X_i,Y_i)-\tilde{\Delta}_j(X_i,Y_i))^2+2\hat{\tilde{\sigma}}_j\sqrt{\frac{1}{N}\sum_{i=1}^N(\hat{\Delta}_j(X_i,Y_i)-\tilde{\Delta}_j(X_i,Y_i))^2}.
    \end{align*}
    Recall that we have already shown the closeness between $\hat{\Delta}_j(X_i,Y_i)=\hat{h}_j(X_i,Y_i;\bX_{\del i,:},\bY_{\del i})$ and $\tilde{\Delta}_j(X_i,Y_i)=h_j(X_i,Y_i;\bX_{\del i,:},\bY_{\del i})$ when bounding $\varepsilon_j^{(1)}$ in the proof of Lemma \ref{lem:bnd_err_terms}, we would like to reuse some of the notations and intermediate results in that proof.
    Let $\alpha_i^{(1)}=\|\mu^{-j}_{-i}(X_i)-\tilde{\mu}^{-j}_{-i}(X_i)\|_2$, $\alpha_i^{(2)}=\|\mu_{-i}(X_i)-\tilde{\mu}_{-i}(X_i)\|_2$, $\alpha_i^{(3)}=\|\tilde{\mu}^{-j}_{-i}(X_i)-\mu^{*-j}_{-i}(X_i)\|_2$, and $\alpha_i^{(4)}=\|\tilde{\mu}_{-i}(X_i)-\mu^{*}_{-i}(X_i)\|_2$, where $\tilde{\mu}^{-j}_{-i}(X_i)$ and $\tilde{\mu}_{-i}(X_i)$ are defined in \eqref{eq:tilde_mu_def}. Then by Assumption \ref{assump:Lip}, we have $|\hat{\Delta}_j(X_i,Y_i)-\tilde{\Delta}_j(X_i,Y_i)|\leq L\sum_{l=1}^4\alpha_i^{(l)}$, and hence
    \begin{align*}
        &|\hat{\sigma}_j^2-\hat{\tilde{\sigma}}_j^2|\\
        \leq&\frac{L^2}{N}\sum_{i=1}^N\left(\sum_{l=1}^4\alpha_i^{(l)}\right)^2+\frac{2L\hat{\tilde{\sigma}}_j}{\sqrt{N}}\sqrt{\sum_{i=1}^N\left(\sum_{l=1}^4\alpha_i^{(l)}\right)^2}\\
        \leq&\frac{L^2}{N}\left(\sum_{l=1}^4\sum_{i=1}^N\alpha_i^{(l)}\right)^2+\frac{2L\hat{\tilde{\sigma}}_j}{\sqrt{N}}\sum_{i=1}^N\sum_{l=1}^4\alpha_i^{(l)}\\
        \leq&\sigma_j^2\left(\sum_{l=1}^4\frac{L}{\sigma_j\sqrt{N}}\sum_{i=1}^N\alpha_i^{(l)}\right)^2+2\hat{\tilde{\sigma}}_j\sigma_j\left(\sum_{l=1}^4\frac{L}{\sigma_j\sqrt{N}}\sum_{i=1}^N\alpha_i^{(l)}\right).
    \end{align*}
    Recall that we have shown in the proof of Lemma \ref{lem:bnd_err_terms} that $|\varepsilon_j^{(1)}|\leq \sum_{l=1}^4\frac{L}{\sigma_j\sqrt{N}}\sum_{i=1}^N\alpha_i^{(l)}\overset{p}{\rightarrow}0$, and we have just shown in (i) and (ii) that $\frac{\tilde{\sigma}_j^2}{\sigma_j^2}\rightarrow 1$, $\frac{\hat{\tilde{\sigma}}_j^2}{\tilde{\sigma}_j^2}\overset{p}{\rightarrow} 1$. Hence, $\frac{|\hat{\sigma}_j^2-\hat{\tilde{\sigma}}_j^2|}{\hat{\tilde{\sigma}}_j^2}\overset{p}{\rightarrow}0$, or equivalently, $\frac{\hat{\sigma}_j^2}{\hat{\tilde{\sigma}}_j^2}\overset{p}{\rightarrow}1$, which completes our proof.
\end{itemize}
\end{proof}
\begin{proof}[Proof of Corollary \ref{cor:coverage-width}]
We combine Theorem \ref{thm:normal_approx} and Theorem \ref{thm:var_est}, and apply Slutsky's theorem to obtain
\begin{align*}
    \sqrt{N}\hat{\sigma}_j^{-1}(\bar{\Delta}_j-\Delta_j)\overset{d}{\rightarrow}\mathcal{N}(0,1).
\end{align*}
Then the coverage probability satisfies
\begin{align*}
    \lim_{N\rightarrow \infty}\bP(\Delta_j\in \hat{\bC}_j)=\lim_{N\rightarrow \infty}\bP(\sqrt{N}\hat{\sigma}_j^{-1}|\bar{\Delta}_j-\Delta_j|\leq z_{\alpha/2})=1-\alpha.
\end{align*}
In addition, since $\sqrt{N}\sigma_j^{-1}|\hat{\bC}_j|=2z_{\alpha/2}\frac{\hat{\sigma}_j}{\sigma_j}$ and Theorem \ref{thm:var_est} suggests that $\frac{\hat{\sigma}_j^2}{\sigma_j^2}\overset{p}{\rightarrow}1$, we have $\sqrt{N}\sigma_j^{-1}|\hat{\bC}_j|\overset{p}{\rightarrow}2z_{\alpha/2}$.
\end{proof}
\subsection{Proof of Lemma \ref{lem:bnd_err_terms}}
We prove the convergence in probability results for the three error terms in Lemma \ref{lem:bnd_err_terms} separately.
\subsubsection{Bounding $\varepsilon_j^{(1)}$}\label{sec:bnd_err1}
Here we prove the convergence in probability result for $\varepsilon_j^{(1)}$ by concentrating the random minipatch algorithm around its population version. First note that by the Lipschitz condition (Assumption \ref{assump:Lip}), one can show that
\begin{align*}
    |\varepsilon_j^{(1)}|\leq\frac{L}{\sigma_j\sqrt{N}}\sum_{i=1}^N\left(\|\mu^{*-j}_{-i}(X_i)-\mu^{-j}_{-i}(X_i)\|_2+\|\mu^*_{-i}(X_{i})-\mu_{-i}(X_i)\|_2\right).
\end{align*}

Recall that we have defined $\mu_{I,F}(X_i)=(H(\bX_{I,F},\bY_{I_k}))(X_{i,F})$ as the prediction of the base learner trained on $(\bX_{I,F},\bY_{I})$. Thus $\mu_{-i}(X_i)$, $\mu^{-j}_{-i}(X_i)$, $\mu^*_{-i}(X_i)$, and $\mu^{*-j}_{-i}(X_i)$ can be written out as follows:
\begin{align*}
    \mu_{-i}(X_i)&=\frac{1}{\sum_{k=1}^K\ind(i\notin I_k)}\sum_{k=1}^K\ind(i\notin I_k)\mu_{I_k,F_k}(X_i),\\
    \mu_{-i}^{-j}(X_i)&=\frac{1}{\sum_{k=1}^K\ind(i\notin I_k)\ind(j\notin F_k)}\sum_{k=1}^K\ind(i\notin I_k)\ind(j\notin F_k)\mu_{I_k,F_k}(X_i),\\
    \mu^*_{-i}(X_i)&=\frac{1}{\binom{N-1}{n}\binom{M}{m}}\sum_{\substack{I\subset[N],|I|=n,\\F\subset[M],|F|=m}}\ind(i\notin I)\mu_{I,F}(X_i)\\
    &=\frac{\binom{N}{n}}{\binom{N-1}{n}}\frac{1}{K}\sum_{k=1}^K\bE[\ind(i\notin I_k)\hat{\mu}_{I_k,F_k}(X_i)|\bX,\bY],\\
    \mu_{-i}^{*-j}(X_i)&=\frac{1}{\binom{N-1}{n}\binom{M-1}{m}}\sum_{\substack{I\subset[N],|I|=n,\\F\subset[M],|F|=m}}\ind(i\notin I)\ind(j\notin F)\mu_{I,F}(X_i)\\
    &=\frac{\binom{N}{n}\binom{M}{m}}{\binom{N-1}{n}\binom{M-1}{m}}\frac{1}{K}\sum_{k=1}^K\bE[\ind(i\notin I_k)\ind(j\notin F_k)\mu_{I_k,F_k}(X_i)|\bX,\bY].
\end{align*}

In addition, we define the following intermediate predictors that would be helpful in the proofs:
\begin{equation}\label{eq:tilde_mu_def}
    \begin{split}
        \tilde{\mu}^{-j}_{-i}(X_i)&=\frac{\binom{N}{n}\binom{M}{m}}{\binom{N-1}{n}\binom{M-1}{m}}\frac{1}{K}\sum_{k=1}^K\ind(i\notin I_k)\ind(j\notin F_k)\mu_{I_k,F_k}(X_i),\\   
\tilde{\mu}_{-i}(X_i)&=\frac{\binom{N}{n}}{\binom{N-1}{n}}\frac{1}{K}\sum_{k=1}^K\ind(i\notin I_k)\mu_{I_k,F_k}(X_i).
    \end{split}
\end{equation}

Then $\epsilon_j^{(1)}$ can be further bounded as follows:
\begin{align*}
    |\varepsilon_j^{(1)}|&\leq\frac{L}{\sigma_j\sqrt{N}}\sum_{i=1}^N\left[\|\mu^{-j}_{-i}(X_i)-\mu^{*-j}_{-i}(X_i)\|_2+\|\mu_{-i}(X_{i})-\mu^*_{-i}(X_i)\|_2\right]\\
    &\leq\frac{L}{\sigma_j\sqrt{N}}\sum_{i=1}^N\Big[\|\mu^{-j}_{-i}(X_i)-\tilde{\mu}^{-j}_{-i}(X_i)\|_2+\|\mu_{-i}(X_{i})-\tilde{\mu}_{-i}(X_i)\|_2+\\&\hspace{20mm}\|\tilde{\mu}^{-j}_{-i}(X_i)-\mu^{*-j}_{-i}(X_i)\|_2+\|\tilde{\mu}_{-i}(X_{i})-\mu^*_{-i}(X_i)\|_2\Big].
\end{align*}
Let 
\begin{align*}
    \RNum{1}=\frac{L}{\sigma_j\sqrt{N}}\sum_{i=1}^N\|\mu^{-j}_{-i}(X_i)-\tilde{\mu}^{-j}_{-i}(X_i)\|_2,\quad\RNum{2}&=\frac{L}{\sigma_j\sqrt{N}}\sum_{i=1}^N\|\mu_{-i}(X_i)-\tilde{\mu}_{-i}(X_i)\|_2,\\
    \RNum{3}=\frac{L}{\sigma_j\sqrt{N}}\sum_{i=1}^N\|\tilde{\mu}^{-j}_{-i}(X_i)-\mu^{*-j}_{-i}(X_i)\|_2,\quad\RNum{4}&=\frac{L}{\sigma_j\sqrt{N}}\sum_{i=1}^N\|\tilde{\mu}_{-i}(X_i)-\mu^*_{-i}(X_i)\|_2,
\end{align*} 
and we will upper bound these four terms separately. 
\begin{enumerate}
    \item To deal with term $\RNum{1}$, we fist let $\hat{p}_{i,j}=\frac{\sum_{k=1}^K\ind(i\notin I_k)\ind(j\notin F_k)}{K}$ and $p_{i,j}=\frac{(N-n)(M-m)}{NM}$. One can then show that
    \begin{align*}
        |\RNum{1}|\leq &\frac{L\sqrt{N}}{\sigma_j}\frac{1}{NK}\sum_{i,k}|\hat{p}_{i,j}^{-1}-p_{i,j}^{-1}|\ind(i\notin I_k)\ind(j\notin F_k)\|\mu_{I_k,F_k}(X_i)\|_2\\
        \leq &\frac{L\sqrt{N}}{\sigma_j}\frac{1}{NK}\sum_{i,k}\ind(i\notin I_k)\|\mu_{I_k,F_k}(X_i)\|_2 \max_{1\leq i\leq N}|\hat{p}_{i,j}^{-1}-p_{i,j}^{-1}|.
    \end{align*}
    Now we first prove an upper bound for $\frac{1}{NK}\sum_{i,k}\ind(i\notin I_k)\|\mu_{I_k,F_k}(X_i)\|_2$ with high probability. In particular, let $\gamma_k = \frac{1}{N}\sum_{i=1}^N\ind(i\notin I_k)\|\mu_{I_k,F_k}(X_i)\|_2$. Then by Assumption \ref{assump:bnd_mu} and Jensen's inequality, we have $$\bE_{I_k,F_k}(\gamma_k|\bX,\bY)\leq \left[\frac{1}{N}\sum_{i=1}^N\bE_{I,F}[\ind(i\notin I)\|\mu_{I,F}(X_i)\|_2^2]\right]^{\frac{1}{2}}\leq B.$$ Furthermore, we can apply Jensen's inequality again to obtain $$\mathrm{Var}(\gamma_k|\bX,\bY)\leq \bE(\gamma_k^2|\bX,\bY)\leq \frac{1}{N}\sum_{i=1}^N\bE_{I,F}[\ind(i\notin I)\|\mu_{I,F}(X_i)\|_2^2]\leq B^2.$$ Therefore, we apply Chebyshev's inequality to reveal that
    \begin{align*}
        \bP\left(\frac{1}{K}\sum_k\gamma_k>2B\right)&\leq \bP\left(\frac{1}{K}\sum_k[\gamma_k - \bE(\gamma_k)]>B\right)\\
        &\leq \frac{\mathrm{Var}(\frac{1}{K}\sum_k\gamma_k|\bX,\bY)}{B^2}\\
        &=\frac{\mathrm{Var}(\gamma_k|\bX,\bY)}{KB^2}\leq \frac{1}{K}.
    \end{align*}
    Here, we have applied the fact that $\gamma_1,\,\dots,\,\gamma_K$ are conditionally independent given $\bX$, $\bY$. Hence the tail probability for $\RNum{1}$ can be further decomposed as follows
    \begin{align*}
        \bP(|\RNum{1}|>\epsilon)&\leq \bP\left(\frac{1}{K}\sum_k\gamma_k>2B\right) + \bP\left(\max_{1\leq i\leq N}|\hat{p}_{i,j}^{-1}-p_{i,j}^{-1}|>\frac{\epsilon\sigma_j}{2BL\sqrt{N}}\right)\\
        &\leq \frac{1}{K} + \sum_{i=1}^N\bP\left(|\hat{p}_{i,j}^{-1}-p_{i,j}^{-1}|>\frac{\epsilon\sigma_j}{2BL\sqrt{N}}\right).
    \end{align*}
    Note that if $|\hat{p}_{i,j}-p_{i,j}|\leq \frac{p_{i,j}}{2}$, $|\hat{p}_{i,j}^{-1}-p_{i,j}^{-1}|=\frac{|\hat{p}_{i,j}-p_{i,j}|}{\hat{p}_{i,j}p_{i,j}}\leq\frac{2|\hat{p}_{i,j}-p_{i,j}|}{p_{i,j}^2}$. Thus 
    $|\hat{p}_{i,j}^{-1}-p_{i,j}^{-1}|>\frac{\epsilon\sigma_j}{2BL\sqrt{N}}$ implies 
    $$
    |\hat{p}_{i,j}-p_{i,j}|>\min\left\{\frac{p_{i,j}}{2},\frac{\epsilon\sigma_jp_{i,j}^2}{4BL\sqrt{N}}\right\}.
    $$
    Since $\hat{p}_{i,j}-p_{i,j}=\sum_{k=1}^K\frac{1}{K}\left[\ind(i\notin I_k)\ind(j\notin F_k)-p_{i,j}\right]$ is a sum of independent bounded random variables with mean zero, we can apply the Hoeffding's inequality \cite[see e.g., Proposition 2.5 in][and examples therein]{wainwright2019high} to obtain that 
    $$
    \bP\left(|\hat{p}_{i,j}-p_{i,j}|>\min\left\{\frac{p_{i,j}}{2},\frac{\epsilon\sigma_jp_{i,j}^2}{4BL\sqrt{N}}\right\}\right)\leq \exp\left\{-K\min\left\{\frac{p^2_{i,j}}{2},\frac{\epsilon^2\sigma_j^2p_{i,j}^4}{8B^2L^2N}\right\}\right\},
    $$
    which further implies that
    \begin{align*}
        \bP(|\RNum{1}|>\epsilon)&\leq\frac{1}{K}+\sum_{i=1}^N\exp\left\{-K\min\left\{\frac{p^2_{i,j}}{2},\frac{\epsilon^2\sigma_j^2p_{i,j}^4}{8B^2L^2N}\right\}\right\}\\
        &\leq \frac{1}{K}+\exp\left\{\log N-K\min\left\{\frac{\min_ip^2_{i,j}}{2},\frac{\epsilon^2\sigma_j^2\min_ip_{i,j}^4}{8B^2L^2N}\right\}\right\}.
    \end{align*}
    By Assumption \ref{assump:mpsize}, $p_{i,j}=(1-\frac{n}{N})(1-\frac{m}{M})\geq (1-\gamma)^2$. Since $K\gg (\frac{B^2L^2N}{\sigma_j^{2}}+1)\log N$, there exists $N_0>0$ such that when $N\geq N_0$, the number of minipatches $K(N)\geq \Big[\frac{12}{(1-\gamma)^8\epsilon^2}+\frac{3}{(1-\gamma)^4}\Big]\Big(\frac{B^2L^2N}{\sigma_j^{2}}+1\Big)\log N$, which implies that
    \begin{align*}
        &\exp\left\{\log N-K\min\left\{\frac{(1-\gamma)^4}{2},\frac{\epsilon^2\sigma_j^2(1-\gamma)^8}{8B^2L^2N}\right\}\right\}\\
        \leq&\exp\left\{\log N-\Big(\frac{B^2L^2N}{\sigma_j^{2}}+1\Big)\min\left\{\frac{3}{2},\frac{3\sigma_j^2}{2B^2L^2N}\right\}\log N\right\}\\
        \leq&\exp\left\{-\frac{1}{2}\log N\right\},
    \end{align*}
    when $N\geq N_0$. Therefore, for any $\epsilon>0$, $\lim_{N\rightarrow \infty}\bP(|\RNum{1}|>\epsilon)=0$, or equivalently, $\RNum{1}\overset{p}{\rightarrow}0$.
    \item For term $\RNum{2}$, following similar arguments to bounding term $\RNum{1}$, we have that for any $\epsilon>0$,
    \begin{align*}
        \bP(|\RNum{2}|>\epsilon) &\leq \frac{1}{K}+\sum_{i=1}^N\bP\left(|\hat{p}_i-p_i|>\min\left\{\frac{p_i}{2},\frac{\epsilon\sigma_jp_i^2}{8BL\sqrt{N}}\right\}\right)\\
        &\leq\frac{1}{K}+\exp\left\{\log N-K\min\left\{\frac{\min_ip_i^2}{2},\frac{\epsilon^2\sigma_j^2\min_ip_i^4}{8B^2L^2N}\right\}\right\},
    \end{align*}
    where $\hat{p}_i=\frac{\sum_{k=1}^K\ind(i\notin I_k)}{K}$ and $p_i=\frac{N-n}{N}$. Since $p_i\geq p_{i,j}$, using the same argument for showing the consistency of $\RNum{1}$, we have $\lim_{N\rightarrow \infty}\bP(|\RNum{2}|>\epsilon)=0$.
    \item While for term $\RNum{3}$, we first define $Z_k\in \bR^{Nd}$ as follows:
    \begin{align*}
        &(Z_k)_{((i-1)d+1):id}\\
    =&\frac{1}{K}\ind(i\notin I_k)\ind(j\notin F_k)\mu_{I_k,F_k}(X_i)-\frac{1}{K}\bE\left[\ind(i\notin I_k)\ind(j\notin F_k)\mu_{I_k,F_k}(X_i)|\bX,\bY\right],
    \end{align*}
    for $i=1,\dots,N$.     
    Now we can write out $\RNum{3}$ as follows:
    \begin{align*}
        \RNum{3}&=\frac{LNM}{\sigma_j\sqrt{N}(N-n)(M-m)}\sum_{i=1}^N\left\|\sum_{k=1}^K(Z_k)_{((i-1)d+1):id}\right\|_2\\
        &\leq\frac{L\sqrt{N}}{\sigma_j(1-\gamma)^2}\frac{1}{N}\sum_{i=1}^N\left\|\sum_{k=1}^K(Z_k)_{((i-1)d+1):id}\right\|_2.
    \end{align*}
    Note that
    \begin{align*}
        &\bE\Bigg[\Bigg(\frac{1}{N}\sum_{i=1}^N\Bigg\|\sum_{k=1}^K(Z_k)_{((i-1)d+1):id}\Bigg\|_2\Bigg)^2|\bX,\bY\Bigg]\\
        &\quad\leq \frac{1}{N}\sum_{i=1}^N\bE\Bigg(\Bigg\|\sum_{k=1}^K(Z_k)_{((i-1)d+1):id}\Bigg\|_2^2|\bX,\bY\Bigg)\\
        &\quad\leq \frac{1}{N}\sum_{i=1}^N\frac{1}{K}\bE\Bigg[\ind(i\notin I_k)\ind(j\notin F_k)\|\mu_{I_k,F_k}(X_i)\|_2^2|\bX,\bY\Bigg]\\
        &\quad \leq \frac{B^2}{K},
    \end{align*}
    where the last line is due to Assumption \ref{assump:bnd_mu}. Therefore, we can apply Chebyshev's inequality and get the following:
    \begin{align*}
        \bP(|\RNum{3}|>\epsilon)&\leq \bP(\frac{1}{N}\sum_{i=1}^N\left\|\sum_{k=1}^K(Z_k)_{((i-1)d+1):id}\right\|_2>\frac{\epsilon \sigma_j(1-\gamma)^2}{L\sqrt{N}}\\
        &\leq \frac{B^2L^2N}{\epsilon^2 \sigma_j^2(1-\gamma)^4K}.
    \end{align*}
    Since we have assumed $K\gg (\frac{L^2B^2N}{\sigma_j^{2}}+1)\log N$, for any $\epsilon>0$, we have $\lim_{N\rightarrow \infty}\bP(|\RNum{3}|>\epsilon)=0$.
    \item By redefining $$(Z_k)_{((i-1)d+1):id}=\frac{1}{K}\ind(i\notin I_k)\mu_{I_k,F_k}(X_i)-\frac{1}{K}\bE[\ind(i\notin I_k)\mu_{I_k,F_k}(X_i)],$$ and following almost the same argument as above, we can also show that for any $\epsilon>0$, $\lim_{N\rightarrow \infty}\bP(|\RNum{4}|>\epsilon)=0$.
\end{enumerate}
Therefore, combing all the convergence in probability results for $\RNum{1}$, $\RNum{2}$, $\RNum{3}$, and $\RNum{4}$, we have $\varepsilon^{(1)}_j\overset{p}{\rightarrow}0$.
\subsubsection{Bounding $\varepsilon_j^{(2)}$}\label{sec:bnd_err2}
First note that
\begin{equation*}
\begin{split}
     \big|\varepsilon_j^{(2)}\big|&\leq \frac{1}{\sigma_j\sqrt{N}}\sum_{i=1}^N\big|\bE_{(X,Y)}\{\err(Y,\mu^{*-j}_{-i}(X))-\err(Y,\mu^{*-j}(X))\}\big|\\
    &\hspace{22mm}+\big|\bE_{(X,Y)}\{\err(Y,\mu^{*}_{-i}(X))-\err(Y,\mu^{*}(X))\}\big|\\
    &\leq\frac{L}{\sigma_j\sqrt{N}}\sum_{i=1}^N\bE_{X}\|\mu^{*-j}_{-i}(X)-\mu^{*-j}(X)\|_2+\bE_{X}\|\mu^{*}_{-i}(X)-\mu^{*}(X)\|_2,
\end{split}
\end{equation*}
where the expectation is taken over the test data $(X,Y)$, conditioning on the training data $\bX,\,\bY$. We can then further bound $\bE|\varepsilon_j^{(2)}|$ as follows:
\begin{equation}\label{eq:epsilon2_bnd0}
    \begin{split}
        \bE\big|\epsilon_j^{(2)}\big|^2&\leq \frac{L^2N}{\sigma_j^2}\bE\left[\frac{1}{N}\sum_{i=1}^N\bE_X\|\mu^{*-j}_{-i}(X)-\mu^{*-j}(X)\|_2+\bE_X\|\mu^{*}_{-i}(X)-\mu^{*}(X)\|_2\right]^2\\
        &\leq \frac{2L^2N}{\sigma_j^2}\bE\left[\frac{1}{N}\sum_{i=1}^N\bE_X\|\mu^{*-j}_{-i}(X)-\mu^{*-j}(X)\|_2^2+\bE_X\|\mu^{*}_{-i}(X)-\mu^{*}(X)\|_2^2\right]\\
        &\leq \frac{2L^2N}{\sigma_j^2}\big(\bE\|\mu^{*-j}_{-i}(X)-\mu^{*-j}(X)\|_2^2+\bE\|\mu^{*}_{-i}(X)-\mu^{*}(X)\|_2^2\big),
    \end{split}
\end{equation}
where the final expectation is taken over the random training data, test data, and the random subsampling of minipatches; the second line is due to Jensen's inequality.
Recall the definition of $\mu^{*-j}_{-i}(X)$, $\mu^{*-j}(X)$, $\mu^{*}_{-i}(X)$, $\mu^{*}(X)$, and $\mu_{I,F}(X)$ in the beginning of Section \ref{sec:proof}. Then we can write
\begin{align*}
    \mu_{-i}^{*-j}(X)&=\frac{1}{\binom{N-1}{n}\binom{M-1}{m}}\sum_{\substack{I\subset[N],|I|=n,\\F\subset[M],|F|=m}}\ind(i\notin I)\ind(j\notin F)\mu_{I,F}(X),\\
    \mu^{*-j}(X)&=\frac{1}{\binom{N}{n}\binom{M-1}{m}}\sum_{\substack{I\subset[N],|I|=n,\\F\subset[M],|F|=m}}\ind(j\notin F)\mu_{I,F}(X)\\
    &=\frac{N-n}{N}\mu_{-i}^{*-j}(X)+\frac{n}{N}\frac{1}{\binom{N-1}{n-1}\binom{M-1}{m}}\sum_{\substack{I\subset[N],|I|=n,\\F\subset[M],|F|=m}}\ind(i\in I)\ind(j\notin F)\mu_{I,F}(X);\\
    \mu^*_{-i}(X)&=\frac{1}{\binom{N-1}{n}\binom{M}{m}}\sum_{\substack{I\subset[N],|I|=n,\\F\subset[M],|F|=m}}\ind(i\notin I)\mu_{I,F}(X),\\
    \mu^*(X)&=\frac{1}{\binom{N}{n}\binom{M}{m}}\sum_{\substack{I\subset[N],|I|=n,\\F\subset[M],|F|=m}}\mu_{I,F}(X)\\
    &=\frac{N-n}{N}\mu^*_{-i}(X)+\frac{n}{N}\frac{1}{\binom{N-1}{n-1}\binom{M}{m}}\sum_{\substack{I\subset[N],|I|=n,\\F\subset[M],|F|=m}}\ind(i\in I)\mu_{I,F}(X),
\end{align*}
and hence
\begin{align*}
    \mu_{-i}^{*-j}(X)-\mu^{*-j}(X)&=\frac{n}{N}\Bigg[\mu_{-i}^{*-j}(X)-\frac{1}{\binom{N-1}{n-1}\binom{M-1}{m}}\sum_{\substack{I\subset[N],|I|=n,\\F\subset[M],|F|=m}}\ind(i\in I)\ind(j\notin F)\mu_{I,F}(X)\Bigg]\\
    &=\frac{n}{N}\bigg[\bE_{\substack{I_1: i\notin I_1,\\F: j\notin F}}\big(\hat{\mu}_{I_1,F}(X)|\bX,\bY\big)-\bE_{\substack{I_2: i\in I_2,\\F: j\notin F}}\big(\mu_{I_2,F}(X)|\bX,\bY\big)\bigg],
\end{align*}
where the expectations on the last line are taken over the random subsampling of $I_1,\,I_2\subset[N]$ and $F\subset[M]$, with $I_1$ including sample $i$ and $I_2$ excluding sample $i$. Now we construct a joint distribution over $I_1,\,I_2$ that agree with the marginal distribution for both $I_1$ and $I_2$: suppose that we first randomly subsample $I_0\subset[N]$ with size $|I_0|=n-1$, and then we let $I_1=I_0\cup\{i\}$, $I_2=I_0\cup\{i'\}$ with $i'$ randomly selected from $[N]\backslash I_1$. Under this construction, the marginal distribution of $I_1$ and $I_2$ are the same as random subsampling under the constraints that $i\in I_1$, $i\notin I_2$; in addition, $I_1$ and $I_2$ only differ by one sample. Then we can write
\begin{align*}
    \mu_{-i}^{*-j}(X)-\mu^{*-j}(X)&=\frac{n}{N}\bE_{\substack{I_1,\,I_2,\\F\subset[M]: j\notin F}}\big[\mu_{I_1,F}(X)-\mu_{I_2,F}(X)|\bX,\bY\big]\\
    &=\frac{n}{N}\bE_{\substack{I_0,\,i',\\F\subset[M]: j\notin F}}\big[\mu_{I_0\cup\{i\},F}(X)-\mu_{I_0\cup\{i'\},F}(X)|\bX,\bY\big].
\end{align*}
Hence 
\begin{align}\label{eq:loo_full_mse}
    \bE\|\mu_{-i}^{*-j}(X)-\mu^{*-j}(X)\|_2^2&\leq \frac{n^2}{N^2}\bE_{\bX,\bY,X}\bE_{\substack{I_0,\,i',\\F\subset[M]: j\notin F}}\big\|\mu_{I_0\cup\{i\},F}(X)-\mu_{I_0\cup\{i'\},F}(X)|\bX,\bY\big\|_2^2\\
    &\leq \frac{n^2}{N^2}\frac{\binom{M}{m}}{\binom{M-1}{m}}\bE_{\bX,\bY,X}\bE_{\substack{I_0,\,i',\\F\subset[M]}}\big\|\mu_{I_0\cup\{i\},F}(X)-\mu_{I_0\cup\{i'\},F}(X)|\bX,\bY\big\|_2^2\\
    &\leq \frac{n^2M}{N^2(M-m)}\stb(m,n)\leq \frac{n^2\stb(m,n)}{(1-\gamma)N^2}.
\end{align}
Similarly, we can write 
\begin{align*}
    \mu_{-i}^{*}(X)-\mu^{*}(X)&=\frac{n}{N}\bE_{\substack{I_0,\,i',\\F\subset[M]}}\big[\mu_{I_0\cup\{i\},F}(X)-\mu_{I_0\cup\{i'\},F}(X)|\bX,\bY\big],
\end{align*}
and 
\begin{align}\label{eq:loco_loo_mse}
    \bE\|\mu_{-i}^{*}(X)-\mu^{*}(X)\|_2^2&\leq \frac{n^2\stb(m,n)}{N^2}.
\end{align}
Therefore, combining these with \eqref{eq:epsilon2_bnd0}, we can upper bound $\bE\big|\epsilon_j^{(2)}\big|^2$ as follows:
\begin{equation}
    \begin{split}
        \bE\big|\epsilon_j^{(2)}\big|^2&\leq \frac{2(2-\gamma)}{1-\gamma}\frac{L^2n^2\stb(m,n)}{\sigma_j^2N}.
    \end{split}
\end{equation}
By Assumption \ref{assump:mpsize}, $\bE\big|\epsilon_j^{(2)}\big|^2\rightarrow0$, which implies $|\varepsilon_j^{(2)}|\overset{p}{\rightarrow}0$.
\subsubsection{Bounding $\varepsilon_j^{(3)}$}\label{sec:bnd_err3}
Recall the definition of $\tilde{h}_j(X_i,Y_i;\bX_{\del i,:},\bY_{\del i})$, $h_j(X_i,Y_i;\bX_{\del i,:},\bY_{\del i})$, and $h_j(X_i,Y_i)$ in the beginning of Section \ref{sec:proof}. Then we can write 
\begin{align*}
    \varepsilon^{(3)}_{i,j}=h_j(X_i,Y_i;\bX_{\del i,:},\bY_{\del i})-\bE[h_j(X_i,Y_i;\bX_{\del i,:},\bY_{\del i})|\bX_{\del i,:},\bY_{\del i}]-h_j(X_i,Y_i)+\bE[h_j(X_i,Y_i)].
\end{align*}
Recall our definition of $\bar{h}_j(X_i,Y_i)$ in \eqref{eq:bar_h_j}, and let
$$h_j'(X_i,Y_i;\bX_{\del i,:},\bY_{\del i})=h_j(X_i,Y_i;\bX_{\del i,:},\bY_{\del i})-\bE[h_j(X_i,Y_i;\bX_{\del i,:},\bY_{\del i})|\bX_{\del i,:},\bY_{\del i}].$$ Then we can further decompose $\varepsilon^{(3)}_{i,j}$ as follows:
\begin{align*}
    \varepsilon^{(3)}_{i,j}&=h_j'(X_i,Y_i;\bX_{\del i,:},\bY_{\del i})-\bE[h_j'(X_i,Y_i;\bX_{\del i,:},\bY_{\del i})|X_i,Y_i]\\
    &\hspace{3mm}+\bar{h}_j(X_i,Y_i)-h_j(X_i,Y_i)+\bE[h_j(X_i,Y_i)]-\bE[\bar{h}_j(X_i,Y_i)].
\end{align*}
Denote 
$h_j'(X_i,Y_i;\bX_{\del i,:},\bY_{\del i})-\bE[h_j'(X_i,Y_i;\bX_{\del i,:},\bY_{\del i})|X_i,Y_i]$ by $E_{i}$, and we will first deal with $E_i$ in the following by leveraging a asymptotic linearity result established in \cite{bayle2020cross}, and then show an upper bound for $|\bar{h}_j(X_i,Y_i)-h_j(X_i,Y_i)+\bE[h_j(X_i,Y_i)]-\bE[\bar{h}_j(X_i,Y_i)]|$.

\paragraph{Bounding $E_i$} In fact, $h_j(X_i,Y_i;\bX_{\del i,:},\bY_{\del i})$ can be viewed as a leave-one-out test value, which is a special case of the cross-validation test error in \cite{bayle2020cross}. 
We would like to apply Theorem 2 in \cite{bayle2020cross} with $h_n'(Z_i,Z_{B_j})$ substituted by $h_j'(X_i,Y_i;\bX_{\del i,:},\bY_{\del i})$. Let $(X_{N+1},Y_{N+1})$ be a sample from $\cP$ which is independent from $(\bX,\bY)$, and denote by $(\bX^{\del l}_{\del i,:},\bY^{\del l}_{\del i})$ the $N-1$ training set with sample $i$ excluded, and sample $l$ replaced by $(X_{N+1},Y_{N+1})$. One key quantity in \cite{bayle2020cross}, the loss stability of $h_j(\cdot,\cdot;\cdot,\cdot)$, can be written out as
\begin{align*}
    \gamma_{loss}(h_j)&=\frac{1}{N-1}\sum_{l\neq i}\bE[(h_j'(X_i,Y_i;\bX_{\del i,:},\bY_{\del i})-h_j'(X_i,Y_i;\bX^{\del l}_{\del i,:},\bY^{\del l}_{\del i}))^2]\\
    &\leq \frac{1}{N-1}\sum_{l\neq i}\bE[(h_j(X_i,Y_i;\bX_{\del i,:},\bY_{\del i})-h_j(X_i,Y_i;\bX^{\del l}_{\del i,:},\bY^{\del l}_{\del i}))^2]\\
    &\leq \frac{2L^2}{N-1}\sum_{l\neq i}\bE\big\|\mu^*(X_{i,\backslash j};\bX_{\backslash i,\backslash j},\bY_{\backslash i}) - \mu^*(X_{i,\backslash j};\bX_{\backslash i,\backslash j}^{\backslash l},\bY_{\backslash i}^{\backslash l})\big\|_2^2\\
    &\quad+\bE\big\|\mu^*(X_i;\bX_{\backslash i,:},\bY_{\backslash i}) - \mu^*(X_i;\bX_{\backslash i,:}^{\backslash l},\bY_{\backslash i}^{\backslash l})\big\|_2^2.
\end{align*}
Let $I^{\del l}=\begin{cases}I,&l\notin I\\
\{i\neq l: i\in I\}\cup\{N+1\}, &l\in I\end{cases}$ denote the index set with $l$ replaced by $N+1$, then we have
\begin{align}\label{eq:mu_stb_bnd}
    &\big\|\mu^*(X_{i,\del j};\bX_{\del i,\del j},\bY_{\del i})-\mu^*(X_{i,\del j};\bX^{\del l}_{\del i,\del j},\bY^{\del l}_{\del i})\big\|_2\\
    &\quad\leq\frac{1}{\binom{N-1}{n}\binom{M-1}{m}}\sum_{\substack{I\subset[N],|I|=n,\\F\subset[M],|F|=m}}\ind(i\notin I)\ind(j\notin F)\ind(l\in I)\big\|\mu_{I,F}(X_i)-\mu_{I^{\del l},F}(X_i)\big\|_2\\
    &\quad \leq\frac{1}{\binom{N-1}{n}\binom{M-1}{m}}\sum_{\substack{I\subset[N],|I|=n,\\F\subset[M],|F|=m}}\ind(i\notin I)\ind(l\in I)\big\|\mu_{I,F}(X_i)-\mu_{I^{\del l},F}(X_i)\big\|_2\\
    &\quad \leq\frac{\binom{N-2}{n-1}\binom{M}{m}}{\binom{N-1}{n}\binom{M-1}{m}}\bE_{\substack{I:i\notin I, l\in I\\F:F\subset[M]}}\big\|\mu_{I,F}(X_i)-\mu_{I^{\del l},F}(X_i)\big\|_2\\
    &\quad=\frac{n}{(1-\gamma)(N-1)}\bE_{\substack{I:i\notin I, l\in I\\F:j\notin F}}\big\|\mu_{I,F}(X_i)-\mu_{I^{\del l},F}(X_i)\big\|_2.
\end{align}
Similarly,
\begin{align*}
    &\big\|\mu(X_{i};\bX_{\del i,:},\bY_{\del i})-\mu(X_{i};\bX^{\del l}_{\del i,:},\bY^{\del l}_{\del i})\big\|_2\\
    &\quad\leq \frac{n}{N-1}\bE_{\substack{I:i\notin I, l\in I\\F:F\subset[M]}}\big\|\mu_{I,F}(X_i)-\mu_{I^{\del l},F}(X_i)\big\|_2.
\end{align*}
Therefore, 
\begin{align}\label{eq:loss_stb_bnd}
    \gamma_{loss}(h_j)&\leq \frac{4L^2n^2}{(1-\gamma)^2(N-1)^3}\sum_{l\neq i}\bE\bigg\|\bE_{\substack{I:i\notin I, l\in I\\F:F\subset[M]}}\mu_{I,F}(X_i)-\mu_{I^{\del l},F}(X_i)\bigg\|_2^2\\
    &\leq \frac{4L^2n^2}{(1-\gamma)^2(N-1)^3}\sum_{l\neq i}\bE\bigg(\bE_{\substack{I:i\notin I, l\in I\\F:F\subset[M]}}\big\|\hat{\mu}_{I,F}(X_i)-\hat{\mu}_{I^{\del l},F}(X_i)\big\|_2^2\bigg)\\
    &\leq \frac{4L^2n^2\stb(m,n)}{(1-\gamma)^2(N-1)^2}.
\end{align}
Now we can apply Theorem 2 in \cite{bayle2020cross} to obtain that
\begin{align}\label{eq:var_bnd_linearity}
    \mathrm{Var}\left(\frac{1}{\sigma_j\sqrt{N}}\sum_{i=1}^NE_i\right)\leq \frac{6L^2n^2\stb(m,n)}{\sigma_j^2(N-1)}.
\end{align}
Since $\bE(E_i)=0$ and we have $n^2\stb(m,n)=o\left(\frac{\sigma_j^2}{L^2}N\right)$ by Assumption \ref{assump:mpsize}, \eqref{eq:var_bnd_linearity} suggests $\frac{1}{\sigma_j\sqrt{N}}\sum_{i=1}^NE_i\overset{L_2}{\rightarrow}0$ which then implies $\frac{1}{\sigma_j\sqrt{N}}\sum_{i=1}^NE_i\overset{p}{\rightarrow}0$.

\paragraph{Bounding $\bar{h}_j(X_i,Y_i)-h_j(X_i,Y_i)$}
For bounding $\bar{h}_j(X_i,Y_i)-h_j(X_i,Y_i)$, note that for any $(X,Y)$,
\begin{align*}
    |\bar{h}_j(X,Y)-h_j(X,Y)|&=\bE\left[h_j(X,Y;\bX_{\del i,:},\bY_{\del i})-h_j(X,Y;\bX,\bY)|X, Y\right]\\
    &\leq L\bE[\|\mu^{*-j}_{-i}(X)-\mu^{*-j}(X)\|_2|X]+L\bE[\|\mu^{*}_{-i}(X)-\mu^{*}(X)\|_2|X].
\end{align*}
 Therefore, 
 \begin{align*}
    &\bE\bigg[\frac{1}{\sigma_j\sqrt{N}}\sum_{i=1}^N\left|\bar{h}_j(X_i,Y_i)-h_j(X_i,Y_i)+\bE[h_j(X_i,Y_i)]-\bE[\bar{h}_j(X_i,Y_i)]\right|\bigg]^2\\
    &\quad\leq \frac{1}{\sigma_j^2}\sum_{i=1}^N\bE\bigg(\bar{h}_j(X_i,Y_i)-h_j(X_i,Y_i)\bigg)^2\\
    &\quad\leq \frac{2L^2}{\sigma_j^2}\sum_{i=1}^N\bigg(\bE\|\mu^{*-j}_{-i}(X)-\mu^{*-j}(X)\|_2^2+\bE\|\mu^{*}_{-i}(X)-\mu^{*}(X)\|_2^2\bigg)\\
    &\quad \leq \frac{4L^2n^2\stb(m,n)}{(1-\gamma)\sigma_j^2N}, 
 \end{align*}
  where we applied Jensen's inequality on the second line, and utilized \eqref{eq:loo_full_mse}, \eqref{eq:loco_loo_mse} on the last line. Assumption \ref{assump:mpsize} then further suggests the quantity above converges to zero in probability.
Combining the convergence in probability results for both $$\frac{1}{\sigma_j\sqrt{N}}\sum_{i=1}^N\left|\bar{h}_j(X_i,Y_i)-h_j(X_i,Y_i)+\bE[h_j(X_i,Y_i)]-\bE[\bar{h}_j(X_i,Y_i)]\right|$$ and $\frac{1}{\sigma_j\sqrt{N}}\sum_{i=1}^NE_i$, 
we have $\varepsilon_j^{(3)}\overset{p}{\rightarrow}0$.
\subsubsection{Bounding $\varepsilon_j^{(4)}$}\label{sec:bnd_err4}
Noting that $\varepsilon_j^{(4)}$ also characterizes the deviation of the random minipatch algorithm to the combinatorial average of all minipatches, here we follow similar arguments to those in Section \ref{sec:bnd_err1} to prove the convergence in probability result for $\varepsilon_j^{(4)}$. Recall the definition of $\varepsilon_j^{(4)}$, we have
\begin{equation}\label{eq:err4_decomp}
    \begin{split}
       \varepsilon_j^{(4)} = &\frac{\sqrt{N}}{\sigma_j} \left(\bE[h_j(X,Y;\bX,\bY)|\bX,\bY]-\bE[h_j^{(K)}(X,Y;\bX,\bY)|\bX,\bY]\right) \\
       \leq &\frac{L\sqrt{N}}{\sigma_j}\bE_{X}\left(\|\mu^{*-j}(X) - \mu^{-j}(X)\|_2+\|\mu^*(X) - \mu(X)\|_2|\bX,\bY\right),
    \end{split}
\end{equation}
where the second line is due to the Lipschitz condition (Assumption \ref{assump:Lip}) of the loss function; $\mu^{*-j}(X)$, $\mu^{*}(X)$, $\mu^{-j}(X)$, and $\mu(X)$ are as follows:
\begin{equation*}
    \begin{split}
        \mu(X) &= \frac{1}{K}\sum_{k=1}^K\mu_{I_k,F_k}(X),\quad \mu^*(X) = \frac{1}{K}\sum_{k=1}^K\bE_{I_k,F_k}[\mu_{I_k,F_k}(X)|\bX,\bY,X],\\
        \mu^{-j}(X) &= \frac{1}{K}\sum_{k=1}^K\mu_{\tilde{I}_k,\tilde{F}_k}(X),\quad \mu^{*-j}(X) = \frac{1}{K}\sum_{k=1}^K\bE_{\tilde{I}_k,\tilde{F}_k}[\mu_{\tilde{I}_k,\tilde{F}_k}(X)|\bX,\bY,X],
    \end{split}
\end{equation*}
where $\{(\tilde{I}_k,\tilde{F}_k)\}_{k=1}^K$ is another sequence of random minipatch indices independent from $\{(I_k,F_k)\}_{k=1}^K$, uniformly sampled from $[N]$ and $[M]\backslash j$ with size $(n,m)$.

We start by bounding $\bE_X(\|\mu^*(X) - \mu(X)\|_2|\bX,\bY)$. One thing to note here is that although we only write $(\bX,\bY)$ inside the conditioning argument, we also implicitly condition on $\{(I_k,F_k)\}_{k=1}^K$ when taking the expectation over test data $X$. We follow similar arguments to those for bounding term $\RNum{3}$ in Section \ref{sec:bnd_err1}. We let $$Z_k(X)=\frac{1}{K}\mu_{I_k,F_k}(X) - \frac{1}{K}\bE_{I_k,F_k}(\mu_{I_k,F_k}(X)|\bX,\bY,X)\in \bR^d,$$ and hence $\mu(X) - \mu^{*}(X) = \sum_{k=1}^K Z_k(X)$. Note that
\begin{equation*}
    \begin{split}
        &\bE_{\{I_k,F_k\}_{k=1}^K}\left[\left(\bE_X\Big(\Big\|\sum_{k=1}^KZ_k(X)\Big\|_2|\bX,\bY, \{I_k,F_k\}_{k=1}^K\Big)\right)^2|\bX,\bY\right]\\
        \leq&\bE_X\bE_{\{I_k,F_k\}_{k=1}^K}\left[\Big\|\sum_{k=1}^KZ_k(X)\Big\|_2^2|\bX,\bY\right]\\
        \leq &\frac{CB^2}{K},
    \end{split}
\end{equation*}
where we utilized the independence among $Z_k(X)$ given $X$ and $(\bX,\bY)$, and we have invoked Assumption \ref{assump:bnd_mu} again in the last line. Applying Chebyschev's inequality and plugging in the bound for $\bE_X\|\sum_{k=1}^KZ_k(X)\|_2$ into $\bE_X\left(\|\mu^*(X)-\mu(X)\|_2|\bX,\bY\right)$, we then have
\begin{equation}\label{eq:err4_fullmodel}
    \bP(\bE_X\left(\|\mu(X)-\mu^*(X)\|_2|\bX,\bY\right)>\frac{\epsilon\sigma_j}{L\sqrt{N}})\leq \frac{CB^2L^2N}{\epsilon^2\sigma_j^2K}\rightarrow 0,
\end{equation}
as $K\gg (\frac{B^2L^2N}{\sigma_j^2}+1)\log N$.

    Similarly, we can apply the same argument to bound $\bE_X(\|\mu^{-j}(X) - \mu^{*-j}(X)\|_2|\bX,\bY)$. The argument now hinges on an extension of Assumption \ref{assump:bnd_mu} to the minipatch ensemble without feature $j$: that is, an upper bound for $\bE_X\bE_{\tilde{I}_k, \tilde{F}_k}\|\mu_{\tilde{I}_k,\tilde{F}_k}(X)\|_2^2$ where $\tilde{I}_k,\tilde{F}_k$ are uniformly sampled from $[N]$ and $[M]\backslash j$ with size $n$ and $m$.
    \begin{equation*}
        \begin{split}
            \bE_{\tilde{I}_k, \tilde{F}_k}\|\mu_{\tilde{I}_k,\tilde{F}_k}(X)\|_2^2 = &\frac{1}{\binom{N}{n}\binom{M-1}{m}}\sum_{\substack{I\subset[N]:|I|=n\\F\subset[M]\backslash j, |F| = m}}\|\mu_{I,F}(X)\|_2^2\\
            \leq &\frac{M}{M-m}\frac{1}{\binom{N}{n}\binom{M}{m}}\sum_{\substack{I\subset[N]:|I|=n\\F\subset[M], |F| = m}}\|\mu_{I,F}(X)\|_2^2\\
            = & \frac{M}{M-m}\bE_{I,F}\|\mu_{I,F}(X)\|_2^2 \\
            \leq & \frac{1}{1-\gamma}\bE_{I,F}\|\mu_{I,F}(X)\|_2^2,
        \end{split}
    \end{equation*}
    where the expectation in $\bE_{I,F}\|\mu_{I,F}(X)\|_2^2$ is taken over random minipatch indices uniformaly distributed on $[N]$ and $[M]$. Since $\gamma<1$ is a constant, we still have $\bE_X\bE_{\tilde{I}_k, \tilde{F}_k}\|\mu_{\tilde{I}_k,\tilde{F}_k}(X)\|_2^2\leq CB^2$, hence the above probabilistic bound \eqref{eq:err4_fullmodel} also holds for $\mu^{-j}(X) - \mu^{*-j}(X)$.

Therefore, due to the decomposition in \eqref{eq:err4_decomp}, we have 
\begin{equation*}
    \lim_{N\rightarrow\infty}\bP(|\epsilon_j^{(4)}|>\epsilon)=0.
\end{equation*}

\subsection{Proof of Theorem \ref{thm:coverage_buffer}}\label{sec:proof_buffer}
We prove Theorem \ref{thm:coverage_buffer} by discussing two cases separately: for some fixed constant $c>0$, (i) $\sigma_j\leq \frac{cL\sqrt{\stb(m,n)}n}{\sqrt{N}}\log N$, and (ii) $\sigma_j>\frac{cL\sqrt{\stb(m,n)}n}{\sqrt{N}}\log N$.

\paragraph{Case (i):} First we note that, 
\begin{align*}
    \bP\Big(\Delta_j\notin \hat{\mathbb{C}}_j^{\mathrm{barrier}}\Big) = &\bP\Big(\frac{|\bar{\Delta}_j-\Delta_j|}{\max\{\hat{\sigma}_j/\sqrt{N},\epsilon(N)\}}>z_{\alpha/2}\Big)\\
    \leq &\bP\Big(\frac{|\bar{\Delta}_j-\Delta_j|}{\epsilon(N)}>z_{\alpha/2}\Big).
\end{align*}
By the decomposition in the proof of Theorem \ref{thm:normal_approx}, we also have
\begin{align*}
    \frac{1}{\epsilon(N)}|\bar{\Delta}_j-\Delta_j|\leq&\left|\frac{1}{\epsilon(N)N}\sum_{i=1}^N[h_j(X_i,Y_i) - \mathbb{E}(h_j(X_i,Y_i))]+\sum_{k=1}^4\frac{\sigma_j}{\sqrt{N}\epsilon(N)}\varepsilon_j^{(k)}\right|\\ 
    \leq&\left|\frac{1}{\sigma_j\sqrt{N}}\sum_{i=1}^N[h_j(X_i,Y_i) - \mathbb{E}(h_j(X_i,Y_i))]+\sum_{k=1}^4\frac{\sigma_j}{\sqrt{N}\epsilon(N)}\varepsilon_j^{(k)}\right|,
\end{align*}
where the second line is due to Assumption \ref{assump:buffer} which suggests $\sigma_j\leq \frac{cL\sqrt{\stb(m,n)}n}{\sqrt{N}}\log N\leq \epsilon(N)\sqrt{N}$. 
As has been shown in the proof of Theorem \ref{thm:normal_approx}, 
$$
\frac{1}{\sigma_j\sqrt{N}}\sum_{i=1}^N[h_j(X_i,Y_i) - \mathbb{E}(h_j(X_i,Y_i))]\overset{d}{\rightarrow}\mathcal{N}(0,1);
$$
While for the error terms $\left|\frac{\sigma_j}{\sqrt{N}\epsilon(N)}\varepsilon_j^{(k)}\right|$, note that we can apply the same argument as the proof of Lemma \ref{lem:bnd_err_terms}, except that we replace the factor $\frac{\sigma_j}{\sqrt{N}}$ by $\epsilon(N)$. That is, instead of requiring $n^2\stb(m,n)=o\left(\frac{\sigma_j^2}{L^2}N\right)$ in Assumption \ref{assump:mpsize}, we need $n^2\stb(m,n)=o\left(\frac{\epsilon^2(N)N^2}{L^2}\right)$, which is automatically satisfied since $\epsilon(N)\geq \frac{cL\sqrt{\stb(m,n)}n}{N}\log N$; instead of requiring $K\gg \left(\frac{L^2B^2N}{\sigma_j^{2}}+1\right)\log N$, we need $K\gg \left(\frac{L^2B^2}{\epsilon^2(N)}+1\right)\log N$, which can also be implied by Assumption \ref{assump:buffer}. Therefore, for case (i), we have $\left|\frac{\sigma_j}{\sqrt{N}\epsilon(N)}\varepsilon_j^{(k)}\right|\overset{p}{\rightarrow}0$, and hence
\begin{align*}
    &\liminf_{N\rightarrow \infty}\bP(\Delta_j\in \hat{\mathbb{C}}_j^{\mathrm{barrier}})\\
    \geq &\lim_{N\rightarrow \infty}\bP\left(\left|\frac{1}{\sigma_j\sqrt{N}}\sum_{i=1}^N[h_j(X_i,Y_i) - \mathbb{E}(h_j(X_i,Y_i))]+\sum_{k=1}^4\frac{\sigma_j}{\sqrt{N}\epsilon(N)}\varepsilon_j^{(k)}\right|\leq z_{\alpha/2}\right)\\
    = &1-\alpha.
\end{align*}

\paragraph{Case (ii):} While for the second case, note that $\sigma_j>\frac{cL\sqrt{\stb(m,n)}n}{\sqrt{N}}\log N$ implies that $n^2\stb(m,n)=o(\frac{\sigma_j^2}{L^2}N)$, the minipatch size condition in Assumption \ref{assump:mpsize}; In addition, the requirement on $K$ in Assumption \ref{assump:buffer} also implies $$K\gg \frac{B^2}{\stb(m,n)}\frac{N^2}{n^2\log N} + \log N >  \left(\frac{c^2L^2B^2N}{\sigma_j^2}+1\right)\log N,$$
which is Assumption \ref{assump:mpnumber}. 
Since $\hat{\mathbb{C}}_j^{\mathrm{barrier}}$ has the same center but larger width than $\hat{\mathbb{C}}_j$, when Assumption \ref{assump:Lip}-\ref{assump:mpnumber} hold, $\liminf_{N\rightarrow \infty}\bP(\Delta_j\in \hat{\mathbb{C}}_j^{\mathrm{barrier}})\geq 1-\alpha$ is a direct consequence of Corollary \ref{cor:coverage-width}. 
\subsection{Proof of Theorems \ref{thm:datadriven_coverage} and \ref{thm:datadriven_coverage_barrier}}\label{sec:proof_datadriven}
\begin{proof}[Proof of Theorem \ref{thm:datadriven_coverage}]
    Recall that in Definition \ref{def:mn_oracle}, $(m^{\mathrm{oracle}},\,n^{\mathrm{oracle}})$ are the minimizer of the population LOO residuals.
    As a review of related notations, $(m^*,\,n^*)$ in Definition \ref{def:delta_LOO} satisfies $(m^*,\,n^*) = \argmin_{(m,\,n)\in S}\LOO(m,\,n)$ and hence depends on the training data; $(\hat{m},\,\hat{n})$ given by Algorithm \ref{algo:MPsize_tuning} minimizes $\widehat{\LOO}(m,\,n)$, where $\widehat{\LOO}(m,\,n) = \frac{1}{N}\sum_i\err(Y_i, \mu^{(m,\,n)}_{-i}(X_i))$, with $\mu_{-i}^{(m,\,n)}(\cdot)$ being the ensemble predictor constructed from random minipatches excluding sample $i$. 
    
    Our main proof idea is to show that the selected minipatch sizes $(\hat{m},\,\hat{n})$ by Algorithm \ref{algo:MPsize_tuning} converges to its population counterpart $(m^{\mathrm{oracle}},\,n^{\mathrm{oracle}})$ in probability, and hence the proof reduces to showing the coverage of LOCO-MP with minipatch sizes $(m^{\mathrm{oracle}},\,n^{\mathrm{oracle}})$ that do not depend on the data. Recall the minipatch sizes notations we reviewed above and Definition \ref{def:delta_LOO}, one can show that
    \begin{equation*}
        \begin{split}
            \bP\left((\hat{m},\,\hat{n})\neq (m^{\mathrm{oracle}},\,n^{\mathrm{oracle}})\right)\leq &\bP\left((\hat{m},\,\hat{n})\neq (m^*,\,n^*)\right) + \bP\left((m^*,\,n^*)\neq (m^{\mathrm{oracle}},\,n^{\mathrm{oracle}})\right)\\
            \leq &\bP\left(\exists (m,\,n)\in S, \big|\widehat{\LOO}(m,\,n) - \LOO(m,\,n)\big|\geq \frac{\delta_{\LOO}(S)}{2}\right)\\
            &+ \bP\left(\exists (m,\,n)\in S, \big|\LOO(m,\,n) - \bE\LOO(m,\,n)\big|\geq \frac{\delta_{\LOO}(S)}{2}\right)\\
            \leq &\sum_{(m,\,n)\in S}\bP\left(\big|\widehat{\LOO}(m,\,n) - \LOO(m,\,n)\big|\geq \frac{\delta_{\LOO}(S)}{2}\right)\\
            &+ \sum_{(m,\,n)\in S}\bP\left(\big|\LOO(m,\,n) - \bE\LOO(m,\,n)\big|\geq \frac{\delta_{\LOO}(S)}{2}\right).
        \end{split}
    \end{equation*}
    In the following, we will bound the probability above by bounding the two terms $\big|\widehat{\LOO}(m,\,n) - \LOO(m,\,n)\big|$ and $\big|\LOO(m,\,n) - \bE\LOO(m,\,n)\big|$, separately. First, we apply Assumption \ref{assump:Lip} to obtain the following:
    \begin{equation*}
        \begin{split}
            \big|\widehat{\LOO}(m,\,n) - \LOO(m,\,n)\big|\leq &\frac{L}{N}\sum_{i=1}^N\|\mu_{-i}^*(X_i) - \mu_{-i}(X_i)\|_2\\
            =&\frac{\sigma_j}{\sqrt{N}}[\RNum{2}(m,n) + \RNum{4}(m,n)],
        \end{split}
    \end{equation*}
    where $\RNum{2}(m,n)$, $\RNum{4}(m,n)$ are as defined in Section \ref{sec:bnd_err1}, when minipatch sizes $(m, n)$ are in use for the ensembled predictors. Since Assumptions \ref{assump:bnd_mu}, \ref{assump:mpsize} hold for all candidate minipatch sizes $(m,\,n)\in S$, we can follow the argument in Section \ref{sec:bnd_err1} and obtain the following:
    \begin{equation*}
        \begin{split}
            &\bP\left(\big|\widehat{\LOO}(m,\,n) - \LOO(m,\,n)\big|>\frac{\delta_{\LOO}(S)}{2}\right)\\
            \leq &\frac{1}{K} + \exp\{\log N -K\min\{\frac{(1-\gamma)^2}{2}, \frac{(1-\gamma)^4\delta_{\LOO}^2(S)}{128B^2L^2}\}\}+\frac{16B^2L^2}{\delta_{\LOO}^2(S)(1-\gamma)^2K},
        \end{split}
    \end{equation*}
    where $\gamma\in (0, 1)$ is a constant in Assumption \ref{assump:mpsize}, $B$ is the average prediction bound of MP predictors in Assumption \ref{assump:bnd_mu}, and $L$ is the Lipstchitz constant of the error function in Assumption \ref{assump:Lip}. Now recall that Assumption \ref{assump:delta_LOO_bnd} suggests $\delta_{\LOO}(S)\geq c'LB$ for some constant $c'>0$, the number of candidate MP sizes $s$ is bounded, and Assumption \ref{assump:mpnumber} suggests $K\gg \log N$. Therefore, $\sum_{(m,\,n)\in S}\bP\left(\big|\widehat{\LOO}(m,\,n) - \LOO(m,\,n)\big|\geq \frac{\delta_{\LOO}(S)}{2}\right)\rightarrow 0$.

    While for $\sum_{(m,\,n)\in S}\bP\left(\big|\LOO(m,\,n) - \bE\LOO(m,\,n)\big|\geq \frac{\delta_{\LOO}(S)}{2}\right)$, we can apply the Markov's inequality to get
    \begin{equation}\label{eq:LOO_concentration}
        \begin{split}
            &\sum_{(m,\,n)\in S}\bP\left(\big|\LOO(m,\,n) - \bE\LOO(m,\,n)\big|\geq \frac{\delta_{\LOO}(S)}{2}\right)\\
            \leq&\frac{4}{\delta_{\LOO}^2(S)}\sum_{(m\,n)\in S}\mathrm{Var}\left(\LOO(m,\,n;\,\bX,\,\bY)\right),
        \end{split}
    \end{equation}
    where we write $\LOO(m,\,n)$ as $\LOO(m,\,n;\,\bX,\,\bY)$ to emphasize its dependence on the training data $(\bX,\,\bY)$. Let $(\bX',\,\bY')$ be an i.i.d. copy of the training data $(\bX,\,\bY)$. Since $\LOO(m,\,n;\,\bX,\,\bY)$ is a function of the i.i.d. data $\{(X_i,\,Y_i)\}_{i=1}^N$, by the Efron-Stein inequality \cite[see, e.g., Proposition 1 in][]{boucheron2005moment}, we can bound the variance of $\LOO(m,\,n;\,\bX,\,\bY)$ as follow:
    \begin{equation}\label{eq:LOO_var}
        \mathrm{Var}\left(\LOO(m,\,n;\,\bX,\,\bY)\right) \leq \frac{1}{2}\sum_{l=1}^N\bE\left[\LOO(m,\,n;\,\bX^{\backslash l},\,\bY^{\backslash l}) - \LOO(m,\,n;\,\bX,\,\bY)\right]^2,
    \end{equation}
    where $(\bX^{\backslash l}\,\bY^{\backslash l})$ is obtained by substituting $(X_l,\,Y_l)$ in the original training data $(\bX,\,\bY)$ by $(X_l',Y_l')$. Furthermore, for any $1\leq l\leq N$, we can decompose $\LOO(m,\,n;\,\bX^{\backslash l},\,\bY^{\backslash l}) - \LOO(m,\,n;\,\bX,\,\bY)$ into two error terms:
    \begin{equation*}
        \begin{split}
            &|\LOO(m,\,n;\,\bX^{\backslash l},\,\bY^{\backslash l}) - \LOO(m,\,n;\,\bX,\,\bY)|\\
            \leq&\bigg|\frac{1}{N}\sum_{i\neq l}\err[Y_i,\,\mu^{(m,\,n)}(X_i;\,\bX^{\backslash l}_{\backslash i,:},\,\bY^{\backslash l}_{\backslash i})] - \frac{1}{N}\sum_{i\neq l}\err[Y_i,\,\mu^{(m,\,n)}(X_i;\,\bX_{\backslash i,:},\,\bY_{\backslash i})]\bigg|\\
            &+\frac{1}{N}\big|\err[Y_l',\,\mu^{(m,\,n)}(X_l';\,\bX_{\backslash l,:},\,\bY_{\backslash l})] - \err[Y_l,\,\mu^{(m,\,n)}(X_l;\,\bX_{\backslash l,:},\,\bY_{\backslash l})]\big|\\
            \leq &\frac{L}{N}\sum_{i\neq l}\|\mu^{(m,\,n)}(X_i;\,\bX^{\backslash l}_{\backslash i,:},\,\bY^{\backslash l}_{\backslash i}) - \mu^{(m,\,n)}(X_i;\,\bX_{\backslash i,:},\,\bY_{\backslash i})\|_2\\
            &+\frac{1}{N}\big|\err[Y_l',\,\mu^{(m,\,n)}(X_l';\,\bX_{\backslash l,:},\,\bY_{\backslash l})] - \err[Y_l,\,\mu^{(m,\,n)}(X_l;\,\bX_{\backslash l,:},\,\bY_{\backslash l})]\big|.
        \end{split}
    \end{equation*}
    Therefore, 
    \begin{equation}\label{eq:LOO_err_decomp}
        \begin{split}
            &\bE\left[\LOO(m,\,n;\,\bX^{\backslash l},\,\bY^{\backslash l}) - \LOO(m,\,n;\,\bX,\,\bY)\right]^2\\
            \leq&2L^2\bE\|\mu^{(m,\,n)}(X_i;\,\bX_{\backslash i,:}^{\backslash l}, \bY_{\backslash i}^{\backslash l}) - \mu^{(m,\,n)}(X_i;\,\bX_{\backslash i,:}, \bY_{\backslash i})\|_2^2\\
            &+\frac{2}{N^2}\bE\left[\err[Y_{l}',\,\mu^{(m,\,n)}(X_l';\,\bX_{\backslash l,:},\bY_{\backslash l}]-\err[Y_{l},\,\mu^{(m,\,n)}(X_l;\,\bX_{\backslash l,:},\bY_{\backslash l}]\right]^2\\
           \leq& 2L^2\bE\|\mu^{(m,\,n)}(X_i;\,\bX_{\backslash i,:}^{\backslash l},\, \bY_{\backslash i}^{\backslash l}) - \mu^{(m,\,n)}(X_i;\,\bX_{\backslash i,:},\, \bY_{\backslash i})\|_2^2\\
            &+\frac{4}{N^2}\bE\left[\mathrm{Var}\left[\err[Y_{l},\,\mu^{(m,\,n)}(X_l;\,\bX_{\backslash l,:},\bY_{\backslash l}]|\bX_{\backslash l,:},\bY_{\backslash l}\right]\right].
        \end{split}
    \end{equation}
    Following the same argument as in Section \ref{sec:bnd_err3}, we have
    \begin{equation}\label{eq:LOO_err_term1}
        \begin{split}
            &\bE\|\mu^{(m,\,n)}(X_i;\,\bX_{\backslash i,:}^{\backslash l},\, \bY_{\backslash i}^{\backslash l}) - \mu^{(m,\,n)}(X_i;\,\bX_{\backslash i,:},\, \bY_{\backslash i})\|_2^2\\
            \leq&\frac{n^2}{(N-1)^2}\bE\left(\bE_{\substack{I:i\notin I, l\in I\\F:F\subset[M]}}\big\|\mu_{I,F}(X_i)-\mu_{I^{\del l},F}(X_i)\big\|_2\right)^2\\
            \leq & \frac{n^2}{(N-1)^2}\bE\left(\big\|\mu_{I,F}(X_i)-\mu_{I^{\del l},F}(X_i)\big\|_2^2\right)\\
            =&\frac{n^2}{(N-1)^2}\stb(m,n).
        \end{split}
    \end{equation}
    In addition, since for any r.v. $(Z_1,\,Z_2)$ and function $g(\cdot)$, $\bE[\mathrm{Var}(f(Z_1,\,Z_2)|Z_2)] = \bE f^2(Z_1,\,Z_2) - \bE[\bE f(Z_1,\,Z_2)|Z_2]^2\leq  \bE f^2(Z_1,\,Z_2) - [\bE f(Z_1,\,Z_2)]^2 = \mathrm{Var}(f(Z_1,\,Z_2))$, we have 
    \begin{equation}\label{eq:LOO_err_term2}
        \begin{split}
            &\bE\left[\mathrm{Var}\left[\err[Y_{l},\,\mu^{(m,\,n)}(X_l;\,\bX_{\backslash l,:},\bY_{\backslash l}]|\bX_{\backslash l,:},\bY_{\backslash l}\right]\right]\\
            \leq &\mathrm{Var}\left[\err[Y_{l},\,\mu^{(m,\,n)}(X_l;\,\bX_{\backslash l,:},\bY_{\backslash l}]\right]\\
            \leq &\sigma_{\LOO}^2(S)
        \end{split}
    \end{equation} 
    Therefore, plugging in \eqref{eq:LOO_err_term1}, \eqref{eq:LOO_err_term2}, and \eqref{eq:LOO_err_decomp} into \eqref{eq:LOO_var}, we have
    \begin{equation*}
        \begin{split}
            \mathrm{Var}\left(\LOO(m,\,n;\,\bX,\,\bY)\right) \leq \frac{L^2n^2N}{(N-1)^2}\stb(m,n) + \frac{2\sigma_{\LOO}^2(S)}{N}.
        \end{split}
    \end{equation*}
    Hence \eqref{eq:LOO_concentration} implies that
    \begin{equation*}
        \begin{split}
            &\sum_{(m,\,n)\in S}\bP\left(\big|\LOO(m,\,n) - \bE\LOO(m,\,n)\big|\geq \frac{\delta_{\LOO}(S)}{2}\right)\\
            \leq&\frac{16L^2}{\delta_{\LOO}^2(S)N}\sum_{(m,\,n)\in S}n^2\stb(m,n) + \frac{8s\sigma_{\LOO}^2(S)}{\delta_{\LOO}^2(S)N},
        \end{split}
    \end{equation*}
    where we have also applied the fact that $N\leq 2(N-1)$. Note that Assumption \ref{assump:delta_LOO_bnd} implies $\sigma_{\LOO}^2(S)\geq cL^2B^2$, $\delta_{\LOO}(S)\geq c\sigma_{\LOO}(S)$ for some constant $c>0$, and 
    \begin{equation*}
    \begin{split}
        \sigma_j^2 = &\mathrm{Var}(h_j(X,Y))\\
        \leq &\bE h_j^2(X,Y;\bX,\bY)\\
        \leq &L^2\bE\left\|\mu^{(m^{\mathrm{oracle}},n^{\mathrm{oracle}})}_{\backslash j}(X_{\backslash j};\bX_{\backslash j}, \bY) - \mu^{(m^{\mathrm{oracle}},n^{\mathrm{oracle}})}(X;\bX, \bY)\right\|_2^2\\
        \leq &2L^2\sum_{(m,n)\in S}\bE\left\|\mu^{(m,n)}_{\backslash j}(X_{\backslash j};\bX_{\backslash j}, \bY)\right\|_2^2 + \bE\left\|\mu^{(m,n)}(X;\bX, \bY)\right\|_2^2\\
        \leq&2L^2\sum_{(m,n)\in S}\left(\frac{1}{1-\gamma}\bE\bE_{I,F}\ind(j\notin F)\|\mu_{I,F}(X_F)\|_2^2 + \bE\bE_{I,F}\|\mu_{I,F}(X_F)\|_2^2\right)\\
        \leq&CsL^2B^2,
    \end{split}
\end{equation*}
and $s$ is bounded. Hence one can show that $$\sum_{(m,\,n)\in S}\bP\left(\big|\LOO(m,\,n) - \bE\LOO(m,\,n)\big|\geq \frac{\delta_{\LOO}(S)}{2}\right)\leq \frac{16L^2}{\sigma_j^2N}\sum_{(m,\,n)\in S}n^2\stb(m,n) + \frac{8s}{N}\rightarrow 0,$$
    where Assumption \ref{assump:mpsize} is invoked for all $(m,\,n)\in S$ to show the above convergence to zero result. Therefore, we have now proved that $\bP\big((\hat{m},\,\hat{n}) \neq (m^{\mathrm{oracle}},\,n^{\mathrm{oracle}})\big)\rightarrow 0$.

    Let $\tilde{\mathbb{C}}_j$ be the confidence interval given by Algorithm \ref{algo:loco} with minipatch sizes $(m^{\mathrm{oracle}},\,n^{\mathrm{oracle}})$, and let $\tilde{\Delta}_j$ be the inference target in \eqref{eq:target_inference} with $(m^{\mathrm{oracle}},\,n^{\mathrm{oracle}})$. Since $(m^{\mathrm{oracle}},\,n^{\mathrm{oracle}})$ are independent from the training data, we can invoke Corollary \ref{cor:coverage-width} to obtain $\lim_{N\rightarrow\infty}\bP(\tilde{\Delta}_j\in \tilde{\mathbb{C}}_j)=1-\alpha$. On the other hand, since $\bP\big((\hat{m},\,\hat{n}) \neq (m^{\mathrm{oracle}},\,n^{\mathrm{oracle}})\big)\rightarrow 0$, $\bP(\tilde{\Delta}_j \neq \Delta_j) + \bP(\tilde{\mathbb{C}}_j\neq \mathbb{C}_j)\rightarrow 0$. Therefore,
    \begin{equation*}
        \begin{split}
            &\lim_{N\rightarrow\infty}\bP(\Delta_j\notin \mathbb{C}_j) \leq \lim_{N\rightarrow\infty}\bP(\tilde{\Delta}_j\notin \tilde{\mathbb{C}}_j) + \bP(\tilde{\Delta}_j \neq \Delta_j) + \bP(\tilde{\mathbb{C}}_j\neq \mathbb{C}_j) = \alpha,\\
            &\lim_{N\rightarrow\infty}\bP(\Delta_j\in \mathbb{C}_j) \leq \lim_{N\rightarrow\infty}\bP(\tilde{\Delta}_j\in \tilde{\mathbb{C}}_j) + \bP(\tilde{\Delta}_j \neq \Delta_j) + \bP(\tilde{\mathbb{C}}_j\neq \mathbb{C}_j) \leq 1-\alpha,
        \end{split}
    \end{equation*}
    implying $\lim_{N\rightarrow\infty}\bP(\Delta_j\in \mathbb{C}_j) = 1-\alpha$. The proof of Theorem \ref{thm:datadriven_coverage} is now complete.
\end{proof}

%%%-------------------%-------------%-------------------%-------------------%
\begin{proof}[Proof of Theorem \ref{thm:datadriven_coverage_barrier}]
To prove the valid coverage of LOCO-MP with data-driven selection of the minipatch size under relaxed assumptions and variance barriers, we need to show an upper bound for the errors induced by the dependency between minipatch sizes and the training data. Therefore, we cannot build our argument on the proof of Theorem \ref{thm:datadriven_coverage}, but need a completely new route. We also need some different definitions for some key quantities, which will only be used within this proof:
redefine the function $h_j(X,Y) = \bE_{\bX,\bY}[h_j(X,Y;\bX,\bY)]$ (same as $h_j'(X,\,Y)$ in Assumption \ref{assump:datadriven_h_third_moment}), where $$h_j(X,Y;\bX,\bY) = h_j^{(m^*(\bX,\bY),n^*(\bX,\bY))}(X,Y;\bX,\bY)$$ is defined as in \eqref{eq:h_j_full_datadriven}. The variance parameter $\sigma_j^2 = \mathrm{Var}(h_j(X,Y))$ is also redefined for this new $h_j(X,Y)$; $\tilde{h}_j(X_i,Y_i;\bX_{\backslash i,:},\bY_{\backslash i}) = h_j(X_i,Y_i;\bX_{\backslash i,:},\bY_{\backslash i}) - h_j(X_i,Y_i)$. Let $h_j^{(K,m,n)}(X,Y;\bX,\bY) = \err(Y,\mu^{(m,n)}_{\backslash j}(X_{\backslash j};\bX,\bY)-\err(Y,\mu^{(m,n)}(X;\bX,\bY))$ be the feature importance score of minipatch ensembles with $K$ minipatches and minipatch size $(m,\,n)$. Then we can write our inference target as
$$
\Delta_j = \bE[h_j^{(\hat{m},\,\hat{n})}(X,Y;\bX,\bY)|\bX,\bY],
$$ 
the importance of $j$ for predicting a new sample when the minipatch ensemble predictor with $K$ minipatches is trained and tuned on $(\bX,\bY)$, where the expectation is taken over the new test point $(X,\,Y)$. Here we also let $\tilde{\Delta}_j = \bE[h_j(X,Y;\bX,\bY)|\bX,\bY] = \bE[h_j^{(m^*,\,n^*)}(X,Y;\bX,\bY)|\bX,\bY]$, which serve as an intermediate target close to $\Delta_j$.

Now we start our proof for Theorem \ref{thm:datadriven_coverage_barrier}, where we need to account for the dependency between the training data and the selected MP sizes. The key intuition lies that such dependency is evenly distributed on all samples, and hence the selection of MP sizes depends on each sample in a negligible way. Our proof structure is similar to the proof of Theorems \ref{thm:normal_approx} and \ref{thm:coverage_buffer}, while special attention is needed when controlling each residual error term. First of all, let us consider a sub-sequence of the sample size $\mathcal{N} = \big\{N: \epsilon(N) <\frac{\sigma_j(N)}{\sqrt{N}}\big\}$, where we wrote $\sigma_j$ as $\sigma_j(N)$ to emphasize its dependence on $N$. For any sample size quantity $N\in \mathcal{N}$, Assumption \ref{assump:buffer_datadriven} implies that
\begin{equation*}
    \sum_{l=1}^sn_l^2\stb(m,n) < \frac{\sigma_j^2(N)N}{c^2L^2\log^2N},
\end{equation*}
and $K\gg \big(\frac{L^2B^2N}{\sigma_j^2} + 1\big)\log N$.
By Assumption \ref{assump:datadriven_h_third_moment}, this newly defined $\sigma_j^2$ for this proof only is at most of the same order as the original $\sigma_j^2$ in Section \ref{sec:datadriven_theory}. These then imply that Assumption \ref{assump:mpnumber} holds and Assumption \ref{assump:mpsize} holds for all $(m,\,n)\in S$ when $N\in\mathcal{N}$. Therefore, we can invoke Theorem \ref{thm:datadriven_coverage} on $\mathcal{N}$ to show that $\lim_{N\rightarrow \infty, N\in \mathcal{N}}\bP(\Delta_j\in \hat{\mathbb{C}}_j) = 1-\alpha$; since the width of $\mathbb{C}_j^{\mathrm{barrier}}$ is greater than or equal to the width of $\mathbb{C}_j$, we also have $\lim\inf_{N\rightarrow \infty, N\in \mathcal{N}}\bP(\Delta_j\in \mathbb{C}_j^{\mathrm{barrier}}) \geq 1-\alpha$. Now it remains to prove Theorem \ref{thm:datadriven_coverage_barrier} for $N\in \mathcal{N}^c$; that is, we will assume $\epsilon(N)\geq \frac{\sigma_j}{\sqrt{N}}$ in the following proof.

Similar to the proof of Theorem \ref{thm:normal_approx}, we can decompose the deviation of each feature occlusion score to the inference target as follows:
    \begin{equation*}
    \begin{split}
        &\hat{\Delta}_j(X_i,Y_i) - \Delta_j \\
        =&\hat{h}_j^{(\hat{m},\hat{n})}(X_i,Y_i;\bX_{\backslash i,:},\bY_{\backslash i}) - \bE[h_j(X,Y;\bX,\bY)|\bX,\bY] + \tilde{\Delta}_j - \Delta_j\\
        =&\hat{h}_j^{(\hat{m},\hat{n})}(X_i,Y_i;\bX_{\backslash i,:},\bY_{\backslash i}) - h_j^{(\hat{m},\hat{n})}(X_i,Y_i;\bX_{\backslash i,:},\bY_{\backslash i})\\
        &+h_j^{(\hat{m},\hat{n})}(X_i,Y_i;\bX_{\backslash i,:},\bY_{\backslash i}) - h_j^{(m^*_{-i},n^*_{-i})}(X_i,Y_i;\bX_{\backslash i,:},\bY_{\backslash i})\\
        &+\tilde{h}_j(X_i,Y_i;\bX_{\backslash i,:},\bY_{\backslash i}) - \bE[\tilde{h}_j(X_i,Y_i;\bX_{\backslash i,:},\bY_{\backslash i})|\bX_{\backslash i,:},\bY_{\backslash i}]\\
        &+\bE[h_j(X_i,Y_i;\bX_{\backslash i,:},\bY_{\backslash i})|\bX_{\backslash i,:},\bY_{\backslash i}] - \tilde{\Delta}_j\\
        &+\tilde{\Delta}_j - \Delta_j\\
        & + h_j(X_i,Y_i) - \bE[h_j(X_i,Y_i)].
    \end{split}
    \end{equation*}
    Now let
    \begin{equation}\label{eq:datadriven_decomp_errdef}
        \begin{split}
            \varepsilon_{i,j}^{(1)} = &\hat{h}_j^{(\hat{m},\hat{n})}(X_i,Y_i;\bX_{\backslash i,:},\bY_{\backslash i}) - h_j^{(\hat{m},\hat{n})}(X_i,Y_i;\bX_{\backslash i,:},\bY_{\backslash i}),\\
             \varepsilon_{i,j}^{(2)} = &h_j^{(\hat{m},\hat{n})}(X_i,Y_i;\bX_{\backslash i,:},\bY_{\backslash i}) - h_j^{(m^*_{-i},n^*_{-i})}(X_i,Y_i;\bX_{\backslash i,:},\bY_{\backslash i}),\\
             \varepsilon_{i,j}^{(3)}= & \tilde{h}_j(X_i,Y_i;\bX_{\backslash i,:},\bY_{\backslash i}) - \bE[\tilde{h}_j(X_i,Y_i;\bX_{\backslash i,:},\bY_{\backslash i})|\bX_{\backslash i,:},\bY_{\backslash i}],\\
             \varepsilon_{i,j}^{(4)}= &\bE[h_j(X_i,Y_i;\bX_{\backslash i,:},\bY_{\backslash i})|\bX_{\backslash i,:},\bY_{\backslash i}] - \tilde{\Delta}_j,
        \end{split}
    \end{equation}
    and define $\varepsilon_j^{(k)} = \bigg(\max\big\{\hat{\sigma}_j\sqrt{N}, \epsilon(N)N\big\}\bigg)^{-1}\sum_{i=1}^N\varepsilon_{i,j}^{(k)}$, for $k=1,\dots,4$, In addition, let $\varepsilon_{j}^{(5)}= \min\big\{\sqrt{N}/\hat{\sigma}_j, \epsilon^{-1}(N)\big\}(\tilde{\Delta}_j - \Delta_j)$. Then we can write
    \begin{equation*}
    \begin{split}
        &\frac{\min\big\{\sqrt{N}/\hat{\sigma}_j, \epsilon^{-1}(N)\big\}}{N}\sum_{i=1}^N\big(\hat{\Delta}_j(X_i,Y_i) - \Delta_j\big) \\
       = &\frac{\min\big\{\sqrt{N}/\hat{\sigma}_j, \epsilon^{-1}(N)\big\}}{N}\sum_{i=1}^N\big(h_j(X_i,Y_i) - \bE[h_j(X_i,Y_i)]\big) + \sum_{k=1}^5\varepsilon_j^{(k)},
    \end{split}
    \end{equation*}
    and hence for any $\delta>0$
    \begin{equation*}
    \begin{split}
        &\mathbb{P}\big(\Delta_j\in \hat{\mathbb{C}}_j^{\mathrm{barrier}}\big)\\
        = &1-\bP\bigg(\bigg|\frac{\min\big\{\sqrt{N}/\hat{\sigma}_j, \epsilon^{-1}(N)\big\}}{N}\sum_{i=1}^N\big(\hat{\Delta}_j(X_i,Y_i) - \Delta_j\big)\bigg|> z_{\alpha/2}\bigg)\\
        \geq &1 - \bP\bigg(\bigg|\frac{\min\big\{\sqrt{N}/\hat{\sigma}_j, \epsilon^{-1}(N)\big\}}{N}\sum_{i=1}^N\big(h_j(X_i,Y_i) - \bE[h_j(X_i,Y_i)]\big)\bigg|>z_{\alpha/2}-\delta\bigg)\\
        &-\bP\bigg(\sum_{k=1}^5\varepsilon_j^{(k)}> \delta\bigg) \\
        \geq &1 - \bP\bigg(\bigg|\frac{1}{\sqrt{N}\sigma_j}\sum_{i=1}^N\big(h_j(X_i,Y_i) - \bE[h_j(X_i,Y_i)]\big)\bigg|>z_{\alpha/2}-\delta\bigg)\\
        &-\bP\bigg(\sum_{k=1}^5\varepsilon_j^{(k)}> \delta\bigg).
    \end{split}
    \end{equation*}
    Assumption \ref{assump:datadriven_h_third_moment} implies that the Liapounov's condition holds for $h_j(X_i,\,Y_i)$, and hence we can apply the central limit theorem to show that $\lim_{N\rightarrow \infty, N\in \mathcal{N}^c}\bP\bigg(\bigg|\frac{1}{\sqrt{N}\sigma_j}\sum_{i=1}^N\big(h_j(X_i,\,Y_i) - \bE[h_j(X_i,\,Y_i)]\big)\bigg|>z_{\alpha/2}-\delta\bigg) = 2(1-\Phi(z_{\alpha/2}-\delta))$, where $\Phi(\cdot)$ is the distribution function of a standard Gaussian distribution. Therefore, for any $\delta>0$
    \begin{equation}\label{eq:datadriven_coverage_lwbnd}
        \begin{split}
            &\lim\inf_{N\rightarrow \infty,\,N\in \mathcal{N}^c}\mathbb{P}\big(\Delta_j\in \hat{\mathbb{C}}_j^{\mathrm{barrier}}\big)\\
            \geq &2\Phi(z_{\alpha/2}-\delta) - 1 - \bP\bigg(\sum_{k=1}^5\varepsilon_j^{(k)}> \delta\bigg).
        \end{split}    
    \end{equation} 
    In the following, we will show that $\varepsilon_j^{(k)}\overset{p}{\rightarrow} 0$ for $1\leq k\leq 5$; then by the continuity of the distribution function $\Phi(\cdot)$ and \eqref{eq:datadriven_coverage_lwbnd}, we will have $\lim\inf_{N\rightarrow \infty,\,N\in \mathcal{N}^c}\mathbb{P}\big(\Delta_j\in \hat{\mathbb{C}}_j^{\mathrm{barrier}}\big)\geq 2\Phi(z_{\alpha/2}) - 1 = 1-\alpha$. This result combined with our previous proof for $\lim_{N\rightarrow \infty, N\in \mathcal{N}}\bP(\Delta_j\in \hat{\mathbb{C}}_j) = 1-\alpha$ completes our proof. The remaining proof is devoted to proving $\varepsilon_j^{(k)}\overset{p}{\rightarrow} 0$ for $1\leq k\leq 5$, when $\epsilon(N)\geq \frac{\sigma_j}{\sqrt{N}}$.

    \paragraph{Bounding $\varepsilon_j^{(1)}$.} By definition, $\varepsilon_j^{(1)}$ captures the deviation of feature importance score computed from the random minipatch algorithm from its population version (infinite $K$), when the minipatch size is chosen through data-driven tuning. We can then simply bound it by the maximum of $\varepsilon_j^{(1)}(m,n)$ over all candidate minipatch sizes:
    $$
    \varepsilon_j^{(1)} \leq \frac{\sigma_j}{\epsilon(N)\sqrt{N}}\max_{1\leq l\leq s}\varepsilon_j^{(1)}(m_l,n_l), 
    $$
    where $\varepsilon_j^{(1)}(m_l,n_l)= \frac{1}{\sigma_j\sqrt{N}}\sum_{i=1}^N\hat{h}_j^{(m_l,n_l)}(X_i,Y_i,\bX_{\backslash i,:}\bY_{\backslash i}) - h_j^{(m_l,n_l)}(X_i,Y_i,\bX_{\backslash i,:}\bY_{\backslash i})$. We can then decompose each $\varepsilon_j^{(1)}(m_l,n_l)$ into four terms as in the proof in Section \ref{sec:bnd_err1} bound them accordingly. In particular, let $\RNum{1}(m_l,\,n_l)$, $\RNum{2}(m_l,\,n_l)$, $\RNum{3}(m_l,\,n_l)$, $\RNum{4}(m_l,\,n_l)$ be defined as in Section \ref{sec:bnd_err1}, when minipatch sizes $(m_l,\,n_l)$ are considered. Since Assumption \ref{assump:bnd_mu} holds for all $(m_l,\,n_l)$, and Assumption \ref{assump:buffer_datadriven} requires $\frac{m}{M},\,\frac{n}{N}\leq \gamma$, $K\gg (\frac{B^2L^2}{\epsilon^2(N)} + 1)\log N$, we can then follow the argument in Section \ref{sec:bnd_err1} and obtain the following: for any $\delta>0$,
    \begin{equation*}
        \begin{split}
            &\bP\bigg(\max_{1\leq l\leq s}|\RNum{1}(m_l,n_l)|>\frac{\delta\epsilon(N)\sqrt{N}}{\sigma_j}\bigg)\\
            \leq &\sum_{l=1}^s\bP\bigg(|\RNum{1}(m_l,n_l)|>\frac{\delta\epsilon(N)\sqrt{N}}{\sigma_j}\bigg)\\
            \leq &\frac{s}{K} + 
            s\exp\big\{-\frac{1}{2}\log N\big\}\ll \frac{s}{\log N},\\
            &\bP(\max_{1\leq l\leq s}|\RNum{2}(m_l,n_l)|>\frac{\delta\epsilon(N)\sqrt{N}}{\sigma_j})\ll \frac{s}{\log N},\\
            &\bP(\max_{1\leq l\leq s}|\RNum{3}(m_l,n_l)|>\epsilon)\leq \frac{sB^2L^2}{\delta^2\epsilon^2(N)(1-\gamma)^4K},\\
            &\bP(\max_{1\leq l\leq s}|\RNum{4}(m_l,n_l)|>\epsilon)\leq \frac{sB^2L^2}{\delta^2\epsilon^2(N)(1-\gamma)^2K}.
        \end{split}
    \end{equation*}
    Since $s$ is bounded, the four probabilities above also converge to zero as $N$ tends to infinity. Therefore, we have $\varepsilon_j^{(1)}\overset{p}{\rightarrow} 0$.
    \paragraph{Bounding $\varepsilon_j^{(2)}$.} $\varepsilon_j^{(2)}$ captures the change in the LOCO-LOO feature importance score if, during each LOO procedure, the minipatch sizes are selected without access to each left-out sample $i$, using the deterministic minipatch ensembles. As we will show in the following, the main steps are to bound the probabilities that (i) the minipatch sizes $(\hat{m},\hat{n})$ selected by Algorithm \ref{algo:MPsize_tuning} differ from $(m^*, n^*)$ selected using the deterministic MP ensemble; and (ii) the minipatch sizes $(m^*, n^*)$ selected using the full data set differ from $(m^*_{-i}, n^*_{-i})$ selected without sample $i$. 
    
    For any $\epsilon>0$, we first decompose the tail probability of $\varepsilon_j^{(2)}$ as follows:
    \begin{equation}\label{eq:datadriven_err2_decomp}
        \begin{split}
            \bP(|\varepsilon_j^{(2)}|>\epsilon)\leq \bP(\hat{m} \neq m^*\text{ or }\hat{n} \neq n^*) + \bP(|\varepsilon_j^{(2)}|>\epsilon, \hat{m} = m^*,\,\hat{n} =n^*),
        \end{split}
    \end{equation}
    where $(m^*,n^*) = \argmin_{(m,n)\in S}\LOO(m,n)$ as defined in Definition \ref{def:delta_LOO}, with $\LOO(m,n) = \frac{1}{N}\sum_i\err(Y_i,\mu^{*(m,n)}(X_i;\bX_{\backslash i,:},\bY_{\backslash i}))$; $(\hat{m},\hat{n}) = \argmin_{(m,n)\in S}\widehat{\LOO}(m,n)$ for $\widehat{\LOO}(m,n) = $ \\$\frac{1}{N}\sum_i\err(Y_i,\mu^{(m,n)}(X_i;\bX_{\backslash i,:},\bY_{\backslash i})$. $\mu^{(m,n)}(\cdot;\cdot,\cdot)$ is the ensemble predictor constructed from $K$ random minipatches, while $\mu^{*(m,n)}(\cdot;\cdot,\cdot)$ is constructed from the deterministic minipatch ensemble. Recall $\delta_{\LOO}(S)$ in Definition \ref{def:delta_LOO}. Then if $|\LOO(m,n) - \widehat{\LOO}(m,n)|\leq \frac{1}{2}\delta_{\LOO}(S)$ for all $(m, n) \in S$, we have 
    \begin{equation*}
        \begin{split}
            \widehat{\LOO}(m^*, n^*) \leq &\LOO(m^*, n^*) + \frac{1}{2}\delta_{\LOO}(S)\\
            = & \min_{(m,n)\neq (m^*,n^*)}\LOO(m, n) - \frac{1}{2}\delta_{\LOO}(S)\\
            \leq &\min_{(m,n)\neq (m^*,n^*)}\widehat{\LOO}(m, n),
        \end{split}
    \end{equation*}
    which then implies $(m^*, n^*) = (m, n)$. One can then bound $\bP(\hat{m} \neq m^*\text{ or }\hat{n} \neq n^*)$ as follows:
    \begin{equation*}
        \begin{split}
            \bP(\hat{m} \neq m^*\text{ or }\hat{n} \neq n^*)\leq & \bP\left(\exists (m,n)\in S, |\LOO(m,n) - \widehat{\LOO}(m,n)|> \frac{1}{2}\delta_{\LOO}(S)\right).
        \end{split}
    \end{equation*}
Note that 
\begin{equation*}
    \begin{split}
        |\LOO(m,n) - \widehat{\LOO}(m,n)|\leq &\frac{L}{N}\sum_{i=1}^N\|\mu^{*(m,n)}(X_i;\bX_{\backslash i,:},\bY_{\backslash i})-\mu^{(m,n)}(X_i;\bX_{\backslash i,:},\bY_{\backslash i})\|\\
        =&\frac{\sigma_j}{\sqrt{N}}(\RNum{2}(m,n)+\RNum{4}(m,n)),
    \end{split}
\end{equation*} 
where $\RNum{2}(m,n)+\RNum{4}(m,n)$ are defined as earlier when bounding $\varepsilon_j^{(1)}$. Therefore, utilizing the arguments in Section \ref{sec:bnd_err1} for bounding $\RNum{2}(m,n)$ and $\RNum{4}(m,n)$, we have
\begin{equation}\label{eq:MPdiff_finiteK_prob}
    \begin{split}
        &\bP(\hat{m} \neq m^*\text{ or }\hat{n} \neq n^*) \\
        \leq & \bP\left(\exists (m,n)\in S, |\LOO(m,n) - \widehat{\LOO}(m,n)|> \frac{1}{2}\delta_{\LOO}(S)\right)\\
        \leq & \sum_{(m,n)\in S}\bP\left(\RNum{2}(m,n) > \frac{\delta_{\LOO}(S)\sqrt{N}}{4\sigma_j}\right) + \sum_{(m,n)\in S}\bP\left(\RNum{4}(m,n) > \frac{\delta_{\LOO}(S)\sqrt{N}}{4\sigma_j}\right)\\
        \leq & \frac{s}{K} + s\exp\left\{\log N - K\min\left\{\frac{(1-\gamma)^2}{2},\frac{(1-\gamma)^4\delta_{\LOO}^2(S)}{128B^2L^2}\right\}\right\} + \frac{sB^2L^2}{\delta_{\LOO}^2(S)(1-\gamma)^2K}\\
        \leq& \frac{s}{K} + s\exp\left\{-\frac{1}{2}\log N\right\} + \frac{1}{\log N}\rightarrow 0,
    \end{split}
\end{equation}
where we have applied the lower bound for $\delta_{\LOO}(S)$ in Assumption \ref{assump:delta_LOO_bnd}, bounded $s$ assumption, and the requirement for $K$ in Assumption \ref{assump:mpnumber} in the last line. With this result combined with \eqref{eq:datadriven_err2_decomp}, we now only need to show that $\lim_{N\rightarrow \infty}\bP(|\varepsilon_j^{(2)}|>\epsilon, \hat{m}=m^*,\hat{n}=n^*)=0$.

When the event $\{\hat{m}=m^*,\hat{n}=n^*\}$ holds true, we can then write $\varepsilon_{i,j}^{(2)} = h_j^{(m^*,n^*)}(X_i,Y_i;\bX_{\backslash i,:}, \bY_{\backslash i}) - h_j^{(m^*_{-i},n^*_{-i})}(X_i,Y_i;\bX_{\backslash i,:}, \bY_{\backslash i})$. Hence, 
\begin{equation*}
    \begin{split}
        \varepsilon_j^{(2)}\ind(\hat{m}=m^*,\hat{n}=n^*) \leq &\frac{1}{\epsilon(N)N}\sum_{i=1}^N\varepsilon_{i,j}^{(2)}\\
        =&\frac{1}{\epsilon(N)N}\sum_{i=1}^N\big\{\ind(m^*\neq m_{-i}^*\text{ or }n^*\neq n^*_{-i})\\
        &\cdot[h_j^{(m^*,n^*)}(X_i,Y_i;\bX_{\backslash i,:}, \bY_{\backslash i}) - h_j^{(m^*_{-i},n^*_{-i})}(X_i,Y_i;\bX_{\backslash i,:}, \bY_{\backslash i})]\big\}\\
        \leq &\frac{1}{\epsilon(N)}\sqrt{\frac{1}{N}\sum_{i=1}^N\ind(m^*\neq m_{-i}^*\text{ or }n^*\neq n^*_{-i})}\\
        &\cdot \sqrt{\frac{1}{N}\sum_{i=1}^N \delta_{\LOCO}^2(j,S;X_i,Y_i,\bX_{\backslash i,:},\bY_{\backslash i})}\\
        \leq &\frac{\delta_{\LOCO}(j,S)}{\epsilon(N)}\sqrt{\frac{1}{N}\sum_{i=1}^N\ind(m^*\neq m_{-i}^*,\text{ or }n^*\neq n^*_{-i})},
    \end{split}
\end{equation*}
where we have utilized the Definition \ref{def:delta_LOCO} in the third and fourth inequalities. 
Therefore, we can reduce the problem to bounding the probability that $(m^*_{-i}, n^*_{-i})\neq (m^*, n^*)$:
\begin{equation}\label{eq:epsilon2_datadriven_bnd}
    \begin{split}
        &\bP(|\varepsilon_j^{(2)}|>\epsilon, \hat{m}=m^*,\hat{n}=n^*)\\
    \leq & \bP\left(\frac{1}{N}\sum_{i=1}^N\ind(m^*\neq m_{-i}^*\text{ or }n^*\neq n^*_{-i})>\frac{\epsilon^2\epsilon^2(N)}{\delta^2_{\LOCO}(j,S)}\right)\\
    \leq &\frac{\delta^2_{\LOCO}(j,S)}{\epsilon^2\epsilon^2(N)}\bP(m^*\neq m_{-i}^*\text{ or }n^*\neq n^*_{-i}).
    \end{split}
\end{equation}
Similar to the previous argument for bounding the probability that $(\hat{m},\hat{n})\neq (m^*,n^*)$, since $(m^*_{-i},n^*_{-i}) = \argmin_{(m,n)\in S}\LOO_{-i}(m,n)$, we have
\begin{equation*}
    \bP(m^*\neq m_{-i}^*\text{ or }n^*\neq n^*_{-i})\leq \sum_{(m,n)\in S}\bP\left(|\LOO(m,n) - \LOO_{-i}(m,n)| \geq \frac{1}{2}\delta_{\LOO}(S)\right),
\end{equation*}
where $\LOO_{-i}(m,n) = \frac{1}{N-1}\sum_{l\neq i}\err(Y_l,\mu^{*(m,n)}(X_i;\bX_{\backslash\{i,l\},:},\bY_{\backslash\{i,l\}}))$. Let $\widetilde{\LOO}_{-i}(m,n) = \frac{1}{N-1}\sum_{l\neq i}\err(Y_l,\mu^{*(m,n)}(X_l;\bX_{\backslash i,:}, \bY_{\backslash i}))$. We can then bound $|\LOO(m,n) - \LOO_{-i}(m,n)|$ as follows:
\begin{equation*}
    \begin{split}
        &|\LOO(m,n) - \LOO_{-i}(m,n)|\\
        \leq & |\LOO(m,n) - \widetilde{\LOO}_{-i}(m,n)| + |\widetilde{\LOO}_{-i}(m,n) - \LOO_{-i}(m,n)|\\
        \leq & \frac{1}{N}\left[\frac{1}{N-1}\sum_{l\neq i}\err(Y_l,\mu^{*(m,n)}(X_l;\bX_{\backslash l,:},\bY_{\backslash l})) - \err(Y_i, \mu^{*(m,n)}(X_i;\bX_{\backslash i,:},\bY_{\backslash i}))\right]\\
        & + \frac{L}{N-1}\sum_{l\neq i}\|\mu^{*(m,n)}(X_l;\bX_{\backslash l,:},\bY_{\backslash l}) - \mu^{*(m,n)}(X_l;\bX_{\backslash \{l,i\},:},\bY_{\backslash \{l,i\}})\|_2,
    \end{split}
\end{equation*}
which then implies
\begin{equation*}
    \begin{split}
        &\bE|\LOO(m,n) - \LOO_{-i}(m,n)|^2\\
        \leq&\frac{2}{N^2}\bE\left[\err(Y_l,\mu^{*(m,n)}(X_l;\bX_{\backslash l,:},\bY_{\backslash l})) - \err(Y_i, \mu^{*(m,n)}(X_i;\bX_{\backslash i,:},\bY_{\backslash i}))\right]^2\\
        & + 2L^2\bE\|\mu^{*(m,n)}(X_l;\bX_{\backslash l,:},\bY_{\backslash l}) - \mu^{*(m,n)}(X_l;\bX_{\backslash \{l,i\},:},\bY_{\backslash \{l,i\}})\|_2^2\\
        \leq & \frac{8}{N^2}\sigma_{\LOO}^2(S) + 2L^2\frac{n^2\stb(m,n)}{(N-1)^2}.
    \end{split}
\end{equation*}
In the last line above, we applied Definition \ref{def:sigma_LOO} to bound the first term, while the bound for the second term is due to the same argument as those for bounding $\bE\|\mu^*_{-i}(X)- \mu^*(X)\|_2^2$ in Section \ref{sec:bnd_err2}. Therefore, 
\begin{equation}\label{eq:MPdiff_LOO_prob}
    \begin{split}
        &\bP(m^*\neq m_{-i}^*\text{ or }n^*\neq n^*_{-i})\\
        \leq &4\delta_{\LOO}^{-2}(S)\sum_{(m,n)\in S}\bE|\LOO(m,n) - \LOO_{-i}(m,n)|^2\\
        \leq &4\delta_{\LOO}^{-2}(S)\left(\frac{8s}{N^2}\sigma_{\LOO}^2(S) + 2L^2\frac{\sum_{(m,n)\in S}n^2\stb(m,n)}{(N-1)^2}\right)\\
        \leq&\frac{Cs}{N^2} + \frac{C\sum_{(m,n)\in S}n^2\stb(m,n)}{B^2N^2},
    \end{split}
\end{equation}
where in the last line, we have applied Assumption \ref{assump:delta_LOO_bnd} which suggests $\frac{\sigma_{\LOO}^2(S)}{\delta_{\LOO}^2(S)}$, $\frac{L^2B^2}{\delta_{\LOO}^2(S)}$ being bounded. Furthermore, \eqref{eq:MPdiff_LOO_prob} and \eqref{eq:epsilon2_datadriven_bnd}, combined together, lead to the following:
\begin{equation*}
    \begin{split}
        &\bP(|\varepsilon_j^{(2)}|>\epsilon, \hat{m}=m^*,\hat{n}=n^*)\\
        \leq&\frac{Cs\delta^2_{\LOCO}(j,S)}{\epsilon^2\epsilon^2(N)N^2} + \frac{C\delta^2_{\LOCO}(j,S)\sum_{(m,n)\in S}n^2\stb(m,n)}{\epsilon^2\epsilon^2(N)B^2N^2}\\
        \leq&\frac{Cs\delta_{\LOCO}^2(j,S)}{\epsilon^2\epsilon^2(N)N^2} + \frac{C\delta_{\LOCO}^2(j,S)}{\epsilon^2L^2B^2\log^2N},
    \end{split}
\end{equation*}
where in the last line, we have also applied Assumption \ref{assump:buffer_datadriven} to bound the second term. In addition, recall that we have assumed $\epsilon(N)\geq \sigma_j/\sqrt{N}$ before bounding $\varepsilon_j^{(1)}$. Then by the fact that $s$ is bounded and Assumption \ref{assump:delta_LOCO_bnd}, the first term in the last line above satisfies
$\frac{32s\delta_{\LOCO}^2(j,S)}{\epsilon^2\epsilon^2(N)N^2}\leq\frac{C}{\epsilon^2 N}\rightarrow 0$ for any $\epsilon>0$. Similarly, the second term $\frac{8\delta_{\LOCO}^2(j,S)}{c^2\epsilon^2L^2B^2\log^2N}\leq \frac{C\sigma_j^2}{\epsilon^2L^2B^2\log^2N}$. Noting that 
\begin{equation}\label{eq:sigma_j_bnd}
    \begin{split}
        \sigma_j^2 = &\mathrm{Var}(h_j(X,Y))\\
        \leq &\bE h_j^2(X,Y;\bX,\bY)\\
        \leq &L^2\bE\left\|\mu^{*(m^*(\bX,\bY),n^*(\bX,\bY))}_{\backslash j}(X_{\backslash j};\bX_{\backslash j}, \bY) - \mu^{*(m^*(\bX,\bY),n^*(\bX,\bY))}(X;\bX, \bY)\right\|_2^2\\
        \leq &2L^2\sum_{(m,n)\in S}\bE\left\|\mu^{*(m,n)}_{\backslash j}(X_{\backslash j};\bX_{\backslash j}, \bY)\right\|_2^2 + \bE\left\|\mu^{*(m,n)}(X;\bX, \bY)\right\|_2^2\\
        \leq&2L^2\sum_{(m,n)\in S}\left(\frac{1}{1-\gamma}\bE\bE_{I,F}\ind(j\notin F)\|\mu_{I,F}(X_F)\|_2^2 + \bE\bE_{I,F}\|\mu_{I,F}(X_F)\|_2^2\right)\\
        \leq&CL^2B^2,
    \end{split}
\end{equation}
where we have applied Assumption \ref{assump:bnd_mu} in the last line. Therefore, $\lim_{N\rightarrow \infty}\bP(|\varepsilon_j^{(2)}|>\epsilon, \hat{m}=m^*,\hat{n}=n^*)=0$, which further implies $\varepsilon_j^{(2)}\overset{p}{\rightarrow}0$. 

\paragraph{Bounding $\varepsilon_j^{(3)}$.} The proof is similar to Section \ref{sec:bnd_err3}, while for a different definition of the feature importance function $h_j(\cdot,\cdot;\cdot,\cdot)$. Recall that $\varepsilon_j^{(3)} = \big(\max\big\{\hat{\sigma}_j\sqrt{N}, \epsilon(N)N\big\}\big)^{-1}\sum_{i=1}^N\varepsilon_{i,j}^{(3)}$, where 
$$
\varepsilon_{i,j}^{(3)} = h_j(X_i,Y_i;\bX_{\backslash i,:},\bY_{\backslash i}) - \bE[h_j(X_i,Y_i;\bX_{\backslash i,:},\bY_{\backslash i})|\bX_{\backslash i,:},\bY_{\backslash i}] - h_j(X_i,Y_i) + \bE(h_j(X_i,Y_i)).
$$
Similar to the proof in Section \ref{sec:bnd_err3}, we let 
$$
h_j'(X_i,Y_i;\bX_{\backslash i,:},\bY_{\backslash i}) = h_j(X_i,Y_i;\bX_{\backslash i,:},\bY_{\backslash i}) - \bE[h_j'(X_i,Y_i;\bX_{\backslash i,:},\bY_{\backslash i})|\bX_{\backslash i}, \bY_{\backslash i}],
$$
and 
\begin{equation}\label{eq:bar_h_def_datadriven}
    \bar{h}_j(X_i,Y_i) = \bE_{\bX_{\backslash i,:},\bY_{\backslash i}}[h_j(X_i,Y_i;\bX_{\backslash i,:}, \bY_{\backslash i})|X_i,Y_i]
\end{equation}
With these new notations, we can then rewrite $\varepsilon_{i,j}^{(3)}$ as follows:
\begin{equation}\label{eq:epsilon_3_decomp}
\begin{split}
    \varepsilon_{i,j}^{(3)} = &h_j'(X_i,Y_i;\bX_{\backslash i,:},\bY_{\backslash i}) - \bE[h_j'(X_i,Y_i;\bX_{\backslash i,:},\bY_{\backslash i})|\bX_{\backslash i,:},\bY_{\backslash i}]\\
    &+ \bar{h}_j(X_i,Y_i) - h_j(X_i,Y_i) + \bE(h_j(X_i,Y_i)) - \bE(\bar{h}_j(X_i,Y_i))\\
    =:&E_i + \bar{h}_j(X_i,Y_i) - h_j(X_i,Y_i) + \bE(h_j(X_i,Y_i)) - \bE(\bar{h}_j(X_i,Y_i)).
\end{split}
\end{equation}
The arguments by far are the same as Section \ref{sec:bnd_err3}; however, the main difference lies that the minipatch sizes in our new function $h_j(X_i,Y_i;\bX_{\backslash i,:},\bY_{\backslash i})$ also depend on the corresponding training data $(\bX_{\backslash i,:},\bY_{\backslash i})$. In the following, we will first bound $E_i$ by utilizing a result in \cite{bayle2020cross} and showing that the stability condition in \cite{bayle2020cross} holds for our function $h_j(X_i,Y_i;\bX_{\backslash i,:},\bY_{\backslash i})$; we will then show a bound for $\bar{h}_j(X_i,Y_i) - h_j(X_i,Y_i) + \bE(h_j(X_i,Y_i)) - \bE(\bar{h}_j(X_i,Y_i))$.

Let's first focus on the loss stability of the new $h_j(X_i,Y_i;\bX_{\backslash i,:},\bY_{\backslash i})$:
\begin{equation}\label{eq:loss_stb_bnd_main}
    \begin{split}
        \gamma_{loss}(h_j) = &\frac{1}{N-1}\sum_{l\neq i}\bE\bigg[\big(h_j'\big(X_i,Y_i;\bX_{\backslash i,:},\bY_{\backslash i}\big) - h_j'\big(X_i,Y_i;\bX^{\backslash l}_{\backslash i,:}, \bY^{\backslash l}_{\backslash i}\big)\big)^2\bigg]\\
        \leq &\bE\bigg[\big(h_j\big(X_i,Y_i;\bX_{\backslash i,:},\bY_{\backslash i}\big) - h_j\big(X_i,Y_i;\bX^{\backslash l}_{\backslash i,:}, \bY^{\backslash l}_{\backslash i}\big)\big)^2\bigg]\\
        \leq &2\bE\bigg[\big(h_j^{(m_{-i}^*,n_{-i}^*)}\big(X_i,Y_i;\bX_{\backslash i,:},\bY_{\backslash i}\big) - h_j^{(m_{-i}^*,n_{-i}^*)}\big(X_i,Y_i;\bX^{\backslash l}_{\backslash i,:}, \bY^{\backslash l}_{\backslash i}\big)\big)^2\bigg]\\
        &+2\bE\bigg[\big(h_j^{(m_{-i}^*,n_{-i}^*)}\big(X_i,Y_i;\bX^{\backslash l}_{\backslash i,:}, \bY^{\backslash l}_{\backslash i}\big) - h_j^{(m_{-i}^{\backslash l *},n_{-i}^{\backslash l*})}\big(X_i,Y_i;\bX^{\backslash l}_{\backslash i,:}, \bY^{\backslash l}_{\backslash i}\big)\big)^2\bigg]
    \end{split}
\end{equation}
where the second line is due to the fact that variance is always bounded by the second moment; the third line decomposes the difference between $h_j\big(X_i,Y_i;\bX_{\backslash i,:},\bY_{\backslash i}\big)$ and $h_j\big(X_i,Y_i;\bX^{\backslash l}_{\backslash i,:}, \bY^{\backslash l}_{\backslash i}\big)$ into the effects of different minipatch sizes and different training data, where $m_{-i}^{\backslash l *} = m^*(\bX_{\backslash i,:}^{\backslash l},\bY_{\backslash i,:}^{\backslash l})$, $n_{-i}^{\backslash l*} = n^*(\bX_{\backslash i,:}^{\backslash l},\bY_{\backslash i,:}^{\backslash l})$. For the first term in the decomposition above, we have
\begin{equation}\label{eq:loss_stb_bnd1}
    \begin{split}
        &2\bE\bigg[\big(h_j^{(m_{-i}^*,n_{-i}^*)}\big(X_i,Y_i;\bX_{\backslash i,:},\bY_{\backslash i}\big) - h_j^{(m_{-i}^*,n_{-i}^*)}\big(X_i,Y_i;\bX^{\backslash l}_{\backslash i,:}, \bY^{\backslash l}_{\backslash i}\big)\big)^2\bigg]\\
        \leq&2\sum_{(m,n)\in S}\bE\bigg[\big(h_j^{(m,n)}\big(X_i,Y_i;\bX_{\backslash i,:},\bY_{\backslash i}\big) - h_j^{(m,n)}\big(X_i,Y_i;\bX^{\backslash l}_{\backslash i,:}, \bY^{\backslash l}_{\backslash i}\big)\big)^2\bigg]\\
        \leq &\frac{8L^2\sum_{(m,n)\in S}n^2\stb(m,n)}{(1-\gamma)^2(N-1)^2},
    \end{split}
\end{equation}
where we have applied the bound for $\bE\bigg[\big(h_j^{(m,n)}\big(X_i,Y_i;\bX_{\backslash i,:},\bY_{\backslash i}\big) - h_j^{(m,n)}\big(X_i,Y_i;\bX^{\backslash l}_{\backslash i,:}, \bY^{\backslash l}_{\backslash i}\big)\big)^2\bigg]$ in \eqref{eq:loss_stb_bnd}. While for the second term in \eqref{eq:loss_stb_bnd_main}, we have
\begin{equation}\label{eq:loss_stb_bnd2}
    \begin{split}
        &2\bE\bigg[\big(h_j^{(m_{-i}^*,n_{-i}^*)}\big(X_i,Y_i;\bX^{\backslash l}_{\backslash i,:}, \bY^{\backslash l}_{\backslash i}\big) - h_j^{(m_{-i}^{\backslash l *},n_{-i}^{\backslash l*})}\big(X_i,Y_i;\bX^{\backslash l}_{\backslash i,:}, \bY^{\backslash l}_{\backslash i}\big)\big)^2\bigg]\\
        \leq&2\bE\bigg[\big(\delta_{\LOCO}^2\big(j,S;X_i,Y_i,\bX^{\backslash l}_{\backslash i,:}, \bY^{\backslash l}_{\backslash i}\big)\ind{(m_{-i}^*,n_{-i}^*)\neq (m_{-i}^{\backslash l*},n_{-i}^{\backslash l*})}\bigg]\\
        \leq&2\sqrt{\bE\bigg[\delta_{\LOCO}^4\big(j,S;X_i,Y_i,\bX^{\backslash l}_{\backslash i,:}, \bY^{\backslash l}_{\backslash i}\big)\bigg]}\sqrt{\bP\Big((m_{-i}^*,n_{-i}^*)\neq \Big(m_{-i}^{\backslash l*},n_{-i}^{\backslash l*}\Big)\Big)}\\
        \leq &2\delta_{\LOCO}^2(j,S)\sqrt{\bP\Big((m_{-i}^*,n_{-i}^*)\neq \Big(m_{-i}^{\backslash l*},n_{-i}^{\backslash l*}\Big)\Big)}.
    \end{split}
\end{equation}
where the second and last line is due to Definition \ref{def:delta_LOCO}, and the third line utilizes the Cauchy-Schwarz inequality. To bound the probability of $(m_{-i}^*,n_{-i}^*)\neq \Big(m_{-i}^{\backslash l*},n_{-i}^{\backslash l*}\Big)$, let 
\begin{equation}\label{eq:LOO_stb_def}
    \begin{split}
        \LOO_{-i}(m,n) = &\frac{1}{N-1}\sum_{i'\neq i}\err(Y_{i'},\mu^{*(m,n)}(X_{i'};\bX_{\backslash (i,i'),:}, \bY_{\backslash (i,i')})),\\
        \LOO_{-i}^{\backslash l}(m,n) = &\frac{1}{N-1}\Big(\sum_{i'\neq i,l}\err\big(Y_{i'},\mu^{*(m,n)}\big(X_{i'};\bX^{\backslash l}_{\backslash (i,i'),:}, \bY^{\backslash l}_{\backslash (i,i')}\big)\big)\\
        &\hspace{1.2cm}+ \err(Y_{N+1},\mu^{*(m,n)}(X_{N+1};\bX_{\backslash (i,l),:}, \bY_{\backslash (i,l)}))\Big).
    \end{split}
\end{equation}
We can then show the following:
\begin{equation}\label{eq:loss_stb_prob}
    \begin{split}
        &\bP\Big((m_{-i}^*,n_{-i}^*)\neq \Big(m_{-i}^{\backslash l*},n_{-i}^{\backslash l*}\Big)\Big)\\
        \leq &\sum_{(m,n)\in S}\bP\Bigg(\Big|\LOO_{-i}(m,n)-\LOO_{-i}^{\backslash l}(m,n)\Big|>\frac{\delta_{\LOO}}{2}\Bigg)\\
        \leq &\frac{16}{\delta^{4}_{\LOO}(S)}\sum_{(m,n)\in S}\bE\Big|\LOO_{-i}(m,n)-\LOO_{-i}^{\backslash l}(m,n)\Big|^4.
    \end{split}
\end{equation}
By the definition of $\LOO_{-i}(m,n)$ and $\LOO_{-i}^{\backslash l}(m,n)$ in \eqref{eq:LOO_stb_def}, we can write
\begin{equation*}
    \begin{split}
        &\Big|\LOO_{-i}(m,n)-\LOO_{-i}^{\backslash l}(m,n)\Big|\\
        =&\Bigg|\frac{1}{N-1}\Bigg(\sum_{i'\neq i,l}\Big(\err\big(Y_{i'},\mu^{*(m,n)}\big(X_{i'};\bX_{\backslash (i,i'),:}, \bY_{\backslash (i,i')}\big)\big)\\
        &\hspace{2.4cm}-\err\big(Y_{i'},\mu^{*(m,n)}\big(X_{i'};\bX^{\backslash l}_{\backslash (i,i'),:}, \bY^{\backslash l}_{\backslash (i,i')}\big)\big)\Big)\\
        &\hspace{1.2cm}+ \err\big(Y_{l},\mu^{*(m,n)}\big(X_{l};\bX_{\backslash (i,l),:}, \bY_{\backslash (i,l)}\big)\big)\\
        &\hspace{1.2cm}- \err\big(Y_{N+1},\mu^{*(m,n)}\big(X_{N+1};\bX_{\backslash (i,l),:}, \bY_{\backslash (i,l)}\big)\big)\Bigg)\Bigg|\\
        \leq&\Bigg|\frac{1}{N-1}\Bigg(\sum_{i'\neq i,l}\frac{Ln}{N-2}\bE_{\substack{I:i,i'\notin I,l\in I\\ F:F\subset [M]}}\|\mu_{I,F}(X_{i'}) - \mu_{I^{\backslash l},F}(X_{i'})\|_2\Big)\\
        &\hspace{1.2cm}+ \err\big(Y_{l},\mu^{*(m,n)}\big(X_{l};\bX_{\backslash (i,l),:}, \bY_{\backslash (i,l)}\big)\big)\\
        &\hspace{1.2cm}- \err\big(Y_{N+1},\mu^{*(m,n)}\big(X_{N+1};\bX_{\backslash (i,l),:}, \bY_{\backslash (i,l)}\big)\big)\Bigg)\Bigg|,
    \end{split}
\end{equation*}
where the last inequality uses similar arguments to those in \eqref{eq:mu_stb_bnd} in Section \ref{sec:bnd_err3}. By Jensen's inequality, we have
\begin{equation}\label{eq:LOO_stb_4moment}
    \begin{split}
        &\bE\Big|\LOO_{-i}(m,n)-\LOO_{-i}^{\backslash l}(m,n)\Big|^4\\
        \leq&8\bE\Bigg|\frac{1}{N-1}\sum_{i'\neq i,l}\frac{Ln}{N-2}\bE_{\substack{I:i,i'\notin I,l\in I\\ F:F\subset [M]}}\|\mu_{I,F}(X_{i'}) - \mu_{I^{\backslash l},F}(X_{i'})\|_2\Bigg|^4\\
        &+ \frac{8}{(N-1)^4}\bE\Big|\err\big(Y_{l},\mu^{*(m,n)}\big(X_{l};\bX_{\backslash (i,l),:}, \bY_{\backslash (i,l)}\big)\big)\\
        &\hspace{2.2cm}- \err\big(Y_{N+1},\mu^{*(m,n)}\big(X_{N+1};\bX_{\backslash (i,l),:}, \bY_{\backslash (i,l)}\big)\big)\Big|^4\\
        \leq&\frac{8}{N-1}\sum_{i'\neq i,l}\frac{L^4n^4}{(N-2)^4}\bE\Big|\bE_{\substack{I:i,i'\notin I,l\in I\\ F:F\subset [M]}}\|\mu_{I,F}(X_{i'}) - \mu_{I^{\backslash l},F}(X_{i'})\|_2\Big|^4\\
        &+ \frac{8}{(N-1)^4}\bE\Big|\err\big(Y_{l},\mu^{*(m,n)}\big(X_{l};\bX_{\backslash (i,l),:}, \bY_{\backslash (i,l)}\big)\big)\\
        &\hspace{2.2cm}- \err\big(Y_{N+1},\mu^{*(m,n)}\big(X_{N+1};\bX_{\backslash (i,l),:}, \bY_{\backslash (i,l)}\big)\big)\Big|^4\\
        \leq &\frac{8}{N-1}\sum_{i'\neq i,l}\frac{L^4n^4}{(N-2)^4}\bE\bE_{\substack{I:i,i'\notin I,l\in I\\ F:F\subset [M]}}\|\mu_{I,F}(X_{i'}) - \mu_{I^{\backslash l},F}(X_{i'})\|_2^4\\
        &+  \frac{128}{(N-1)^4}\kappa_{\LOO}(S)\sigma_{\LOO}^4(S)\\
        \leq&\frac{CL^4n^4\stb^2(m,n)}{N^4}+\frac{C\sigma_{\LOO}^4(S)}{N^4},
    \end{split}
\end{equation}
where the second inequality applies the following Lemma \ref{lem:diff_4moment} on $\err\big(Y_{l},\mu^{*(m,n)}\big(X_{l};\bX_{\backslash (i,l),:}, \bY_{\backslash (i,l)}\big)\big)$ and $\err\big(Y_{N+1},\mu^{*(m,n)}\big(X_{N+1};\bX_{\backslash (i,l),:}, \bY_{\backslash (i,l)}\big)\big)$, as well as Definition \ref{def:sigma_LOO}; the last inequality applies Assumptions \ref{assump:stb_kurtosis} and \ref{assump:delta_LOO_bnd}, and assumes $N$ to be sufficiently large.
\begin{lem}\label{lem:diff_4moment}
    Suppose that $X_1$ and $X_2$ are independent and identically distributed, and $Z$ is independent from $X_1,\,X_2$. Then if $g(X_1,Z)$ has finite fourth moment, we have
    $$
    \bE\big(g(X_1,Z) - g(X_2,Z)\big)^4 \leq 16\bE\big[g(X_1,Z) - \bE g(X_1,Z)\big]^4.
    $$
\end{lem}
Now we can combine \eqref{eq:loss_stb_bnd_main}, \eqref{eq:loss_stb_bnd1}, \eqref{eq:loss_stb_bnd2}, \eqref{eq:loss_stb_prob}, and \eqref{eq:LOO_stb_4moment} to obtain a final bound for the loss stability:
\begin{equation*}
    \begin{split}
        \gamma_{loss}(h_j) \leq &\frac{8L^2\sum_{(m,n)\in S}n^2\stb(m,n)}{(1-\gamma)^2(N-1)^2} \\
        &+ \frac{C\delta_{\LOCO}^2(j,S)}{\delta_{\LOO}^2(S)}\Bigg(\frac{L^4\sum_{(m,n)\in S}n^4\stb^2(m,n)}{N^4}+\frac{s\sigma_{\LOO}^4(S)}{N^4}\Bigg)^{\frac{1}{2}}\\
        \leq &\frac{8L^2\sum_{(m,n)\in S}n^2\stb(m,n)}{(1-\gamma)^2(N-1)^2} \\
        &+ \frac{C\delta_{\LOCO}^2(j,S)}{\delta_{\LOO}^2(S)}\Bigg(\frac{L^2\sum_{(m,n)\in S}n^2\stb(m,n)}{N^2}+\frac{\sqrt{s}\sigma_{\LOO}^2(S)}{N^2}\Bigg)\\
        \leq &\frac{8L^2\sum_{(m,n)\in S}n^2\stb(m,n)}{(1-\gamma)^2(N-1)^2}\\
        &+\frac{CL^2\sum_{(m,n)\in S}n^2\stb(m,n)}{N^2}+\frac{C\sqrt{s}\delta^2_{\LOCO}(j,S)}{N^2}\Bigg)\\
        \leq &\frac{CL^2\sum_{(m,n)\in S}n^2\stb(m,n)}{N^2} + \frac{\sigma_j^2}{N^2}
    \end{split}
\end{equation*}
where the third inequality applies the fact that $\delta_{\LOCO}(j,S)\leq \sigma_j^2$ (Assumption \ref{assump:delta_LOCO_bnd} with the new definition of $h_j$), $\sigma_j^2\leq CL^2B^2$ as shown in \eqref{eq:sigma_j_bnd}, $\delta_{\LOO}(S)\geq cLB,\,c\sigma_{\LOO}(S)$ (Assumption \ref{assump:delta_LOO_bnd}). 
Therefore, by Theorem 2 in \cite{bayle2020cross}, we have $$\mathrm{Var}\big[\big(\max\big\{\hat{\sigma}_j\sqrt{N}, \epsilon(N)N\big\}\big)^{-1}\sum_{i=1}^NE_i\big]\leq C\gamma_{loss}(h_j)\epsilon^{-2}(N)\leq C\log^{-2}N + \frac{1}{N}\rightarrow 0,$$
which is due to Assumption \ref{assump:buffer_datadriven} and the fact that we are focusing the case $\epsilon(N)\geq \frac{\sigma_j}{\sqrt{N}}$. Now that we have dealt with the term $E_i$ in \eqref{eq:epsilon_3_decomp}, to show that $\varepsilon_j^{(3)}\overset{p}{\rightarrow}0$, it remains to bound $\epsilon^{-1}(N)N^{-1}\sum_{i=1}^N \big(\bar{h}_j(X_i,Y_i) - h_j(X_i,Y_i) + \bE(\bar{h}_j(X_i,Y_i) - h_j(X_i,Y_i))\big)$. To achieve this, we first note that
\begin{equation}\label{eq:epsilon3_term2}
    \begin{split}
        &\bE\Big[\epsilon^{-1}(N)N^{-1}\sum_{i=1}^N \big(\bar{h}_j(X_i,Y_i) - h_j(X_i,Y_i) + \bE(\bar{h}_j(X_i,Y_i) - h_j(X_i,Y_i))\big)\Big]^2\\
        = &\epsilon^{-2}(N)N^{-1}\bE\Big[\big(\bar{h}_j(X_i,Y_i) - h_j(X_i,Y_i) + \bE(\bar{h}_j(X_i,Y_i) - h_j(X_i,Y_i))\big)\Big]^2\\
        \leq &\epsilon^{-2}(N)N^{-1}\bE_{(X,Y)\sim \cP}\big(\bar{h}_j(X,Y) - h_j(X,Y)\big)^2\\
        = &\epsilon^{-2}(N)N^{-1}\bE_{(X,Y)\sim \cP}\Big[\bE_{\bX,\bY}\big(h_j(X,Y;\bX_{\backslash i,:},\bY_{\backslash i}) - h_j(X,Y;\bX,\bY)|X,Y\big)\Big]^2\\
        \leq &\epsilon^{-2}(N)N^{-1}\bE\big(h_j(X,Y;\bX_{\backslash i,:},\bY_{\backslash i}) - h_j(X,Y;\bX,\bY)\big)^2.
    \end{split}
\end{equation}
where the second line is due to that $\{(X_i,Y_i)\}_{i=1}^N$ are independent, identically distributed, the third line is due to the fact that $\bE(Z^2)\geq \mathrm{Var}(Z)$, and the fourth line utilizes the definition of $\bar{h}_j(X,Y)$ in \eqref{eq:bar_h_def_datadriven}. Now we can bound $\bE\big(h_j(X,Y;\bX_{\backslash i,:},\bY_{\backslash i}) - h_j(X,Y;\bX,\bY)\big)^2$ using similar arguments to bounding $\gamma_{loss}(h_j)\leq \bE\big(h_j(X_i,Y_i;\bX_{\backslash i,:},\bY_{\backslash i}) - h_j(X_i,Y_i;\bX^{\backslash l}_{\backslash i,:},\bY^{\backslash l}_{\backslash i})\big)^2$ in \eqref{eq:loss_stb_bnd_main}; the only difference lies that we look at the difference in $h_j$ when removing one training sample instead of replacing it with a new sample. In particular, 
\begin{equation}
    \begin{split}
        &\bE\big(h_j(X,Y;\bX_{\backslash i,:},\bY_{\backslash i}) - h_j(X,Y;\bX,\bY)\big)^2\\
        \leq&2\bE\big(h_j^{(m^*,n^*)}(X,Y;\bX_{\backslash i,:},\bY_{\backslash i}) - h_j^{(m^*,n^*)}(X,Y;\bX,\bY)\big)^2\\
        &+2\bE\big(h_j^{(m_{-i}^*,n_{-i}^*)}(X,Y;\bX_{\backslash i,:},\bY_{\backslash i}) - h_j^{(m^*,n^*)}(X,Y;\bX_{\backslash i,:},\bY_{\backslash i}\big)^2\\
        \leq &\frac{CL^2\sum_{(m,n)\in S}n^2\stb(m,n)}{N^2} + 2\delta^2_{\LOCO}(j,S)\sqrt{\bP\big((m^*,n^*)\neq (m_{-i}^*,n_{-i}^*)\big)}\\
        \leq &\frac{CL^2\sum_{(m,n)\in S}n^2\stb(m,n)}{N^2} + C\delta^2_{\LOCO}(j,S)\big(\frac{1}{N}+\frac{\sqrt{\sum_{(m,n)\in S}n^2\stb(m,n)}}{BN}\big)\\
        \leq &\frac{CL^2\sum_{(m,n)\in S}n^2\stb(m,n)}{N^2} + \frac{C\sigma_j^2}{N}+\frac{C\sigma_j L\sqrt{\sum_{(m,n)\in S}n^2\stb(m,n)}}{N}.
    \end{split}
\end{equation}
where the second inequality follows similar arguments in \eqref{eq:loss_stb_bnd1}, \eqref{eq:loss_stb_bnd2}, and utilizes \eqref{eq:loco_loo_mse}; the third inequality follows \eqref{eq:MPdiff_LOO_prob}; the last inequality is due to $\delta_{\LOCO}^2(j,S)\leq C\sigma_j^2\leq C'L^2B^2$ (Assumption \ref{assump:delta_LOCO_bnd} and \eqref{eq:sigma_j_bnd}). Recall that we have focused on the case where $\epsilon(N)>\sigma_j/\sqrt{N}$, \eqref{eq:epsilon3_term2} further implies that
\begin{equation*}
    \begin{split}
        &\bE\Big[\epsilon^{-1}(N)N^{-1}\sum_{i=1}^N \big(\bar{h}_j(X_i,Y_i) - h_j(X_i,Y_i) + \bE(\bar{h}_j(X_i,Y_i) - h_j(X_i,Y_i))\big)\Big]^2\\
        \leq &\frac{CL^2\sum_{(m,n)\in S}n^2\stb(m,n)}{N^3\epsilon^2(N)} + \frac{C\sigma_j^2}{N^2\epsilon^2(N)}+\frac{C\sigma_j L\sqrt{\sum_{(m,n)\in S}n^2\stb(m,n)}}{N^2\epsilon^2(N)}\\
        \leq &\frac{CL^2\sum_{(m,n)\in S}n^2\stb(m,n)}{N^3\epsilon^2(N)} + \frac{C}{N}+\frac{CL\sqrt{\sum_{(m,n)\in S}n^2\stb(m,n)}}{N^{3/2}\epsilon(N)}\\
        \leq &\frac{C}{N\log^2 N} + \frac{C}{N}+\frac{C}{\sqrt{N}\log N}\rightarrow 0.
    \end{split}
\end{equation*}
The last line utilizes Assumption \ref{assump:buffer}. Therefore, we have shown $\varepsilon_j^{(3)}\overset{p}{\rightarrow}0$.
\paragraph{Bounding $\varepsilon_j^{(4)}$.} $\varepsilon_j^{(4)}$ averages over the difference between the inference target with $N-1$ training data vs. using the full data. Recall that $\varepsilon_j^{(4)} = \frac{1}{\max\{\hat{\sigma}_j \sqrt{N},\epsilon(N)N\}}\sum_{i=1}^N\varepsilon_{i,j}^{(4)}$, where 
\begin{equation*}
    \begin{split}
        \varepsilon_{i,j}^{(4)} = &\bE\left[h_j^{(m^*_{-i},n^*_{-i})}(X_i,Y_i;\bX_{\backslash i,:}, \bY_{\backslash i})|\bX_{\backslash i,:}, \bY_{\backslash i}\right]-\bE\left[h_j^{(m^*,n^*)}(X,Y;\bX, \bY)|\bX, \bY\right]\\
        = & \bE\left[h_j^{(m^*_{-i},n^*_{-i})}(X_i,Y_i;\bX_{\backslash i,:}, \bY_{\backslash i})-h_j^{(m^*,n^*)}(X_i,Y_i;\bX_{\backslash i,:}, \bY_{\backslash i})|\bX_{\backslash i,:}, \bY_{\backslash i}\right]\\
        &+ \bE\left[h_j^{(m^*,n^*)}(X_i,Y_i;\bX_{\backslash i,:}, \bY_{\backslash i})|\bX_{\backslash i,:}, \bY_{\backslash i}\right]-\bE\left[h_j^{(m^*,n^*)}(X,Y;\bX, \bY)|\bX, \bY\right].
    \end{split}
\end{equation*}
Here, we have decomposed $\varepsilon_{i,j}^{(4)}$ into two error terms, one from the different minipatch sizes and the other from the training data when $(m^*,n^*)$ is selected. For notational convenience, throughout the proof for bounding $\varepsilon_j^{(4)}$, we let $h_{i,j}^{(m^*_{-i},n^*_{-i})}(X,Y) = h_j^{(m^*_{-i},n^*_{-i})}(X,Y;\bX_{\backslash i,:}, \bY_{\backslash i})$, $h_{j}^{(m^*,n^*)}(X,Y) = h_j^{(m^*,n^*)}(X,Y;\bX, \bY)$, and similarly define $h_{i,j}^{(m^*,n^*)}(X,Y)$. Also only in this proof, we denote $\frac{1}{\epsilon(N)N}\sum_{i=1}^N\bE\left[h_{i,j}^{(m^*_{-i},n^*_{-i})}(X_i,Y_i)-h_{i,j}^{(m^*,n^*)}(X_i,Y_i)|\bX_{\backslash i,:}, \bY_{\backslash i}\right]$ by $E_1$, and denote $\frac{1}{\epsilon(N)N}\sum_{i=1}^N \bE\left[h_{i,j}^{(m^*,n^*)}(X_i,Y_i)|\bX_{\backslash i,:}, \bY_{\backslash i}\right]-\bE\left[h_j^{(m^*,n^*)}(X,Y)|\bX, \bY\right]$ by $E_2$. Our goal is to show that both $E_1$ and $E_2$ converge to zero in probability. To bound $E_1$, we can apply similar arguments to the proof for bounding $\varepsilon_j^{(2)}$. First note that
\begin{equation*}
    \begin{split}
        &\bE\left[h_{i,j}^{(m^*_{-i},n^*_{-i})}(X_i,Y_i)-h_{i,j}^{(m^*,n^*)}(X_i,Y_i)|\bX_{\backslash i,:}, \bY_{\backslash i}\right]\\
        =&\bE\Big[\left(h_{i,j}^{(m^*_{-i},n^*_{-i})}(X_i,Y_i)-h_{i,j}^{(m^*,n^*)}(X_i,Y_i)\right)\ind(m^*\neq m^*_{-i},\text{ or }n^*\neq n^*_{-i})|\bX_{\backslash i,:}, \bY_{\backslash i}\Big]\\
        \leq &\sqrt{\bP(m^*\neq m^*_{-i},\text{ or }n^*\neq n^*_{-i}|\bX_{\backslash i,:}, \bY_{\backslash i})}\\
        &\cdot\sqrt{\bE\left[\left(h_{i,j}^{(m^*_{-i},n^*_{-i})}(X_i,Y_i)-h_{i,j}^{(m^*,n^*)}(X_i,Y_i)\right)^2|\bX_{\backslash i,:}, \bY_{\backslash i}\right]},
    \end{split}
\end{equation*}
which then implies
\begin{equation*}
    \begin{split}
        &E_1\\
        \leq&\frac{1}{\epsilon(N)}\sqrt{\frac{1}{N}\sum_{i=1}^N \bP(m^*\neq m^*_{-i},\text{ or }n^*\neq n^*_{-i}|\bX_{\backslash i,:}, \bY_{\backslash i})}\\
        &\cdot\sqrt{\frac{1}{N}\sum_{i=1}^N\bE\left[\left(h_{i,j}^{(m^*_{-i},n^*_{-i})}(X_i,Y_i)-h_{i,j}^{(m^*,n^*)}(X_i,Y_i)\right)^2|\bX_{\backslash i,:}, \bY_{\backslash i}\right]}\\
        \leq &\frac{\delta_{\LOCO}(j,S)}{\epsilon(N)}\sqrt{\frac{1}{N}\sum_{i=1}^N \bP(m^*\neq m^*_{-i},\text{ or }n^*\neq n^*_{-i}|\bX_{\backslash i,:}, \bY_{\backslash i})}.
    \end{split}
\end{equation*}
 Therefore, for any $\epsilon>0$, 
\begin{equation*}
    \begin{split}
        \bP(E_1>\epsilon)\leq &\bP\left(\sqrt{\frac{1}{N}\sum_{i=1}^N \bP(m^*\neq m^*_{-i},\text{ or }n^*\neq n^*_{-i}|\bX_{\backslash i,:}, \bY_{\backslash i})}>\frac{\epsilon\epsilon(N)}{\delta_{\LOCO}(j,S)}\right)\\
        \leq&\bP(m^*\neq m^*_{-i},\text{ or }n^*\neq n^*_{-i})\frac{\delta_{\LOCO}^2(j,S)}{\epsilon^2\epsilon^2(N)})\\
        \leq&\left[\frac{32s\sigma_{\LOO}^2(S)}{\delta^2_{\LOO}(S)N^2} + \frac{8L^2\sum_{(m,n)\in S}n^2\stb(m,n)}{\delta^2_{\LOO}(S)(N-1)^2}\right]\frac{\delta_{\LOCO}^2(j,S)}{\epsilon^2\epsilon^2(N)})\\
        \leq & \frac{C\sigma_j^2}{\epsilon^2 \epsilon^2(N)N^2} + \frac{C\sigma_j^2}{\epsilon^2\delta^2_{\LOO}(S)\log^2N},
    \end{split}
\end{equation*}
where the third line is due to our bound for $\bP(m^*\neq m^*_{-i},\text{ or }n^*\neq n^*_{-i})$ in \eqref{eq:MPdiff_LOO_prob}; the last line utilizes the fact that $s$ is bounded, Assumptions \ref{assump:delta_LOO_bnd}, \ref{assump:buffer_datadriven}, and \ref{assump:delta_LOCO_bnd}. Since $\epsilon(N)\geq \sigma_j/\sqrt{N}$, the first term above converges to zero; following the same argument as in \eqref{eq:sigma_j_bnd}, and recall the fact that $\delta_{\LOO}(S)\geq cLB$, the second term above can also be shown to converge to zero. Therefore, we have $\lim_{N\rightarrow \infty}\bP(E_1>\epsilon) = 0$ for any $\epsilon>0$. 

While for the second term $E_2$, we note that it can be bounded as follows:
\begin{equation*}
    E_2\leq \max_{(m,n)\in S}\frac{1}{\epsilon(N)N}\sum_{i=1}^N \bE\left[h_{i,j}^{(m,n)}(X_i,Y_i)|\bX_{\backslash i,:}, \bY_{\backslash i}\right]-\bE\left[h_j^{(m,n)}(X,Y)|\bX, \bY\right].
\end{equation*}
We can then follow the same argument as in Section \ref{sec:bnd_err2} to show that $$\frac{1}{\epsilon(N)N}\sum_{i=1}^N \bE\left[h_{i,j}^{(m,n)}(X_i,Y_i)|\bX_{\backslash i,:}, \bY_{\backslash i}\right]-\bE\left[h_j^{(m,n)}(X,Y)|\bX, \bY\right]$$ converges to zero in probability for any $(m,n)\in S$. Since $s=|S|$ is bounded, we immediately have the convergence in probability result for $E_2$ and hence $\varepsilon_j^{(4)}\overset{p}{\rightarrow}0$.

\paragraph{Bounding $\varepsilon_j^{(5)}$.} Recall that $\varepsilon_j^{(5)} = \min\{\sqrt{N}/\hat{\sigma}_j,\epsilon^{-1}(N)\}(\tilde{\Delta}_j-\Delta_j)$, where\\ $\tilde{\Delta}_j = \bE\big[h_j^{(m^*,n^*)}(X,Y;\bX,\bY)|\bX,\bY\big]$, and $\Delta_j = \bE\big[h_j^{(K,\hat{m},\hat{n})}(X,Y;\bX,\bY)|\bX,\bY\big]$. We note that by following the same arguments as in the Section \ref{sec:bnd_err4} in the proof of Lemma \ref{lem:bnd_err_terms}, we have
\begin{equation*}
    \epsilon^{-1}(N)|\tilde{\Delta}_j| - \bE\big[h_j^{(K,m^*,n^*)}(X,Y;\bX,\bY)|\bX,\bY\big]|\overset{p}{\rightarrow} 0.
\end{equation*}
Furthermore, $\bE\big[h_j^{(K,m^*,n^*)}(X,Y;\bX,\bY)|\bX,\bY\big] \neq \Delta_j$ if and only if $(m^*,n^*)\neq (\hat{m},\hat{n})$. Since we have already shown that $\bP((m^*,n^*)\neq (\hat{m},\hat{n}))\rightarrow 0$ in \eqref{eq:MPdiff_finiteK_prob}, we immediately have $\bP(|\bE\big[h_j^{(K,m^*,n^*)}(X,Y;\bX,\bY)|\bX,\bY\big] 
 - \Delta_j| 0)\rightarrow 0$, and hence $\varepsilon_j^{(5)}\overset{p}{\rightarrow}0$.

Having established that $\varepsilon_j^{(k)}\overset{p}{\rightarrow} 0$ for $1\leq k\leq 5$, our proof for Theorem \ref{thm:datadriven_coverage_barrier} is now complete. 
\end{proof}
\begin{proof}[Proof of Lemma \ref{lem:diff_4moment}]
    Let $\mu = \bE g(X_1,Z) = \bE g(X_2,Z)$. We can then write
    \begin{equation*}
        \begin{split}
          &\bE\big(g(X_1,Z) - g(X_2,Z)\big)^4\\
          \leq&\bE\big(\big(g(X_1,Z)-\mu\big) - \big(g(X_2,Z)-\mu\big)\big)^4\\
          =&16\bE\Big[\frac{1}{2}[\big(g(X_1,Z)-\mu\big) - \big(g(X_2,Z)-\mu\big)]\Big]^4\\
          \leq &8\Big[\bE[\big(g(X_1,Z)-\mu\big)]^4 + \bE[\big(g(X_2,Z)-\mu\big)]^4\Big]\\
          =&16\bE[\big(g(X_1,Z)-\mu\big)]^4,
        \end{split}
    \end{equation*}
    where we have applied Jensen's inequality on the third line.
\end{proof}
\subsection{Proof of Theorem \ref{thm:pred_coverage}}\label{sec:proof_pred}
Our proof closely follows the proofs in \cite{barber2021predictive} and \cite{kim2020predictive}, while the main difference lies that our algorithm also subsample features randomly. For completeness, we will write out the full proof, including the steps that are very similar to \cite{barber2021predictive} and \cite{kim2020predictive}.

First we suppose that we have access to a new data point $(X_{N+1},Y_{N+1})$, and we consider the following ``lifted'' algorithm that is similar to the one defined in \cite{kim2020predictive} and is symmetric w.r.t. all $N+1$ data points $(X_1,Y_1),\dots,(X_{N+1},Y_{N+1})$. The extended training data is denoted by $(\bX^*,\bY^*)$ with $\bX^*\in \bR^{(N+1)\times M}$ and $\bY^*\in \bR^{N+1}$.

\begin{algorithm}[H]\label{algo:lifted_mp}~
\SetAlgoLined
\footnotesize
  \DontPrintSemicolon
  \SetKwInOut{Input}{input}\SetKwInOut{Output}{output}
\noindent{\textbf{Input}}: Training data $\{(X_i,Y_i)\}_{i=1}^{N+1}$, minipatch sizes $n$, $m$; number of minipatches $\tilde{K}$, base learner $H$;
\begin{enumerate}
    \item Perform Minipatch Learning: For $k=1,...,\tilde{K}$:
 \begin{enumerate}
   \item Randomly subsample $n$ observations, $I_k \subset [N+1]$, and $m$ features, $F_k \subset [M]$.
        \item Train prediction model $\mu_k$ on $(\bX^*_{I_k,F_k}, \bY^*_{I_k})$: $\mu_k=H(\bX^*_{I_k,F_k}, \bY^*_{I_k})$.
           \end{enumerate}
\item Obtain leave-two-out predictions : For $i_1\neq i_2\in [N+1]$:
\begin{enumerate}
        \item  Obtain the ensembled leave-two-out prediction for $i_1, i_2$:\\ $\mu_{-i_1,-i_2}(X_{i_1}) = \frac{1}{\sum_{k=1}^{K} \ind(i_1 \notin I_k)\ind(i_2\notin I_k)} \sum_{k=1}^{K} \ind(i_1 \notin I_k)\ind(i_2\notin I_k) \mu_{k} (X_{i_1})$;\\
        $\mu_{-i_1,-i_2}(X_{i_2}) = \frac{1}{\sum_{k=1}^{K} \ind(i_1 \notin I_k)\ind(i_2\notin I_k)} \sum_{k=1}^{K} \ind(i_1 \notin I_k)\ind(i_2\notin I_k) \mu_{k} (X_{i_2})$;
           \item  Obtain the non-conformity score/residual for $i_1, i_2$:\\
           $R_{i_1,i_2} = \err(Y_{i_1},\mu_{-i_1,-i_2}(X_{i_1})$;\\
           $R_{i_2,i_1} = \err(Y_{i_2},\mu_{-i_1,-i_2}(X_{i_2})$;
           \end{enumerate}
 \end{enumerate}
 \textbf{Output}: Residuals $(R_{i_1,i_2}:i_1\neq i_2\in [N+1])$.
 \caption{Lifted J+MP Minipatch Predictive Interval}
\end{algorithm}

Define matrix $R\in \bR^{(N+1)\times (N+1)}$ as the leave-two-out non-conformity score matrix, where $R_{i_1,i_2}$ is the output of Algorithm \ref{algo:lifted_mp} if $i_1\neq i_2$, and diagonal entries $R_{i,i}=0$ for all $i$. Also define comparison matrix $A\in \bR^{(N+1)\times (N+1)}$ by letting $A_{i_1,i_2}=\ind(R_{i_1,i_2}>R_{i_2,i_1})$. Let $S(A)=\{i:\sum_{i'=1}^{N+1}A_{i,i'}\geq (1-\alpha)(N+1)\}$ denote the indices of samples which have higher non-conformity scores than $(1-\alpha)(N+1)$ samples. In the following, we will first show the size of $S(A)$ is small and $\bP(N+1\in S(A))\leq 2\alpha$, and then we will build a connection between $N+1\in S(A)$ and $Y_{N+1}\notin \hat{C}_{\alpha}^{\mathrm{J+MP}}$.

By using exactly the same arguments as in the proof of Theorem 1 in \cite{barber2021predictive}, we have $|S(A)|<2\alpha(N+1)$. Now we prove that the distribution of $S(A)$ would not change when $\{(X_i,Y_i)\}_{i=1}^{N+1}$ are arbitrarily exchanged. We first note that the residual matrix $R$ is a function of the extended training data $(\bX^*,\bY^*)$ and subsampled minipatches $(I_1,F_1),\dots,(I_{\tilde{K}},F_{\tilde{K}})$, and we denote this function by $\mathcal{A}$:
$$
R=\mathcal{A}(\bX^*,\bY^*;(I_1,F_1),\dots,(I_{\tilde{K}},F_{\tilde{K}})),
$$
where each entry of $R$ satisfies
$$R_{i_1,i_2}=\err(Y_{i_1},\frac{1}{\sum_{k=1}^{K} \ind(i_1 \notin I_k)\ind(i_2\notin I_k)}\sum_{k=1}^{K} \ind(i_1 \notin I_k)\ind(i_2\notin I_k) H(\bX^*_{I_k,F_k},\bY^*_{I_k})(X_{i_1}).$$
Consider an arbitrary permutation $\sigma$ on $[N+1]$, and let $\Pi_{\sigma}\in \{0,1\}^{(N+1)\times (N+1)}$ be its matrix representation, with $(\Pi_{\sigma})_{\sigma(i),:}=e_i^\top$ for $i=1,\dots,N+1$. Also let $\sigma(I_k)=\{\sigma(i):i\in I_k\}$. Then we can also write $Y_{i_1}=(\Pi_{\sigma}\bY^*)_{\sigma(i_1)}$, and $H(\bX^*_{I_k,F_k},\bY^*_{I_k})(X_{i_1})=H((\Pi_{\sigma}\bX^*)_{\sigma(I_k),F_k},(\Pi_{\sigma}\bY^*)_{\sigma(I_k)})((\Pi_{\sigma}\bX^*)_{\sigma(i_1})$.
Thus 
\begin{align*}
    &(\Pi_{\sigma}R\Pi_{\sigma}^\top)_{\sigma(i_1),\sigma(i_2)}\\
    =&R_{i_1,i_2}\\
    =&\err((\Pi_{\sigma}\bY^*)_{\sigma(i_1)},\frac{\sum_{k=1}^{K} \ind(\sigma(i_1) \notin \sigma(I_k))\ind(\sigma(i_2)\notin \sigma(I_k)) H((\Pi_{\sigma}\bX^*)_{\sigma(I_k),F_k},(\Pi_{\sigma}\bY^*)_{\sigma(I_k)})((\Pi_{\sigma}\bX^*)_{\sigma(i_1)})}{\sum_{k=1}^{K} \ind(\sigma(i_1) \notin \sigma(I_k))\ind(\sigma(i_2)\notin \sigma(I_k))}.
\end{align*}
Therefore,
$$
\Pi_{\sigma}R\Pi_{\sigma}^\top=\mathcal{A}(\Pi_{\sigma}\bX^*,\Pi_{\sigma}\bY^*;(\sigma(I_1),\sigma(F_1)),\dots,(\sigma(I_{\tilde{K}}),\sigma(F_{\tilde{K}}))).
$$
Since $\{(I_k,F_k)\}_{k=1}^{\tilde{K}}$ are independent random sets with uniform distribution over $[N+1]\times [M]$, $\{(I_k,F_k)\}_{k=1}^{\tilde{K}}\overset{\rm d.}{=}\{(\sigma(I_k),\sigma(F_k))\}_{k=1}^{\tilde{K}}$. Meanwhile, by Assumption \ref{assump:data_exchange} and Assumption \ref{assump:alg_orderinvariant}, $\Pi_{\sigma}\bX^*\overset{\rm d.}{=}\bX^*$, $\Pi_{\sigma}\bY^*\overset{\rm d.}{=}\bY^*$, and $H(\cdot)$ is invariant to the order of the input. Hence $\Pi_{\sigma}R\Pi_{\sigma}^\top\overset{\rm d.}{=}R$. Since $A$ and $S(A)$ are functions of $R$, we also have $S(A)\overset{\rm d.}{=}S(\Pi_{\sigma}A\Pi_{\sigma}^\top)$. For any $1\leq i\leq N$, there exists a permutation $\sigma(i)=N+1$, and thus $\bP(N+1\in S(A))=\bP(N+1\in S(\Pi_{\sigma}A\Pi_{\sigma}^\top))=\bP(i\in S(A))$. Since this holds for all $1\leq i\leq N$, $\bP(N+1\in S(A))=\frac{\bE(\sum_{i=1}^{N+1}\ind(i\in S(A))}{N+1}=\frac{\bE(|S(A)|)}{N+1}\leq 2\alpha$.

Now we show a connection between the events $N+1\in S(A)$ and $Y_{N+1}\notin \hat{C}_{\alpha}^{\mathrm{J+MP}}$. Let $K=\sum_{k=1}^{\tilde{K}}\ind(N+1\notin I_k)$, then $K$ follows a binomial distribution with parameters $(\tilde{K},1-\frac{n}{N+1})$ since $I_k$ is randomly sampled from $[N+1]$ without replacement. Collect the minipatches $\{(I_k,F_k):N+1\notin I_k\}$ and notice that $\{(I_k,F_k):N+1\notin I_k\}$ are independent random subsets that are uniformly sampled from $[N]\times [M]$. For each $1\leq i\leq N$, we ensemble the minipatch predictions from $\{(I_k,F_k):i,N+1\notin I_k\}$ for $X_i$ and $X_{N+1}$, and compute their corresponding non-conformity score, then (i) they are exactly $R_{i,N+1}$ and $R_{N+1,i}$ returned from Algorithm \ref{algo:lifted_mp}; (ii) they also share the same joint distribution as $\{(R_i^{LOO},\err(Y_{N+1},\mu_{-i}(X_{N+1}))\}_{i=1}^N$ where $R_i^{LOO}$ and $\mu_{-i}(X_{N+1}))$ are returned from Algorithm \ref{algo:j+mp}. Therefore, $$\bP(\sum_{i=1}^{N}\mathbb{I}(\err(Y_{N+1},\mu_{-i}(X_{N+1}))\geq R^{(LOO)}_i)\geq (1-\alpha)(N+1))=\bP(N+1\in S(A))\leq 2\alpha,$$ verifying that for the classification setting, $\bP(Y_{N+1}\in \hat{C}_{\alpha}^{\mathrm{J+MP}}(X_{N+1}))\geq 1-2\alpha$ with $\hat{C}_{\alpha}^{\mathrm{J+MP}}(X_{N+1})$ being defined in \eqref{eq:pred_interval_class}. While for the regression setting, note that if $Y_{N+1}\notin\hat{C}_{\alpha}^{\mathrm{J+MP}}(X_{N+1})$ for $\hat{C}_{\alpha}^{\mathrm{J+MP}}(X_{N+1})$ defined in \eqref{eq:pred_interval_regress}, we have $\sum_{i=1}^N\ind(Y_{N+1}\geq\mu_{-i}(X_{N+1})+R_i^{LOO})\geq (1-\alpha)(N+1)$ or $\sum_{i=1}^N\ind(Y_{N+1}\leq\mu_{-i}(X_{N+1})-R_i^{LOO})\geq (1-\alpha)(N+1)$ hold, which implies $$\sum_{i=1}^N|\ind(Y_{N+1}-\mu_{-i}(X_{N+1})|\geq R_i^{LOO})\geq (1-\alpha)(N+1).$$ If choosing error function to be the absolute error, we have
\begin{align*}
    &\bP(Y_{N+1}\notin\hat{C}_{\alpha}^{\mathrm{J+MP}}(X_{N+1}))\\
    \leq&\bP(\sum_{i=1}^N|\ind(Y_{N+1}-\mu_{-i}(X_{N+1})|\geq R_i^{LOO})\geq (1-\alpha)(N+1))\\
    \leq&\bP(\sum_{i=1}^{N}\mathbb{I}(\err(Y_{N+1},\mu_{-i}(X_{N+1})\geq R^{(LOO)}_i))\geq (1-\alpha)(N+1))\\
    \leq& 2\alpha,
\end{align*}
which finishes our proof.

\subsection{Proofs of the Theoretical Results in Section \ref{sec:methods}}\label{sec:disc_proof}
In this section, we present the proofs of the theoretical results we presented in Section \ref{sec:thm_disc_target}. 
\begin{proof}[Proof of Theorem \ref{prop:MP_target_full}]
Let $$\Delta_j^* = \lim_{K\rightarrow \infty}\Delta_j = \lim_{K\rightarrow \infty}\bE_{X,Y}[\err(Y,\mu(X;\bX,\bY)-\err(Y,\mu_{\backslash j}(X;\bX,\bY)|\bX,\bY].$$ By Assumption \ref{assump:FiniteExpectationErrMP} and the dominated convergence theorem, the limit and expectation over $X,Y$ can be exchanged. Furthermore, by the Lipschitz continuity of the error function, and by applying the strong law of large numbers on $\mu(X;\bX,\bY)=\frac{1}{K}\sum_{k=1}^K\mu_{I_k,F_k}(X)$ and $\mu_{\backslash j}(X_{\backslash j};\bX_{:,\backslash j},\bY)$, we have $\Delta_j^*=\bE_{X,Y}[\err(Y,\mu^*(X;\bX,\bY)-\err(Y,\mu^*_{\backslash j}(X;\bX,\bY)|\bX,\bY]$, where $\mu^*$ and $\mu^*_{\backslash j}$ are the combinatorial average of all minipatch predictors, defined in \eqref{eq:minipatch_predictor}. For any index sets $I=\{i_1,\dots,i_n\}\subset [N]$ with $i_1<i_2<\dots<i_n$ and $F=\{j_1,\dots,j_m\}\subset [M]$ with $j_1<j_2<\dots<j_m$, let $R_I\in\mathbb{R}^{n\times N}$ and $R_F\in \mathbb{R}^{m\times M}$ be subsampling matrices defined as follows: $(R_I)_{k,l} = \ind(i_k=l)$, $(R_F)_{k,l} = \ind(j_k = l)$. Then we can also write $\boldsymbol{X}_{I,F} = R_I\boldsymbol{X}R_F^\top$. 

When the base learner of each minipatch ensemble is a least squares estimator and $\boldsymbol{X}_{I,F}^\top \boldsymbol{X}_{I,F}$ is of full rank for all $I,\, F$, we have 
\begin{align*}
    H(\boldsymbol{X}_{I,F},\boldsymbol{Y}_{I})(X_{F})=&X_F^\top(\boldsymbol{X}_{I,F}^\top \boldsymbol{X}_{I,F})^{-1}\boldsymbol{X}_{I,F}^\top \boldsymbol{Y}_{I}\\
    =&X^\top R_F^\top(\boldsymbol{X}_{I,F}^\top \boldsymbol{X}_{I,F})^{-1}\boldsymbol{X}_{I,F}^\top (\boldsymbol{X}_{I,:}\beta^* + \epsilon_I)\\
    =&X^\top R_F^\top[\beta_F^* + (\boldsymbol{X}_{I,F}^\top \boldsymbol{X}_{I,F})^{-1}\boldsymbol{X}_{I,F}^\top (\boldsymbol{X}_{I,F^c}\beta^*_{F^c} + \epsilon_I)],
\end{align*} 
and hence 
\begin{align*}
&\mu^*(X;\boldsymbol{X},\boldsymbol{Y})\\
= &X^\top \frac{1}{\binom{N}{n}\binom{M}{m}}\sum_{I\subset[N],F\subset [M]}R_F^\top[\beta_F^* + (\boldsymbol{X}_{I,F}^\top \boldsymbol{X}_{I,F})^{-1}\boldsymbol{X}_{I,F}^\top (\boldsymbol{X}_{I,F^c}\beta^*_{F^c} + \epsilon_I)]\\
=&X^\top \left\{\frac{m}{M}\beta^* + \frac{1}{\binom{N}{n}\binom{M}{m}}\sum_{I\subset[N],F\subset [M]}R_F^\top[(\boldsymbol{X}_{I,F}^\top \boldsymbol{X}_{I,F})^{-1}\boldsymbol{X}_{I,F}^\top (\boldsymbol{X}_{I,F^c}\beta^*_{F^c} + \epsilon_I)]\right\}.
\end{align*}
Similarly, we can write 
\begin{align*}
&\mu^*_{\backslash j}(X_{\backslash j};\boldsymbol{X}_{:,\backslash j},\boldsymbol{Y})\\
= &X^\top \frac{1}{\binom{N}{n}\binom{M-1}{m}}\sum_{I\subset[N],j\notin F}R_F^\top[\beta_F^* + (\boldsymbol{X}_{I,F}^\top \boldsymbol{X}_{I,F})^{-1}\boldsymbol{X}_{I,F}^\top (\boldsymbol{X}_{I,F^c}\beta^*_{F^c} + \epsilon_I)]\\
=&X^\top \left\{\frac{m}{M-1}\beta^{*\backslash j} + \frac{1}{\binom{N}{n}\binom{M-1}{m}}\sum_{I\subset[N],j\notin F}R_F^\top[(\boldsymbol{X}_{I,F}^\top \boldsymbol{X}_{I,F})^{-1}\boldsymbol{X}_{I,F}^\top (\boldsymbol{X}_{I,F^c}\beta^*_{F^c} + \epsilon_I)]\right\},
\end{align*}
where we denote by $\beta^{*\backslash j}\in \mathbb{R}^M$ the true regression parameter but setting the $j$th coordinate as zero: $\beta^{*\backslash j}_j = 0$ and $\beta^{*\backslash j}_{\backslash j} = \beta^*_{\backslash j}$. For any subset $F\subset [M]$ of size $m$, let $\varepsilon_{1,F}, \varepsilon_{2,F}\in \mathbb{R}^M$ be defined as follows:
\begin{equation}\label{eq:target_epsilon_F}
\begin{split}
    \varepsilon_{1,F} =& \frac{1}{\binom{N}{n}}\sum_{I\subset[N],|I|=n}R_F^\top(\boldsymbol{X}_{I,F}^\top \boldsymbol{X}_{I,F})^{-1}\boldsymbol{X}_{I,F}^\top \boldsymbol{X}_{I,F^c}\beta^*_{F^c},\\
    \varepsilon_{2,F} =& \frac{1}{\binom{N}{n}}\sum_{I\subset[N],|I|=n}R_F^\top(\boldsymbol{X}_{I,F}^\top \boldsymbol{X}_{I,F})^{-1}\boldsymbol{X}_{I,F}^\top \epsilon_I.
\end{split}
\end{equation}
Also define the following six $M$-dimensional error terms:
\begin{equation}\label{eq:target_epsilon_def}
    \begin{split}
        \varepsilon_{1,1} = &\frac{1}{\binom{M}{m}}\sum_{F\subset [M], |F|=m}\varepsilon_{1,F}, \quad \varepsilon_{1,2} =  \frac{1}{\binom{M}{m}}\sum_{F\subset [M], |F|=m}\varepsilon_{2,F},\\
    \varepsilon_{2,1} = &\frac{1}{\binom{M-1}{m}}\sum_{j\notin F, |F|=m}\varepsilon_{1,F}, \quad \varepsilon_{2,2} =  \frac{1}{\binom{M-1}{m}}\sum_{j\notin F, |F|=m}\varepsilon_{2,F},\\
    \varepsilon_{3,1} = &\frac{1}{\binom{M-1}{m-1}}\sum_{j\in F, |F|=m}\varepsilon_{1,F}, \quad \varepsilon_{3,2} =  \frac{1}{\binom{M-1}{m-1}}\sum_{j\in F, |F|=m}\varepsilon_{2,F}.
    \end{split}
\end{equation}
Since $\mathbb{E}(\epsilon|\boldsymbol{X})=0$, $\mathbb{E}(\boldsymbol{X}_{I,F}^\top \boldsymbol{X}_{I,F^c})=0$, all these six error terms are of mean zero. We can also write the first two error terms as a function of the latter four errors:
$$
\varepsilon_{1,1} = \frac{M-m}{M}\varepsilon_{2,1} + \frac{m}{M}\varepsilon_{3,1},\quad \varepsilon_{1,2} = \frac{M-m}{M}\varepsilon_{2,2} + \frac{m}{M}\varepsilon_{3,2}.
$$
Now we can further decompose our inference target as follows:
\begin{equation}\label{eq:Delta_decompose}
    \begin{split}
    \Delta_j^* = &\mathbb{E}_{X,Y}\left(Y-X^\top\left(\frac{m}{M-1}\beta^{*\backslash j} + \varepsilon_{2,1} + \varepsilon_{2,2}\right)\right)^2 - \mathbb{E}_{X,Y}\left(Y-X^\top\left(\frac{m}{M}\beta^* + \varepsilon_{1,1} + \varepsilon_{1,2}\right)\right)^2\\
    = &\left\|\beta^* - \frac{m}{M-1}\beta^{*\backslash j} - \varepsilon_{2,1} - \varepsilon_{2,2}\right\|_2^2-\left\|\frac{M-m}{M}\beta^* - \varepsilon_{1,1} - \varepsilon_{1,2}\right\|_2^2\\
    = &\gamma(2-\gamma)\beta_j^{*2} - \left(\frac{2\gamma}{M-1} -\frac{\gamma^2(2M-1)}{(M-1)^2}\right)\|\beta^*_{\backslash j}\|_2^2-2(\varepsilon_{2,1}+\varepsilon_{2,2})^\top (\beta^* - \frac{m}{M-1}\beta^{*\backslash j})\\
    &+2\frac{M-m}{M}(\varepsilon_{1,1}+\varepsilon_{1,2})^\top \beta^*+\|\varepsilon_{1,1}+\varepsilon_{1,2}\|_2^2+ \|\varepsilon_{2,1}+\varepsilon_{2,2}\|_2^2\\
    = &\gamma(2-\gamma)\beta_j^{*2} - \left(\frac{2\gamma}{M-1} -\frac{\gamma^2(2M-1)}{(M-1)^2}\right)\|\beta^*_{\backslash j}\|_2^2+\|\varepsilon_{1,1}+\varepsilon_{1,2}\|_2^2+ \|\varepsilon_{2,1}+\varepsilon_{2,2}\|_2^2\\
    &+2(\varepsilon_{2,1}+\varepsilon_{2,2})^\top\left[\frac{m}{M-1}\beta^{*\backslash j}-(2\gamma-\gamma^2)\beta^*\right]
    +2\gamma(1-\gamma)(\varepsilon_{3,1}+\varepsilon_{3,2})^\top \beta^*.
\end{split}
\end{equation}
In particular, since $$\left\|\frac{m}{M-1}\beta^{*\backslash j}-(2\gamma-\gamma^2)\beta^*\right\|_2\leq 4\gamma^2\beta_j^{*2}+\gamma^2\|\beta^*_{\backslash j}\|_2\leq 4\gamma^2\|\beta^*\|_2^2, $$
\eqref{eq:Delta_decompose} implies that
\begin{equation}\label{eq:MP_target_basic_bnd}
    \begin{split}
        &\left|\Delta_j-\Delta_j^{*(L)}\right|\\
    \leq&4\gamma\sum_{l=1}^2\|\varepsilon_{2,l}\|_2\|\beta^*\|_2+2\sum_{k=1}^2\sum_{l=1}^2\|\varepsilon_{k,l}\|_2^2+2\gamma\sum_{l=1}^2\|(\varepsilon_{3,l})_{\backslash j}\|_2\|\beta^*_{\backslash j}\|_2+2\gamma\sum_{l=1}^2|(\varepsilon_{3,l})_j||\beta^*_{j}|.
    \end{split}
\end{equation}

The following lemma suggests that $\|\varepsilon_{k,l}\|_2$ can be bounded by functions of $\|\varepsilon_{l,F}\|_2^2$ for $1\leq k\leq 3$, $l=1,2$.
\begin{lem}\label{lem:target_epsilon_decomposed}
For $\varepsilon_{k,l}$, $k=1,2,3$, $l=1,2$, defined in \eqref{eq:target_epsilon_def}, we have
\begin{align*}
    \|\varepsilon_{1,l}\|_2\leq &\sqrt{\frac{m}{M}}\sqrt{\frac{1}{\binom{M}{m}}\sum_{F\subset [M], |F|=m}\|\varepsilon_{l,F}\|_2^2},\\
    \|\varepsilon_{2,l}\|_2\leq &\sqrt{\frac{m}{M-1}}\sqrt{\frac{1}{\binom{M-1}{m}}\sum_{j\notin F, |F|=m}\|\varepsilon_{l,F}\|_2^2}\\
    \|(\varepsilon_{3,l})_{\backslash j}\|_2\leq &\sqrt{\frac{m-1}{M-1}\frac{1}{\binom{M-1}{m-1}}\sum_{j\in F, |F|=m}\|(\varepsilon_{l,F})_{\backslash j}\|_2^2},\\
    |(\varepsilon_{3,l})_j|\leq&\sqrt{\frac{1}{\binom{M-1}{m-1}}\sum_{j\in F, |F|=m}(\varepsilon_{l,F})_{j}^2}.
\end{align*}
hold for $l=1,2$.
\end{lem}

\begin{proof}[Proof of Lemma \ref{lem:target_epsilon_decomposed}]
    Recall the definition of $\varepsilon_{1,l}$, we can write
    \begin{align*}
        \|\varepsilon_{1,l}\|_2^2=&\frac{1}{\binom{M}{m}^2}\sum_{F\subset[M], |F|=m}\sum_{F'\subset[M], |F|=m'}\varepsilon_{l,F}^\top\varepsilon_{l,F'}.
    \end{align*}
    By the definition \eqref{eq:target_epsilon_F} of $\varepsilon_{l,F}$ and the definition of subsampling matrix $R_F$, it is straightforward to see that $(\varepsilon_{l,F})_{F^c}=0$. Hence $\varepsilon_{l,F}^\top\varepsilon_{l,F'}\leq \|(\varepsilon_{l,F})_{F'}\|_2\|(\varepsilon_{l,F'})_{F}\|_2$, and 
    \begin{align*}
        \|\varepsilon_{1,l}\|_2^2\leq&\frac{1}{\binom{M}{m}^2}\sum_{F\subset[M], |F|=m}\sum_{F'\subset[M], |F|=m'}\|(\varepsilon_{l,F})_{F'}\|_2\|(\varepsilon_{l,F'})_{F}\|_2\\
        \leq &\frac{1}{2\binom{M}{m}^2}\sum_{F\subset[M], |F|=m}\sum_{F'\subset[M], |F|=m'}\|(\varepsilon_{l,F})_{F'}\|_2^2+\|(\varepsilon_{l,F'})_{F}\|_2^2\\
        \leq&\frac{1}{\binom{M}{m}^2}\sum_{F\subset[M], |F|=m}\sum_{F'\subset[M], |F|=m'}\|(\varepsilon_{l,F})_{F'}\|_2^2\\
        \leq&\frac{1}{\binom{M}{m}^2}\sum_{F\subset[M], |F|=m}\binom{M-1}{m-1}\|\varepsilon_{l,F}\|_2^2\\
        \leq &\frac{m}{M}\frac{1}{\binom{M}{m}}\sum_{F\subset[M], |F|=m}\|\varepsilon_{l,F}\|_2^2.
    \end{align*}
    Thus we have $\|\varepsilon_{1,l}\|_2\leq \sqrt{\frac{m}{M}}\sqrt{\frac{1}{\binom{M}{m}}\sum_{F\subset[M], |F|=m}\|\varepsilon_{l,F}\|_2^2}$, and similarly one can show that $\|\varepsilon_{2,l}\|_2\leq \sqrt{\frac{m}{M-1}}\sqrt{\frac{1}{\binom{M-1}{m}}\sum_{j\notin F, |F|=m}\|\varepsilon_{l,F}\|_2^2}$. While for $\varepsilon_{3,l}$, some calculations show that
    \begin{align*}
        \|(\varepsilon_{3,l})_{\backslash j}\|_2^2\leq &\frac{1}{\binom{M-1}{m-1}^2}\sum_{j\in F, |F|=m}\sum_{j\in F', |F'|=m}\|(\varepsilon_{l,F})_{F'\backslash j}\|_2\|(\varepsilon_{l,F'})_{F\backslash j}\|_2\\
        \leq&\frac{1}{\binom{M-1}{m-1}^2}\sum_{j\in F, |F|=m}\sum_{j\in F', |F'|=m}\|(\varepsilon_{l,F})_{F'\backslash j}\|_2^2\\
        \leq&\frac{1}{\binom{M-1}{m-1}^2}\sum_{j\in F, |F'|=m}\binom{M-2}{m-2}\|(\varepsilon_{l,F})_{\backslash j}\|_2^2\\
        \leq &\frac{m-1}{M-1}\frac{1}{\binom{M-1}{m-1}}\sum_{j\in F, |F'|=m}\|(\varepsilon_{l,F})_{\backslash j}\|_2^2.
    \end{align*}
    In addition,
    \begin{align*}
        (\varepsilon_{3,l})_j^2\leq&\frac{1}{\binom{M-1}{m-1}^2}\sum_{j\in F, |F|=m}\sum_{j\in F', |F|=m}(\varepsilon_{l,F})_j(\varepsilon_{l,F'})_{j}\\
        =&\frac{1}{\binom{M-1}{m-1}}\sum_{j\in F, |F|=m}(\varepsilon_{l,F})_{j}^2.
    \end{align*}
\end{proof}

In the following, we will focus on bounding $\|\varepsilon_{1,F}\|_2^2$ and $\|\varepsilon_{2,F}\|_2^2$.
\paragraph{Bounding $\|\varepsilon_{1,F}\|_2^2$:} 
By the definition \eqref{eq:target_epsilon_F}, we can directly write
\begin{align*}
    \|\varepsilon_{1,F}\|_2^2=&\left\|\frac{1}{\binom{N}{n}}\sum_{I\subset[N], |I|=n}(\boldsymbol{X}_{I,F}^\top \boldsymbol{X}_{I,F})^{-1}\boldsymbol{X}_{I,F}^\top R_I\boldsymbol{X}_{:,F^c}\beta^*_{F^c}\right\|_2^2\\
    :=&(\boldsymbol{X}_{:,F^c}\beta^*_{F^c})^\top \boldsymbol{A}_{F} (\boldsymbol{X}_{:,F^c}\beta^*_{F^c}),
\end{align*}
where we denote by $\boldsymbol{A}_{F}\in \mathbb{R}^{N\times N}$ the following matrix:
$$
\boldsymbol{A}_{F}=\frac{1}{\binom{N}{n}^2}\sum_{I\subset[N], |I|=n}\sum_{I'\subset[N], |I'|=n}R_I^\top \boldsymbol{X}_{I,F}(\boldsymbol{X}_{I,F}^\top \boldsymbol{X}_{I,F})^{-1}(\boldsymbol{X}_{I',F}^\top \boldsymbol{X}_{I',F})^{-1}\boldsymbol{X}_{I',F}^\top R_{I'}.
$$
Since covariates $\{X_i\}_{i=1}^N$ are mean zero independent sub-Gaussian random vectors with sub-Gaussian parameter $C$, we know that conditioning on $\boldsymbol{X}_{:,F}$, $\boldsymbol{X}_{:,F^c}\beta^*_{F^c}\in \mathbb{R}^{N}$ have independent centered sub-Gaussian entries with sub-Gaussian norm $C\|\beta^*_{F^c}\|_2$ and variance $\|\beta^*_{F^c}\|_2$. Hence $\mathbb{E}[\|\varepsilon_{1,F}\|_2^2|\boldsymbol{A}_{F}]=\|\beta_{F^c}^*\|_2^2\mathrm{tr}(\boldsymbol{A}_{F})$.
Furthermore, to concentrate the quadratic $(\boldsymbol{X}_{:,F^c}\beta^*_{F^c})^\top \boldsymbol{A}_{F} (\boldsymbol{X}_{:,F^c}\beta^*_{F^c})^\top$, we can simply apply the Hanson-Wright inequality \citep[see][Theorem 1.1]{rudelson2013hanson} for sub-Gaussian random variables, and obtain the following
\begin{align*}
    \mathbb{P}\left(\left|\|\varepsilon_{1,F}\|_2^2-\mathbb{E}(\|\varepsilon_{1,F}\|_2^2|\boldsymbol{A}_F)\right| >t\right)\leq 2\exp\left\{-c\min\left\{\frac{t^2}{\|\beta_{F^c}^*\|_2^4\|\boldsymbol{A}_{F}\|_F^2}, \frac{t}{\|\beta_{F^c}^*\|_2^2\|\boldsymbol{A}_{F}\|_2}\right\}\right\}.
\end{align*}
Let $t=\|\beta^*_{F^c}\|_2^2\max\{\|\boldsymbol{A}_{F}\|_F\sqrt{\log N}, \|\boldsymbol{A}_{F}\|_2\log N\}$ in the inequality above, then we have 
\begin{equation}\label{eq:MP_target_epsilon_1F}
    \|\varepsilon_{1,F}\|_2^2\leq C\|\beta_{F^c}^*\|_2^2\left(\mathrm{tr}(\boldsymbol{A}_{F})+\|\boldsymbol{A}_{F}\|_F\log N\right),
\end{equation}
with probability at least $1-CN^{-c}$. Here, we have applied the fact that $\|\boldsymbol{A}_{F}\|_2\leq \|\boldsymbol{A}_{F}\|_F$.

Now we focus on bounding the Frobenious norm and trace of matrix $\boldsymbol{A}_{F}$. In particular, 
for the Frobenious norm bound, we have
\begin{align*}
    \|\boldsymbol{A}_{F}\|_F^2=&\sum_{j,k}(\boldsymbol{A}_{F})_{j,k}^2\\
    =&\frac{n^4}{N^4}\sum_{j,k}\left(\frac{1}{\binom{N-1}{n-1}^2}\sum_{j\in I, |I|=n}\sum_{k\in I', |I'|=n}\boldsymbol{X}_{j,F}(\boldsymbol{X}_{I,F}^\top \boldsymbol{X}_{I,F})^{-1}(\boldsymbol{X}_{I',F}^\top \boldsymbol{X}_{I',F})^{-1}\boldsymbol{X}_{k,F}^\top\right)^2\\
    \leq &\frac{n^4}{N^4}\frac{1}{\binom{N-1}{n-1}^2}\sum_{I\subset[N], |I|=n}\sum_{I'\subset[N], |I'|=n}\sum_{j\in I,k\in I'}\left(\boldsymbol{X}_{j,F}(\boldsymbol{X}_{I,F}^\top \boldsymbol{X}_{I,F})^{-1}(\boldsymbol{X}_{I',F}^\top \boldsymbol{X}_{I',F})^{-1}\boldsymbol{X}_{k,F}^\top\right)^2\\
    = &\frac{n^2}{N^2}\frac{1}{\binom{N}{n}^2}\sum_{I\subset[N], |I|=n}\sum_{I'\subset[N], |I'|=n}\left\|\boldsymbol{X}_{I,F}(\boldsymbol{X}_{I,F}^\top \boldsymbol{X}_{I,F})^{-1}(\boldsymbol{X}_{I',F}^\top \boldsymbol{X}_{I',F})^{-1}\boldsymbol{X}_{I',F}^\top\right\|_F^2\\
    \leq &\frac{n^2m}{N^2}\frac{1}{\binom{N}{n}^2}\sum_{I\subset[N], |I|=n}\sum_{I'\subset[N], |I'|=n}\sigma_{\min}^{-2}(\boldsymbol{X}_{I,F})\sigma_{\min}^{-2}(\boldsymbol{X}_{I',F})\\
    \leq&\frac{n^2m}{N^2}\left(\frac{1}{\binom{N}{n}}\sum_{I\subset[N], |I|=n}\sigma_{\min}^{-2}(\boldsymbol{X}_{I,F})\right)^2
\end{align*}
where we have applied the Jensen's inequality on the third line, and the fifth line utilizes the following arguments:
\begin{align*}
    &\left\|\boldsymbol{X}_{I,F}(\boldsymbol{X}_{I,F}^\top \boldsymbol{X}_{I,F})^{-1}(\boldsymbol{X}_{I',F}^\top \boldsymbol{X}_{I',F})^{-1}\boldsymbol{X}_{I',F}^\top\right\|_F^2\\
    =&\mathrm{tr}\left((\boldsymbol{X}_{I,F}^\top \boldsymbol{X}_{I,F})^{-1}(\boldsymbol{X}_{I',F}^\top \boldsymbol{X}_{I',F})^{-1}\right)\\
    \leq & m\|(\boldsymbol{X}_{I,F}^\top \boldsymbol{X}_{I,F})^{-1}\|_2\|(\boldsymbol{X}_{I',F}^\top \boldsymbol{X}_{I',F})^{-1}\|_2\\
    \leq & m\sigma_{\min}^{-2}(\boldsymbol{X}_{I,F})\sigma_{\min}^{-2}(\boldsymbol{X}_{I',F}).
\end{align*}
For the trace of $\boldsymbol{A}_{F}$, similar to the arguments above, we have
\begin{align*}
    \mathrm{tr}(\boldsymbol{A}_{F})=&\sum_{i=1}^N(\boldsymbol{A}_{F})_{i,i}\\
    =&\frac{1}{\binom{N}{n}^2}\sum_{i=1}^N\sum_{i\in I,|I|=n}\sum_{i\in I',|I'|=n}\boldsymbol{X}_{i,F}(\boldsymbol{X}_{I,F}^\top \boldsymbol{X}_{I,F})^{-1}(\boldsymbol{X}_{I',F}^\top \boldsymbol{X}_{I',F})^{-1}\boldsymbol{X}_{i,F}^\top\\
    \leq &\frac{1}{2\binom{N}{n}^2}\sum_{I\subset [N],|I|=n}\sum_{I'\subset [N],|I'|=n}\sum_{i\in I\cap I'}\|(\boldsymbol{X}_{I,F}^\top \boldsymbol{X}_{I,F})^{-1}\boldsymbol{X}_{i,F}\|_2^2+\|(\boldsymbol{X}_{I',F}^\top \boldsymbol{X}_{I',F})^{-1}\boldsymbol{X}_{i,F}^\top\|_2^2\\
    =&\frac{1}{\binom{N}{n}^2}\sum_{I\subset [N],|I|=n}\sum_{I'\subset [N],|I'|=n}\sum_{i\in I\cap I'}\|(\boldsymbol{X}_{I,F}^\top \boldsymbol{X}_{I,F})^{-1}\boldsymbol{X}_{i,F}\|_2^2\\
    =&\frac{1}{\binom{N}{n}^2}\sum_{I\subset [N],|I|=n}\sum_{I'\subset [N],|I'|=n}\|(\boldsymbol{X}_{I,F}^\top \boldsymbol{X}_{I,F})^{-1}\boldsymbol{X}_{I\cap I',F}\|_F^2\\
    = &\frac{n}N{}\frac{1}{\binom{N}{n}}\sum_{I\subset [N],|I|=n}\|(\boldsymbol{X}_{I,F}^\top \boldsymbol{X}_{I,F})^{-1}\boldsymbol{X}_{I,F}\|_F^2\\
    \leq &\frac{mn}{N}\frac{1}{\binom{N}{n}}\sum_{I\subset[N], |I|=n}\sigma_{\min}^{-2}(\boldsymbol{X}_{I,F}).
\end{align*}
Therefore, plugging in these bounds into \eqref{eq:MP_target_epsilon_1F} leads to 
$$
\|\varepsilon_{1,F}\|_2^2\leq C\frac{(m+\sqrt{m}\log N)n}{N}\frac{1}{\binom{N}{n}}\sum_{I\subset[N], |I|=n}\sigma_{\min}^{-2}(\boldsymbol{X}_{I,F})\|\beta_{F^c}^*\|_2^2,
$$
with probability at least $1-CN^{-c}$; and 
$$
\mathbb{E}(\|\varepsilon_{1,F}\|_2^2)\leq \frac{mn}{N}\mathbb{E}\left(\frac{1}{\binom{N}{n}}\sum_{I\subset[N], |I|=n}\sigma_{\min}^{-2}(\boldsymbol{X}_{I,F})\right)\|\beta_{F^c}^*\|_2^2,
$$
\paragraph{Bounding $\|\varepsilon_{2,F}\|_2^2$}: 
Similar to bounding $\|\varepsilon_{1,F}\|_2^2$, here we can write $\|\varepsilon_{2,F}\|_2^2=\epsilon^\top \boldsymbol{A}_{F}\epsilon$.
Since $\epsilon$ consists of $N$ independent sub-Gaussian noise of mean zero, we can also apply Hanson-Wright inequality to obtain
$$
\|\varepsilon_{2,F}\|_2^2\leq C\sigma_{\epsilon}^2\frac{(m+\sqrt{m}\log N)n}{N}\frac{1}{\binom{N}{n}}\sum_{I\subset[N], |I|=n}\sigma_{\min}^{-2}(\boldsymbol{X}_{I,F}),
$$
with probability at least $1-CN^{-c}$; and 
$$
\mathbb{E}(\|\varepsilon_{2,F}\|_2^2)\leq \sigma_{\epsilon}^2\frac{mn}{N}\mathbb{E}\left(\frac{1}{\binom{N}{n}}\sum_{I\subset[N], |I|=n}\sigma_{\min}^{-2}(\boldsymbol{X}_{I,F})\right),
$$
Recall the definition of $\lambda_{m,n}(\boldsymbol{X})$, $\lambda_{m,n}(\boldsymbol{X}_{:,\backslash j})$, $\lambda_{m,n}^{(j)}(\boldsymbol{X})$ and $\overline{\lambda}_{m,n}$ before Theorem \ref{prop:MP_target}.
Then, by Lemma \ref{lem:target_epsilon_decomposed}, we have with probability at least $1-CN^{-c}$ that
\begin{equation*}
    \begin{split}
        \|\varepsilon_{1,l}\|_2^2\leq &C\gamma \lambda_{n,m}(\boldsymbol{X})\frac{m+\sqrt{m}\log N}{N}\|\beta^*\|_2^{2\ind_{l=1}}\sigma_{\epsilon}^{2\ind_{l=1}},\\
        \|\varepsilon_{2,l}\|_2^2\leq &C\frac{m}{M-1} \lambda_{n,m}(\boldsymbol{X}_{:,\backslash j})\frac{m+\sqrt{m}\log N}{N}\|\beta^*\|_2^{2\ind_{l=1}}\sigma_{\epsilon}^{2\ind_{l=2}},\\
        \|(\varepsilon_{3,l})_{\backslash j}\|_2^2\leq &C\frac{m-1}{M-1} \lambda_{n,m}^{(j)}(\boldsymbol{X})\frac{m+\sqrt{m}\log N}{N}\|\beta^*\|_2^{2\ind_{l=1}}\sigma_{\epsilon}^{2\ind_{l=2}},\\
        |(\varepsilon_{3,l})_{j}|^2\leq &C\lambda_{n,m}^{(j)}(\boldsymbol{X})\frac{m+\sqrt{m}\log N}{N}\|\beta^*\|_2^{2\ind_{l=1}}\sigma_{\epsilon}^{2\ind_{l=2}}.
    \end{split}
\end{equation*}
Now we recall \eqref{eq:MP_target_basic_bnd}, and immediately we have
\begin{equation}\label{eq:MP_target_final_bnd}
    \begin{split}
        &\left|\Delta_j^*-\Delta_j^{*(L)}\right|\\
    \leq&C\sqrt{\overline{\lambda}_{m,n}}(\|\beta^*\|_2+\sigma_{\epsilon})(\sqrt{\gamma}\|\beta^*\|_2+|\beta_j^*|)\frac{m(m+\sqrt{m}\log N)^{\frac{1}{2}}}{M\sqrt{N}}\\
    &+C\overline{\lambda}_{m,n}(\|\beta^*\|_2^2+\sigma_{\epsilon}^2)\frac{m(m+\sqrt{m}\log N)}{MN}.
    \end{split}
\end{equation}
The proof is now complete.
\end{proof}

\begin{proof}[Proof of Theorem \ref{prop:coverage_tilde_Delta_full}]
    First we note that
    \begin{align*}
        \bP(\Delta_j^{*(\mathrm{L})}\in \hat{\mathbb{C}}_j^{\mathrm{barrier}})=&\bP(\frac{|\bar{\Delta}_j - \Delta_j^{*(\mathrm{L})}|}{\hat{\sigma}_j/\sqrt{N} + \epsilon(N)}\leq z_{\alpha/2})\\
        \geq &\bP(\frac{|\bar{\Delta}_j - \Delta_j^*|}{\hat{\sigma}_j/\sqrt{N} + \epsilon(N)}+\frac{N|\Delta_j^* - \Delta_j^{*(\mathrm{L})}|}{cLBn\log N}\leq z_{\alpha/2}),
    \end{align*}
    where the last line is due to $\epsilon(N)\geq \frac{cLBn}{N}\log N$ in Assumption \ref{assump:buffer}. In fact, using the same proof as those of Theorem \ref{thm:coverage_buffer} (skipping the bound for $\varepsilon_j^{(4)}$), one can immediately show that $\liminf_{N\rightarrow \infty}\bP(\frac{|\bar{\Delta}_j - \Delta_j^*|}{\hat{\sigma}_j/\sqrt{N} + \epsilon(N)}\leq z_{\alpha/2})\geq 1-\alpha$. While for dealing with the error term $\frac{N|\Delta_j - \Delta_j^{*(\mathrm{L})}|}{cLBn\log N}$, we would like to apply Theorem \ref{prop:MP_target_full}. When $\|\beta^*\|_2\leq C$, with probability at least $1-N^{-c}$, 
    $$
    |\Delta_j^*-\Delta_j^{*(\mathrm{L})}|\leq C\frac{\sqrt{\overline{\lambda}_{m,n}}\gamma(m+\sqrt{m}\log N)^{\frac{1}{2}}}{\sqrt{N}}+C\frac{\overline{\lambda}_{m,n}\gamma(m+\sqrt{m}\log N)}{N}.
    $$
    Since \eqref{eq:gamma_cond_coverage} implies
    \begin{equation*}
    \gamma\ll LB\min\left\{\frac{n\log N}{\sqrt{\overline{\lambda}_{m,n}N(m+\sqrt{m}\log N)}},\frac{n\log N}{\overline{\lambda}_{m,n}(m+\sqrt{m}\log N)}\right\},
\end{equation*}
   we have
    \begin{align*}
        \frac{N|\Delta_j^* - \Delta_j^{*(\mathrm{L})}|}{cLBn\log N}\leq C\gamma\frac{\sqrt{\overline{\lambda}_{m,n}N(m+\sqrt{m}\log N)}+\overline{\lambda}_{m,n}(m+\sqrt{m}\log N)}{LBn\log N}.
    \end{align*}
    Therefore, $\liminf_{n\rightarrow \infty}\bP(\Delta_j^{*(\mathrm{L})}\in \hat{\mathbb{C}}_j^{\mathrm{barrier}})=1-\alpha$.
\end{proof}
\begin{proof}[Proof of Proposition \ref{prop:MP_target_corr}]
Most of our proof of Proposition \ref{prop:MP_target_corr} follows similar arguments to the proof of Theorem \ref{prop:MP_target_full}, except that for each minipatch $I,F$, we will decorrelate $\boldsymbol{X}_{I,F^c}$ and $\boldsymbol{X}_{I,F}$, and consider the effect of out-of-patch features $F^c$ upon in-patch features $F$; We will focus on the main steps that are different, and omit the repeated steps. 

\paragraph{Concentration of $\bE\Delta_j^*$} Recall that we have defined $\Delta_j^* =\lim_{K\rightarrow \infty}\Delta_j$ in the proof of Theorem \ref{prop:MP_target_full}, and have shown that $\Delta_j^*=\bE_{X,Y}[\err(Y,\mu^*(X;\bX,\bY)-\err(Y,\mu^*_{\backslash j}(X;\bX,\bY)|\bX,\bY]$. Now we show that $\bE\Delta_j^*$ is close to $\Delta_j^{*(L)}$. In particular, for a given minipatch $(I,F)$, let $\tilde{\boldsymbol{X}}_{I,F^c}=\boldsymbol{X}_{I,F^c} - \boldsymbol{X}_{I,F}\boldsymbol{\Sigma}_{F,F}^{-1}\boldsymbol{\Sigma}_{F,F^c}$, which satisfies $\mathbb{E}(\boldsymbol{X}_{I,F}^\top \tilde{\boldsymbol{X}}_{I,F^c}) = 0$. Then we can write
\begin{equation*}
\begin{split}
    H(\boldsymbol{X}_{I,F},\boldsymbol{Y}_{I})(X_F) = &X^\top R_F^\top[\beta_F^* + (\boldsymbol{X}_{I,F}^\top \boldsymbol{X}_{I,F})^{-1}\boldsymbol{X}_{I,F}^\top (\boldsymbol{X}_{I,F^c}\beta^*_{F^c} + \epsilon_I)]\\
    = &X^\top R_F^\top [\beta_F^* + \boldsymbol{\Sigma}_{F,F}^{-1}\boldsymbol{\Sigma}_{F,F^c}\beta_{F^c}^*+(\boldsymbol{X}_{I,F}^\top \boldsymbol{X}_{I,F})^{-1}\boldsymbol{X}_{I,F}^\top (\tilde{\boldsymbol{X}}_{I,F^c}\beta^*_{F^c} + \epsilon_I)].
\end{split}
\end{equation*}
Let $$\beta^{(m)*} = \frac{m}{M}\beta^*+\frac{1}{\binom{M}{m}}\sum_{F\subset [M]}R_F^\top \boldsymbol{\Sigma}_{F,F}^{-1}\boldsymbol{\Sigma}_{F,F^c}\beta_{F^c}^*,
$$
$$
\beta^{(m,-j)*} = \frac{m}{M-1}\beta^{*\backslash j}+\frac{1}{\binom{M-1}{m}}\sum_{F\subset [M],j\notin F}R_F^\top \boldsymbol{\Sigma}_{F,F}^{-1}\boldsymbol{\Sigma}_{F,F^c}\beta_{F^c}^*,
$$ 
with $\beta^{*\backslash j}\in \mathbb{R}^M$ satisfying $\beta^{*\backslash j}_j=0$ and $\beta^{*\backslash j}_{\backslash j}=\beta^*_{\backslash j}$. By the definition of $\mu^*(X;\boldsymbol{X},\boldsymbol{Y})$ and $\mu^*_{\backslash j}(X_{\backslash j};\boldsymbol{X}_{:,\backslash j},\boldsymbol{Y})$, we further have
\begin{align*}
    &\mu^*(X;\boldsymbol{X},\boldsymbol{Y})\\
    =&X^\top \left\{\beta^{(m)*}+\frac{1}{\binom{N}{n}\binom{M}{m}}\sum_{I\subset[N],F\subset [M]}R_F^\top[(\boldsymbol{X}_{I,F}^\top \boldsymbol{X}_{I,F})^{-1}\boldsymbol{X}_{I,F}^\top (\tilde{\boldsymbol{X}}_{I,F^c}\beta^*_{F^c} + \epsilon_I)]\right\},
\end{align*}
and 
\begin{align*}
    &\mu^*_{\backslash j}(X_{\backslash j};\boldsymbol{X}_{:,\backslash j},\boldsymbol{Y})\\
    =&X^\top \left\{\beta^{(m,-j)*} + \frac{1}{\binom{N}{n}\binom{M-1}{m}}\sum_{I\subset[N],j\notin F}R_F^\top[(\boldsymbol{X}_{I,F}^\top \boldsymbol{X}_{I,F})^{-1}\boldsymbol{X}_{I,F}^\top (\tilde{\boldsymbol{X}}_{I,F^c}\beta^*_{F^c} + \epsilon_I)]\right\}.
\end{align*}
Similar to the proof of Theorem \ref{prop:MP_target_full}, here we define $$
\varepsilon_{1,F} = \frac{1}{\binom{N}{n}}\sum_{I\subset[N],|I|=n}R_F^\top(\boldsymbol{X}_{I,F}^\top \boldsymbol{X}_{I,F})^{-1}\boldsymbol{X}_{I,F}^\top \tilde{\boldsymbol{X}}_{I,F^c}\beta^*_{F^c},
$$
$$
    \varepsilon_{2,F} = \frac{1}{\binom{N}{n}}\sum_{I\subset[N],|I|=n}R_F^\top(\boldsymbol{X}_{I,F}^\top \boldsymbol{X}_{I,F})^{-1}\boldsymbol{X}_{I,F}^\top \epsilon_I.
$$
Also define the following six $M$-dimensional error terms:
\begin{equation*}
    \begin{split}
        \varepsilon_{1,1} = &\frac{1}{\binom{M}{m}}\sum_{F\subset [M], |F|=m}\varepsilon_{1,F}, \quad \varepsilon_{1,2} =  \frac{1}{\binom{M}{m}}\sum_{F\subset [M], |F|=m}\varepsilon_{2,F},\\
    \varepsilon_{2,1} = &\frac{1}{\binom{M-1}{m}}\sum_{j\notin F, |F|=m}\varepsilon_{1,F}, \quad \varepsilon_{2,2} =  \frac{1}{\binom{M-1}{m}}\sum_{j\notin F, |F|=m}\varepsilon_{2,F}.
    \end{split}
\end{equation*}
Since $\mathbb{E}(\epsilon|\boldsymbol{X})=0$, $\mathbb{E}(\boldsymbol{X}_{I,F}^\top \tilde{\boldsymbol{X}}_{I,F^c})=0$, all these six error terms are of mean zero. Then we can decompose our inference target as follows:
\begin{equation}\label{eq:Delta_decompose_corr}
    \begin{split}
    \Delta_j^* = &\mathbb{E}_{X,Y}\left(Y-X^\top\left(\beta^{(m,-j)*} + \varepsilon_{2,1} + \varepsilon_{2,2}\right)\right)^2 - \mathbb{E}_{X,Y}\left(Y-X^\top\left(\beta^{(m)*} + \varepsilon_{1,1} + \varepsilon_{1,2}\right)\right)^2\\
    = &\left\|\beta^* - \beta^{(m,-j)*} - \varepsilon_{2,1} - \varepsilon_{2,2}\right\|_{\boldsymbol{\Sigma}}^2-\left\|\beta^* - \beta^{(m)*} - \varepsilon_{1,1} - \varepsilon_{1,2}\right\|_{\boldsymbol{\Sigma}}^2\\
    =&\left\|\beta^* - \beta^{(m,-j)*}\right\|_{\boldsymbol{\Sigma}}^2-\left\|\beta^* - \beta^{(m)*}\right\|_{\boldsymbol{\Sigma}}^2
    -2(\varepsilon_{2,1}+\varepsilon_{2,2})^\top\boldsymbol{\Sigma} (\beta^* - \beta^{(m,-j)*})\\
    &+2(\varepsilon_{1,1}+\varepsilon_{1,2})^\top\boldsymbol{\Sigma} (\beta^* - \beta^{(m)*})+\|\varepsilon_{1,1}+\varepsilon_{1,2}\|_{\boldsymbol{\Sigma}}^2+ \|\varepsilon_{2,1}+\varepsilon_{2,2}\|_{\boldsymbol{\Sigma}}^2.
\end{split}
\end{equation}
Recall our definition of $\Delta_j^{*(L)}=\left\|\beta^* - \beta^{(m,-j)*}\right\|_{\boldsymbol{\Sigma}}^2-\left\|\beta^* - \beta^{(m)*}\right\|_{\boldsymbol{\Sigma}}^2$, we have
\begin{equation*}
    \begin{split}
        |\bE\Delta_j^* - \Delta_j^{*(\mathrm{L})}|= &\mathbb{E}\|\varepsilon_{1,1}+\varepsilon_{1,2}\|_{\boldsymbol{\Sigma}}^2+\mathbb{E}\|\varepsilon_{2,1}+\varepsilon_{2,2}\|_{\boldsymbol{\Sigma}}^2\\
        \leq&2\lambda_{\max}(\boldsymbol{\Sigma})\sum_{k,l=1}^2\|\varepsilon_{k,l}\|_2^2.
    \end{split}
\end{equation*}
Using the same arguments as in the proof of Lemma \ref{lem:target_epsilon_decomposed}, one can further bound $\|\varepsilon_{k,l}\|_2^2$ by the average $\ell_2$ errors of $\varepsilon_{l,F}$, with exactly the same form as in Lemma \ref{lem:target_epsilon_decomposed}. Hence we can bound $|\Delta_j^* - \Delta_j^{*(\mathrm{L})}|$ as follows:
\begin{equation*}
    \begin{split}
        |\bE\Delta_j^* - \Delta_j^{*(\mathrm{L})}|\leq &2\lambda_{\max}(\boldsymbol{\boldsymbol{\Sigma}})\frac{m}{M\binom{M}{m}}\sum_{F\subset [M],|F|=m}(\mathbb{E}\|\varepsilon_{1,F}\|_2^2+\mathbb{E}\|\varepsilon_{2,F}\|_2^2)\\
        &+2\lambda_{\max}(\boldsymbol{\Sigma})\frac{m}{(M-1)\binom{M-1}{m}}\sum_{j\notin F,|F|=m}(\mathbb{E}\|\varepsilon_{1,F}\|_2^2+\mathbb{E}\|\varepsilon_{2,F}\|_2^2).
    \end{split}
\end{equation*}
Now our proof hinges on upper bounds of $\mathbb{E}\|\varepsilon_{1,F}\|_2^2$ and $\mathbb{E}\|\varepsilon_{2,F}\|_2^2$. Recall our definition of matrix $\boldsymbol{A}_F$ in the proof of Theorem \ref{prop:MP_target_full}. We can then write
\begin{equation*}
\begin{split}
    \|\varepsilon_{1,F}\|_2^2 =& (\tilde{\boldsymbol{X}}_{:,F^c}\beta_{F^c}^*)^\top \boldsymbol{A}_F(\tilde{\boldsymbol{X}}_{:,F^c}\beta_{F^c}^*),\\
    \|\varepsilon_{2,F}\|_2^2 =& \epsilon^\top \boldsymbol{A}_F\epsilon.
\end{split}
\end{equation*}
Conditioning on $\boldsymbol{X}_{:,F}$, $\tilde{\boldsymbol{X}}_{:,F^c}\beta_{F^c}^*$ are entry-wise independent Gaussian random variables with mean zero, variance $\beta_{F^c}^{*\top}(\boldsymbol{\Sigma}_{F,F}-\boldsymbol{\Sigma}_{F,F^c}\boldsymbol{\Sigma}_{F^c,F^c}^{-1}\boldsymbol{\Sigma}_{F^c,F})\beta_{F^c}^*\leq \lambda_{\max}(\Sigma)\|\beta_{F^c}^*\|_2^2$. While for the noise $\epsilon$, we have assumed it to be entry-wise independent with mean zero and variance $\sigma_{\epsilon}^2$. As has been shown in the proof of Theorem \ref{prop:MP_target_full}, $\mathrm{tr}(\boldsymbol{A}_F)\leq \frac{mn}{N}\frac{1}{\binom{N}{n}}\sum_{I\subset[N],|I|=n}\sigma_{\min}^{-2}(\boldsymbol{X}_{I,F})$, which then implies
\begin{equation*}
\begin{split}
    \mathbb{E}\|\varepsilon_{1,F}\|_2^2 
    \leq& \lambda_{\max}(\Sigma)\|\beta_{F^c}^*\|_2^2\mathbb{E}(\mathrm{tr}(\boldsymbol{A}_F))\\
    \leq&\lambda_{\max}(\Sigma)\|\beta_{F^c}^*\|_2^2\frac{mn}{N}\frac{1}{\binom{N}{n}}\sum_{I\subset[N],|I|=n}\mathbb{E}\sigma_{\min}^{-2}(\boldsymbol{X}_{I,F}),\\
    \|\varepsilon_{2,F}\|_2^2 \leq & \sigma_{\epsilon}^2\frac{mn}{N}\frac{1}{\binom{N}{n}}\sum_{I\subset[N],|I|=n}\mathbb{E}\sigma_{\min}^{-2}(\boldsymbol{X}_{I,F}).
\end{split}
\end{equation*}
Therefore, we immediately have
\begin{equation*}
    \begin{split}
        |\Delta_j^* - \Delta_j^{*(\mathrm{L})}|\leq &2\lambda_{\max}(\boldsymbol{\boldsymbol{\Sigma}})(\lambda_{\max}(\Sigma)\|\beta_{F^c}^*\|_2^2+\sigma_{\epsilon}^2)\frac{m}{N}\left[\frac{m}{M}\mathbb{E}\lambda_{m,n}(\boldsymbol{X})+\frac{m}{M-1}\mathbb{E}\lambda_{m,n}(\boldsymbol{X_{:,\backslash j}})\right].
    \end{split}
\end{equation*}
\paragraph{$\Delta_j^{*(\mathrm{L})}$ in a special example} 
Now it remains to show the simplified form of $\Delta_1^{*(\mathrm{L})}$ when $\boldsymbol{\Sigma}$ is as specified in Proposition \ref{prop:MP_target_corr}. In this setting, one can immediately see that
\begin{equation*}
    R_F^\top\boldsymbol{\Sigma}_{F,F}^{-1}\boldsymbol{\Sigma}_{F,F^c}R_{F^c} = \begin{cases}
    \boldsymbol{0}_{M\times M},&\text{ if }1,2\in F\text{ or }1,2\in F^c,\\
    \rho \boldsymbol{e}^{(1,2)},&\text{ if }1\in F, 2\in F^c,\\
    \rho \boldsymbol{e}^{(2,1)},&\text{ if }1\in F^c, 2\in F,\\
    \end{cases}
\end{equation*}
where $\boldsymbol{e}^{(1,2)}$, $\boldsymbol{e}^{(2,1)}\in \mathbb{R}^{M\times M}$ are matrices with all entries being zero except for one entry: $\boldsymbol{e}^{(1,2)}_{1,2}=1$, $\boldsymbol{e}^{(2,1)}_{2,1}=1$. Hence we can write 
\begin{equation*}
\begin{split}
    \beta^*-\beta^{(m)*} =& \beta^*-(\gamma\beta^*+\gamma'(1-\gamma)\rho(\beta_2^*,\beta_1^*,0,\dots,0)^\top)\\
    =&(1-\gamma)\begin{pmatrix}
     1&-\gamma'\rho&0&\cdots&0\\
     -\gamma'\rho&1&0&\cdots&0\\
     0&0&1&\cdots&0\\
     \vdots&\vdots&\ddots&\ddots&\vdots\\
     0&\cdots&\cdots&\cdots&1
     \end{pmatrix}\beta^*
\end{split}
\end{equation*}
where we denote $\frac{m}{M-1}$ by $\gamma'$; and 
\begin{equation*}
\begin{split}
     \beta^*-\beta^{(m,-1)*} = &\beta^*-(\gamma'\beta^{*\backslash 1}+\gamma'\rho(0,\beta_1^*,0,\dots,0)^\top)\\
     =&(\beta_1^*,(1-\gamma')\beta_2^*-\gamma'\rho\beta_1^*,(1-\gamma')\beta^*_{\backslash (1,2)})^\top\\
     =&\begin{pmatrix}
     1&0&0&\cdots&0\\
     -\gamma'\rho&1-\gamma'&0&\cdots&0\\
     0&0&1-\gamma'&\cdots&0\\
     \vdots&\vdots&\ddots&\ddots&\vdots\\
     0&\cdots&\cdots&\cdots&1-\gamma'
     \end{pmatrix}\beta^*.
\end{split}
\end{equation*}
By the definition of $\Delta_j^{*(\mathrm{L})}$, we have
\begin{equation}\label{eq:tilde_Delta_corr}
    \begin{split}
        \Delta_j^{*(\mathrm{L})}=&\beta_{(1,2)}^{*\top}\begin{pmatrix}
        1&-\gamma'\rho\\
        0&1-\gamma'\end{pmatrix}\begin{pmatrix}
        1&\rho\\
        \rho&1\end{pmatrix}\begin{pmatrix}
        1&0\\
        -\gamma'\rho&1-\gamma'\end{pmatrix}\beta_{(1,2)}^*\\
        &-(1-\gamma)^2\beta_{(1,2)}^{*\top}\begin{pmatrix}
        1&-\gamma'\rho\\
        -\gamma'\rho&1\end{pmatrix}\begin{pmatrix}
        1&\rho\\
        \rho&1\end{pmatrix}\begin{pmatrix}
        1&-\gamma'\rho\\
        -\gamma'\rho&1\end{pmatrix}\beta_{(1,2)}^*\\
        &+[(1-\gamma')^2-(1-\gamma)^2]\|\beta^*_{\backslash (1,2)}\|_2^2\\
        =&(2\gamma-\gamma^2)(1-2\gamma'\rho^2+\gamma'^2\rho^2)\beta_1^{*2}+[(-2\gamma'+\gamma'^2)(1-\rho^2)+(2\gamma-\gamma^2)(1-2\gamma'\rho^2+\gamma'^2\rho^2)]\beta_2^{*2}\\
        &+2[\gamma'^2(1-\rho^2)\rho+(2\gamma-\gamma^2)(1-2\gamma'+\gamma'^2\rho^2)\rho]\beta_1^*\beta_2^*-(\gamma^2-\gamma'^2-2\gamma+2\gamma')\|\beta^*_{\backslash (1,2)}\|_2^2.
    \end{split}
\end{equation}
When $\rho=0$, we have 
$$
\Delta_j^{*(\mathrm{L})} = \gamma(2-\gamma)\beta_1^{*2}-\gamma\left(\frac{2}{M-1}-\frac{m(2M-1)}{(M-1)^2M}\right)\|\beta^*_{\backslash 1}\|_2^2;
$$
when $\rho=1$, we have
$$
\Delta_j^{*(\mathrm{L})} = \gamma(2-\gamma)(1-\gamma')^2(\beta_1^*+\beta_2^*)^2-\gamma\left(\frac{2}{M-1}-\frac{m(2M-1)}{(M-1)^2M}\right)\|\beta^*_{\backslash (1,2)}\|_2^2.
$$
\end{proof}
\newpage

 \section{Additional Empirical Details and Results}\label{sec:additional_numeric}
 
A Python package of our proposed method is available online: 
\href{https://github.com/DataSlingers/LOCOMP}{https://github.com/DataSlingers/LOCOMP}. 

\subsection{Validating Theoretical Coverage}

\begin{figure}
\centering
    \includegraphics[clip,width=\textwidth]{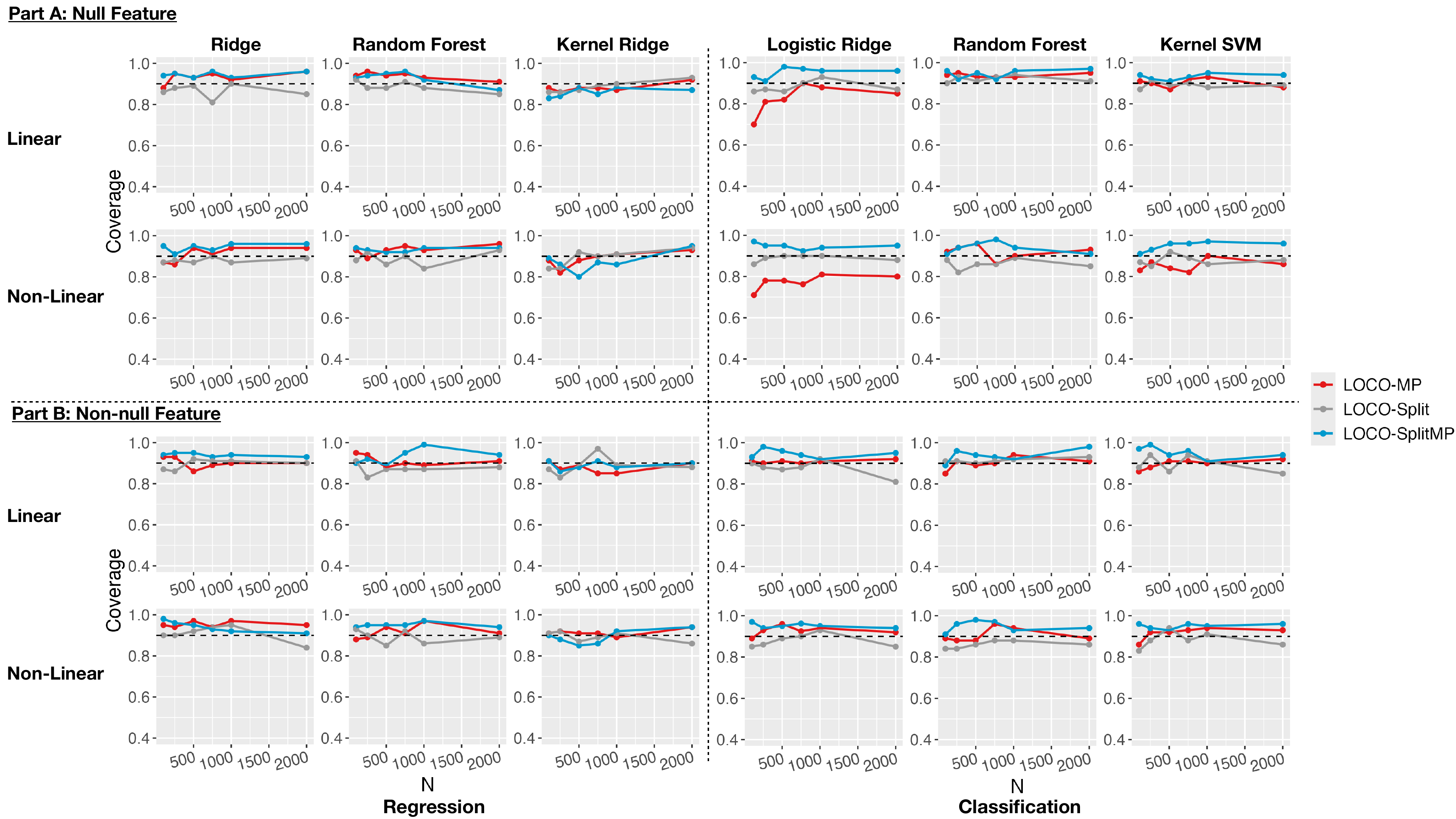}
\caption{\label{fig:validate5} Coverage of the inference target \eqref{eq:target_inference} for null and non-null features in 90\% confidence intervals based on synthetic data, with minipatch sizes $n = N^{0.8}$, and $m = 0.5M$. No buffer constant is applied to all the settings. 
Part A: Coverage for a null feature, with rows corresponding to linear and non-linear simulations, and columns representing different base estimators: ridge regression, decision tree, and kernel SVM, with no buffer constant applied. Left panels display results for regression tasks, while right panels show results for classification tasks. Part B: Coverage for a non-null feature with a SNR of 2, under the same setup of linear and non-linear simulations (rows) and base estimators (columns). No buffer constant is applied. Left panels correspond to regression tasks, and right panels correspond to classification tasks.}
\end{figure}

\begin{figure}
\centering
    \includegraphics[clip,width=\textwidth]{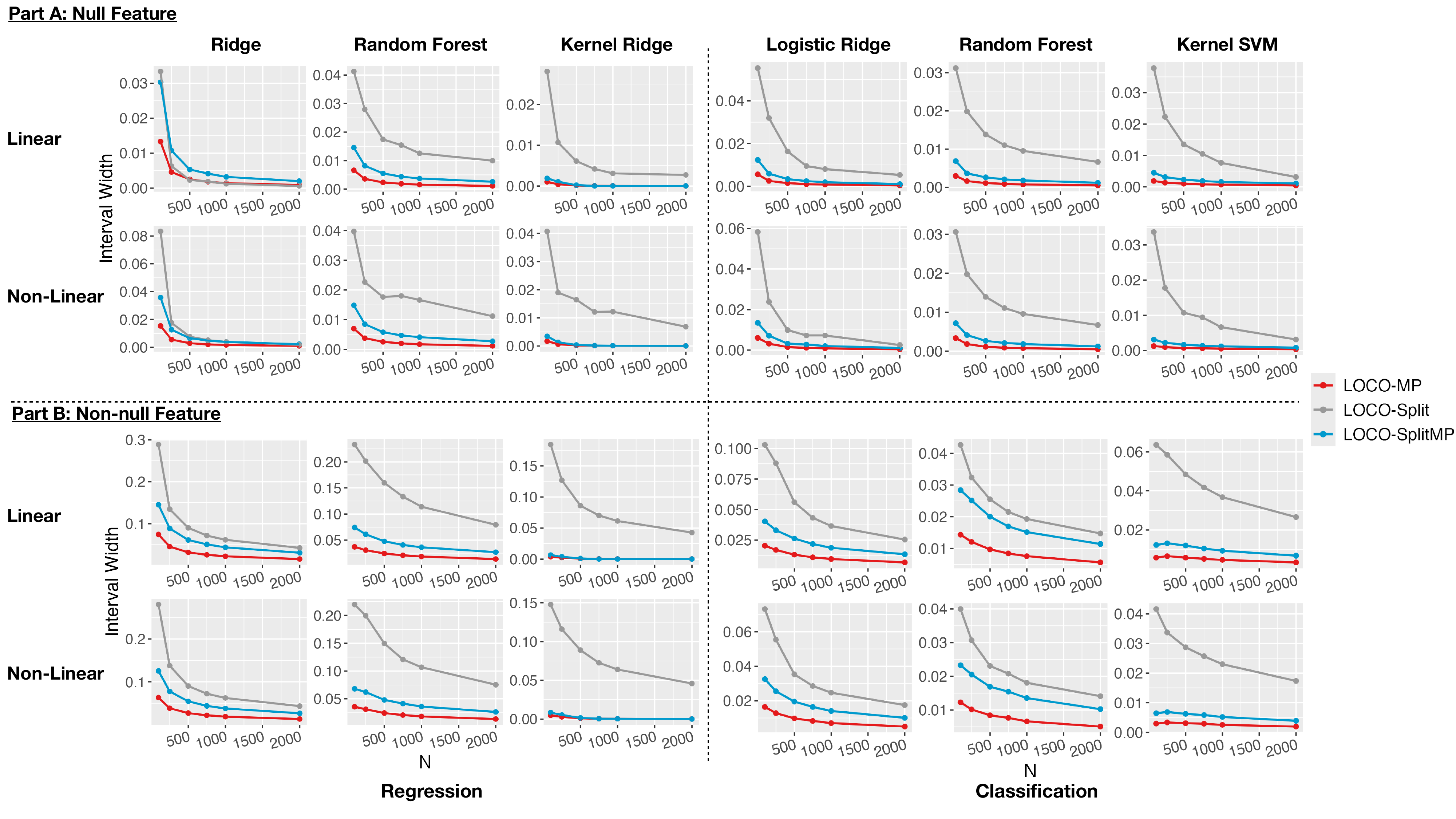}
\caption{\label{fig:validate6} Interval width for the inference target \eqref{eq:target_inference} in 90\% confidence intervals with minipatch sizes $n = N^{0.8}$, and $m = 0.5M$, evaluated on synthetic data. No buffer constant is applied to all the settings. 
Part A: Interval width for a null feature, with rows representing linear and non-linear simulations, and columns corresponding to different base estimators: ridge regression, decision tree, and kernel SVM. Left panels illustrate results for regression tasks, while right panels present results for classification tasks.
Part B: Interval width for a non-null feature with a signal-to-noise ratio (SNR) of 2, using the same arrangement of linear and non-linear simulations (rows) and base estimators (columns). Left panels show results for regression tasks, and right panels show results for classification tasks. LOCO-MP achieves the smallest interval width, which decreases as the sample size $N$ increases.}
\end{figure}

Since our inference target $\Delta_j$ in \eqref{eq:target_inference} involves expectation and hence is hard to compute, we use Monte Carlo approximation with $10,000$ test data points. In particular, we evaluate our trained machine learning model $\mu$ and $\mu_{\backslash j}$ (taking the form of minipatch ensembles in LOCO-MP and LOCO-SplitMP) on the test data, computing the prediction error difference between $\mu$ and $\mu_{\backslash j}$, averaged over the test data set. This approximation serves as a surrogate of the expectation in \eqref{eq:target_inference}. 
Figure~\ref{fig:validate5} and Figure~\ref{fig:validate6} show the coverage and width without adding any buffer, under the same setting as the main paper. 

\begin{figure}
\centering
    \includegraphics[clip,width=\textwidth]{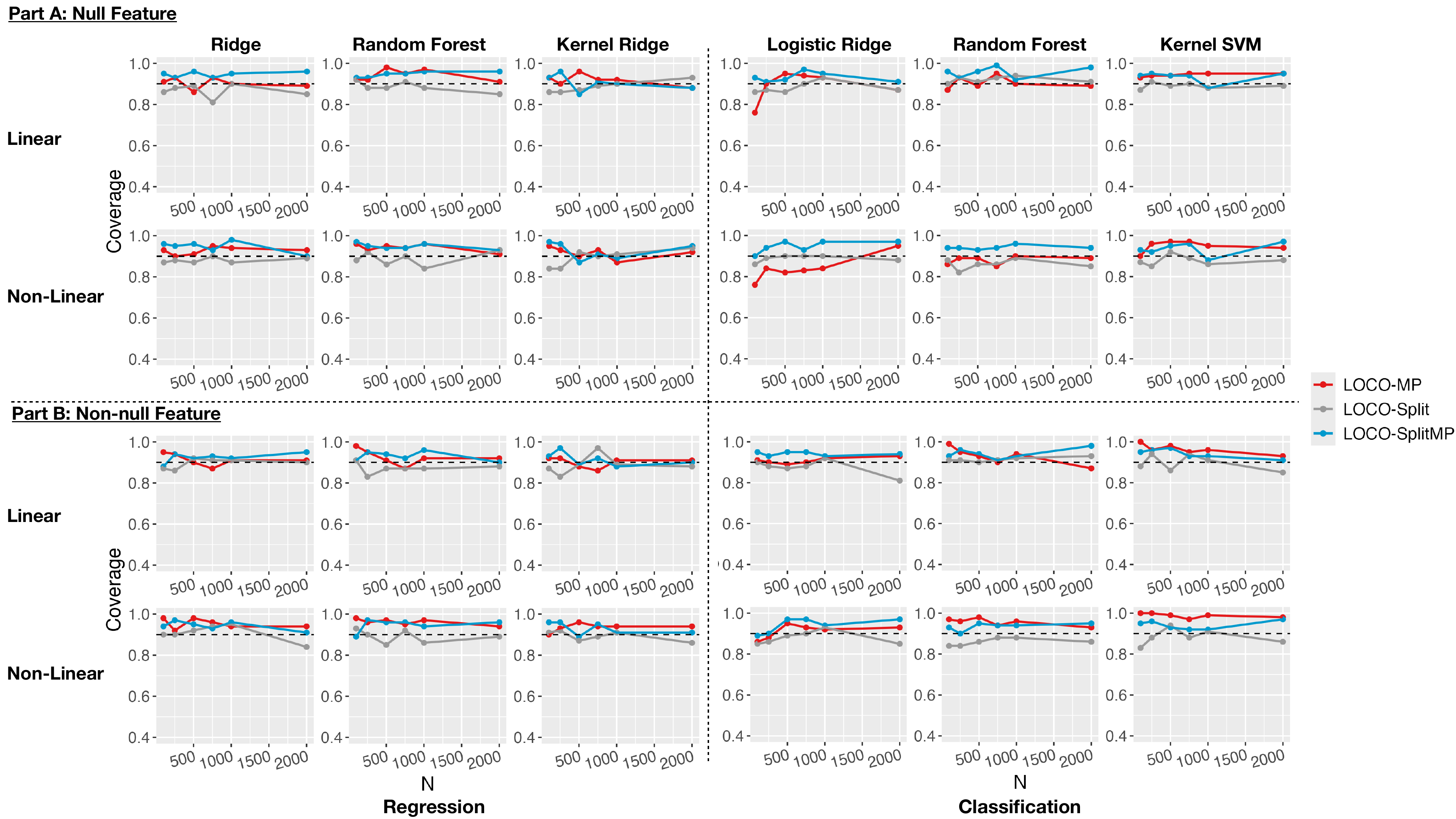}
\caption{\label{fig:validate3} Coverage of the inference target \eqref{eq:target_inference} for null and non-null features in 90\% confidence intervals based on synthetic data, with minipatch sizes $n = \sqrt{N}$, and $m = \sqrt{M}$.
Part A: Coverage for a null feature, with rows corresponding to linear and non-linear simulations, and columns representing different base estimators: ridge regression, decision tree, and kernel SVM, with no buffer constant applied. Left panels display results for regression tasks, while right panels show results for classification tasks. Part B: Coverage for a non-null feature with a SNR of 2, under the same setup of linear and non-linear simulations (rows) and base estimators (columns). No buffer constant is applied. Left panels correspond to regression tasks, and right panels correspond to classification tasks.}
\end{figure}

\begin{figure}
\centering
    \includegraphics[clip,width=\textwidth]{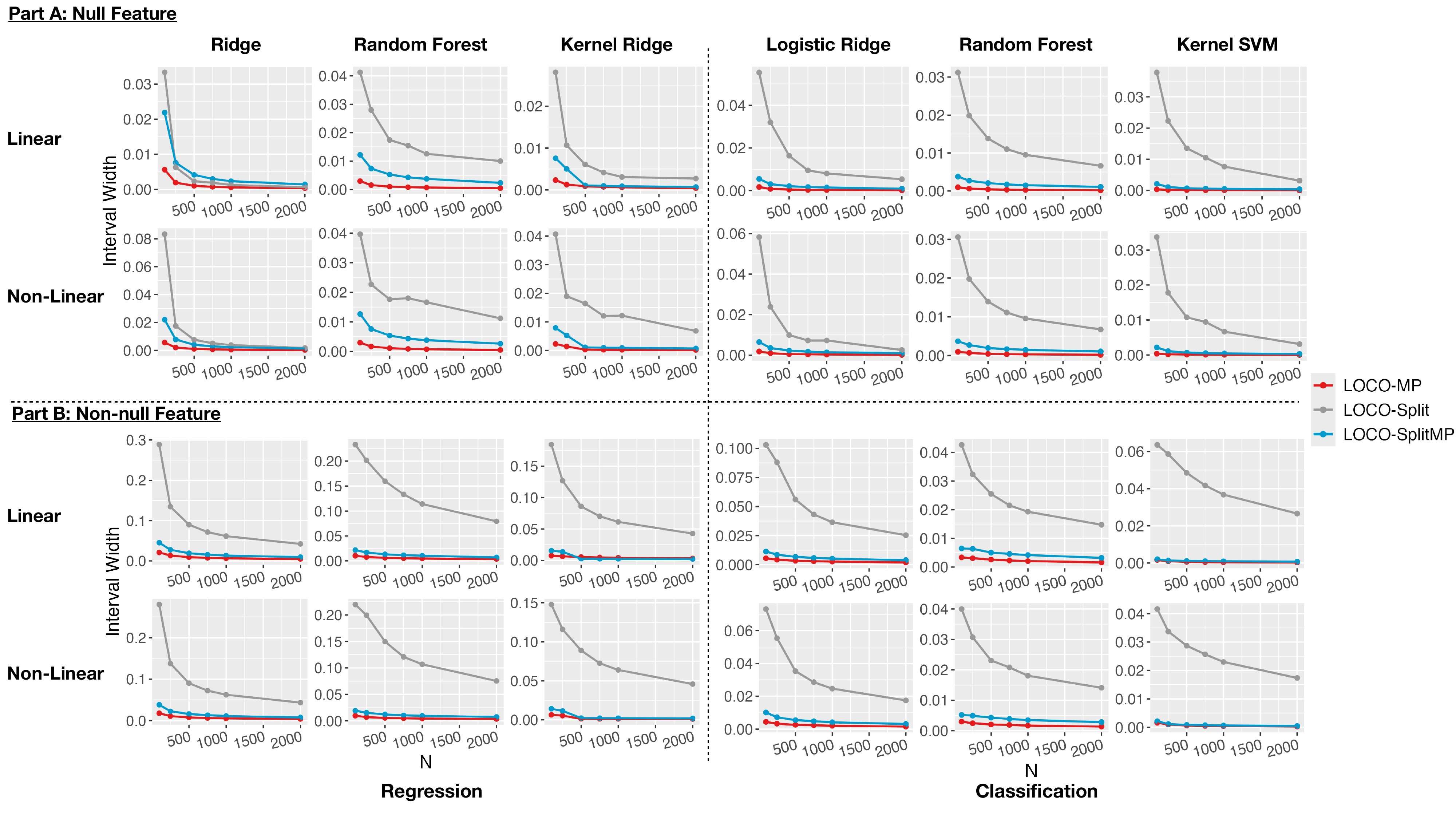}
\caption{\label{fig:validate4} Interval width for the inference target \eqref{eq:target_inference} in 90\% confidence intervals with minipatch sizes $n = \sqrt{N}$, and $m = \sqrt{M}$, evaluated on synthetic data. 
Part A: Interval width for a null feature, with rows representing linear and non-linear simulations, and columns corresponding to different base estimators: ridge regression, decision tree, and kernel SVM. A buffer constant $c=0.005$ is applied. Left panels illustrate results for regression tasks, while right panels present results for classification tasks.
Part B: Interval width for a non-null feature with a signal-to-noise ratio (SNR) of 2, using the same arrangement of linear and non-linear simulations (rows) and base estimators (columns). No buffer constant is applied. Left panels show results for regression tasks, and right panels show results for classification tasks. LOCO-MP achieves the smallest interval width, which decreases as the sample size $N$ increases.}
\end{figure}

We additionally evaluate the feature importance confidence intervals generated by LOCO-MP in terms of valid coverage for the inference target as well as interval width under various scenarios, with a different minipatch size setting that $n = \sqrt{N}$, and $m = \sqrt{M}$. With all other settings the same as stated in the main paper, Figure~\ref{fig:validate3} and Figure~\ref{fig:validate4} demonstrate that LOCO-MP exhibits valid coverage rates for both null feature and signal feature with $SNR = 2$ and generates efficient intervals with width decreasing as $N$ increases, with no buffer added.

\subsection{Additional Comparative Results}

We present additional results in Figure~\ref{fig:loco_interval1} and Figure~\ref{fig:loco_interval2} on the comparison of 90\% confidence intervals constructed with $n = N^{0.8}, m = 0.5M$, with all other settings the same as the main paper. Figure~\ref{fig:loco_interval2} shows the same confidence intervals excluding LOCO-Spilt to demonstrate a clearer comparison between LOCO-MP and LOCO-SplitMP. 

\begin{figure}
\centering
    \includegraphics[clip,width=\textwidth]{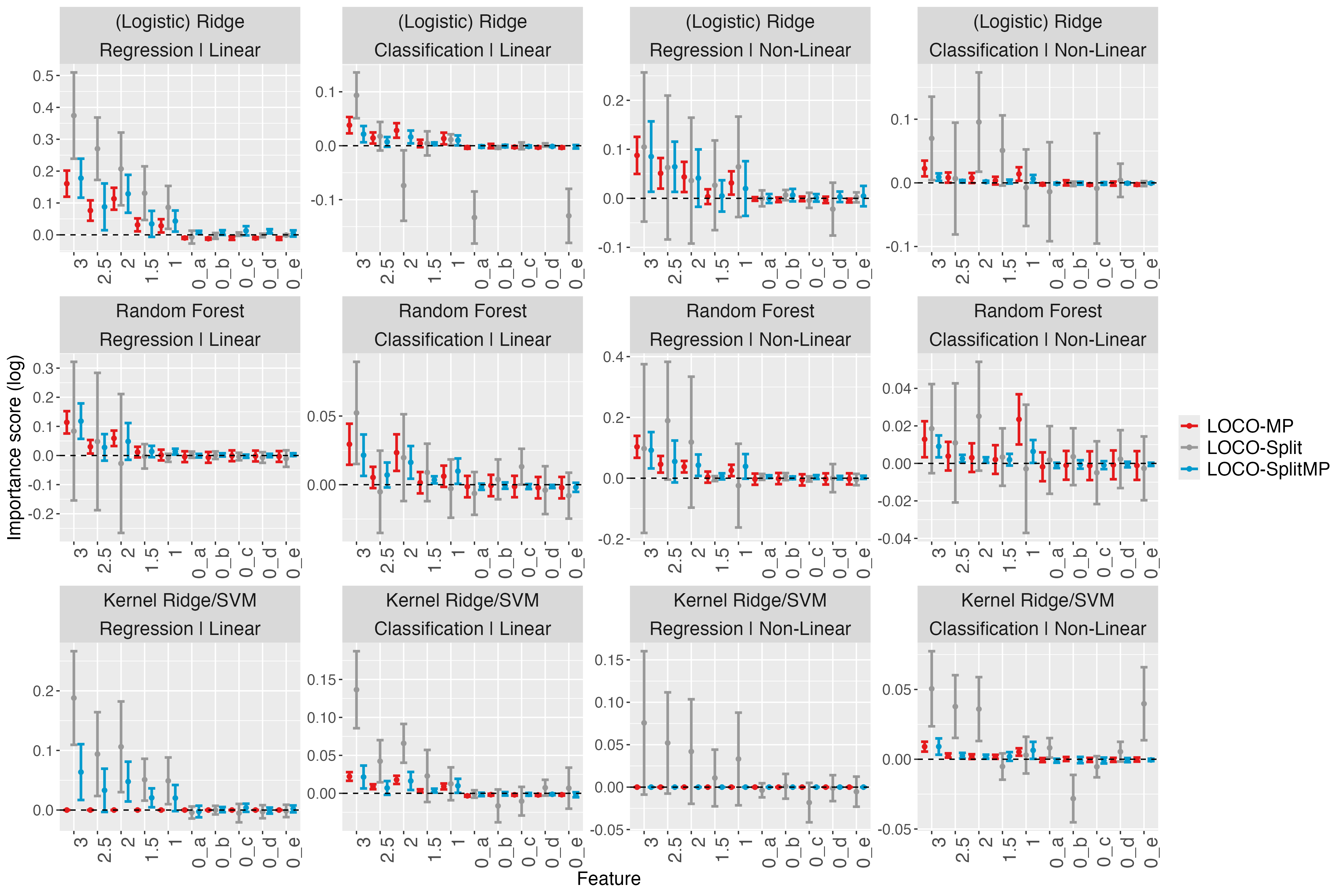}
    \caption{Feature inference on linear and non-linear simulated data, using (logistic) regression, decision tree, and kernel SVM/SVR as the base predictor. Features whose lower bounds of confidence interval greater than zero are statistically significant. The confidence intervals with an upper bound smaller than zero indicate that such a feature would hamper prediction. Overall, LOCO-MP can correctly identify signal features and is among the best in terms of interval efficiency with the smallest widths. The near-zero intervals given by LOCO-MP with kernel ridge regression as the base learner may arise from a poor fit of this predictive model for our nonlinear data-generating model.}
\end{figure}\label{fig:loco_interval1}

\begin{figure}
\centering
    \includegraphics[clip,width=\textwidth]{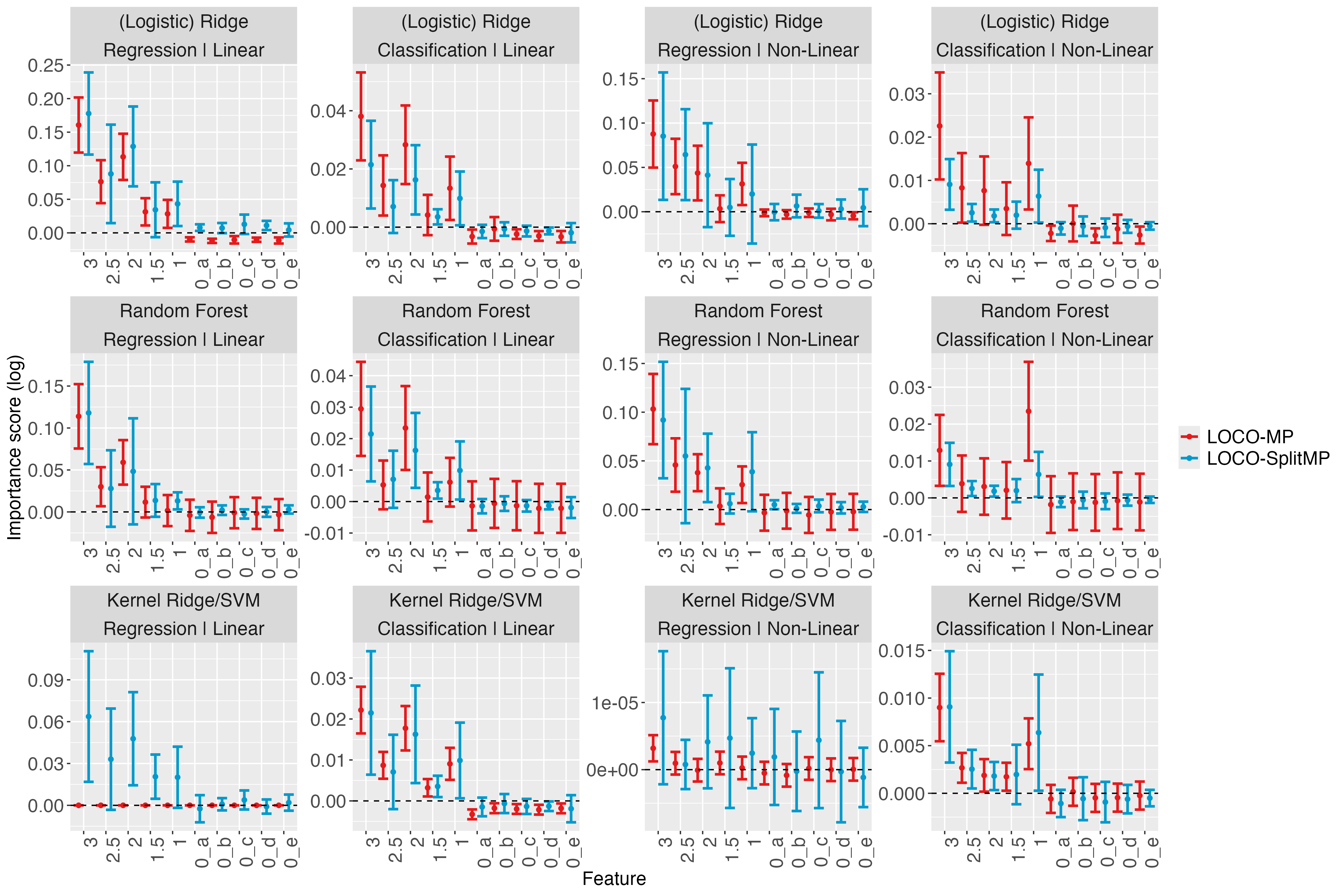}
    \caption{Feature inference comparison between LOCO-MP and LOCO-SplitMP on linear and non-linear simulated data, using (logistic) regression, decision tree, and kernel SVM/SVR as the base predictor. This figure presents the same results as Figure~\ref{fig:loco_interval1} except that it excludes LOCO-Split. Features whose lower bounds of confidence interval greater than zero are statistically significant. The confidence intervals with an upper bound smaller than zero indicate that such a feature would hamper prediction. Overall, LOCO-MP can correctly identify signal features and is among the best in terms of interval efficiency with the smallest widths. The near-zero intervals given by LOCO-MP with kernel ridge regression as the base learner may arise from a poor fit of this predictive model for our nonlinear data-generating model.}
\end{figure}\label{fig:loco_interval2}

\subsection{Case Study on ROSMAP data}

 \begin{table}
\centering
\small 
\resizebox{1\textwidth}{!}{
\begin{tabular}{llllllllllllll}
  \toprule
  Feature & LOCO-MP & LOCO-Split & GCM & VIM & CPI & Floodgate & Feature & LOCO-MP & LOCO-Split & GCM & VIM & CPI & Floodgate\\ 
  \midrule
  TTTY14 & \checkmark &  &  &  &  & \checkmark  & KDM5D &  &  & \checkmark &  &  & \checkmark \\ \hline
  AL162497.1 & \checkmark &  &  &  & \checkmark & \checkmark  & CP &  &  & \checkmark &  &  &  \\ \hline
  RP11-599B13.6 & \checkmark &  &  &  & \checkmark & \checkmark & RN7SL1 &  &  &  &  & \checkmark & \checkmark \\ \hline
  CHI3L2 &  &  & \checkmark &  &  &   &  DDX3Y &  &  &  &  &  & \checkmark \\ \hline

  CSF3 &  &  & \checkmark &  &  &  &   WNT3 &  &  &  &  & \checkmark &  \\ \hline

  CCL2 &  &  & \checkmark &  & \checkmark & \checkmark  &  USP9Y &  &  &  &  &  & \checkmark \\ \hline

  IL1RL1 &  &  & \checkmark &  &  &  &   VGF &  &  &  &  &  & \checkmark \\ \hline

  TXLNG2P &  &  &  &  &  & \checkmark  &  SLC14A1 &  &  & \checkmark &  &  &  \\ \hline

  S100A8 &  &  & \checkmark &  &  &   &  FAM21B &  &  &  &  &  & \checkmark \\ \hline

  SST &  &  &  &  &  & \checkmark &  RPL9 &  &  & \checkmark &  &  & \checkmark \\ \hline
 
  CARTPT &  &  &  &  &  & \checkmark &  C10orf10 &  &  & \checkmark &  &  & \checkmark \\ \hline
 
  FOS &  &  & \checkmark &  &  & \checkmark & C1QB &  &  & \checkmark &  &  &  \\ \hline
  
  C1QA &  &  &  &  & \checkmark &  &  HLA-DQB1 &  &  & \checkmark &  &  &  \\ \hline
 
  HERC2P3 &  &  &  &  &  & \checkmark & SERPINA3 &  &  &  &  & \checkmark & \checkmark \\ \hline
  
  MT1F &  &  & \checkmark &  &  & \checkmark &  EIF1AY &  &  &  &  & \checkmark &  \\ \hline
 
  RP5-857K21.11 &  &  & \checkmark &  & \checkmark &  &   MT1A &  &  & \checkmark &  &  &  \\ \hline

  RNU1-6 &  &  &  &  &  & \checkmark & RNU1-9 &  &  &  &  &  & \checkmark \\ \hline
  
  RNU11 &  &  & \checkmark &  & \checkmark & \checkmark  & ANKRD20A11P &  &  &  &  & \checkmark & \checkmark \\ \hline
 
  PNMA6A &  &  &  &  & \checkmark &  & AC015936.3 &  &  &  &  &  & \checkmark \\ \hline
  
  MTND2P28 &  &  &  &  &  & \checkmark & MTND1P23 &  &  &  &  &  & \checkmark \\ \hline
  
  RP11-742N3.1 &  &  & \checkmark &  &  & &RPL9P8 &  &  & \checkmark &  &  &  \\ \hline
  
  CTA-221G9.10 &  &  &  &  &  & \checkmark &RP11-34P13.13 &  &  & \checkmark &  &  &  \\ \hline
  
  HBB &  &  & \checkmark &  &  & & RP11-318M2.2 &  &  & \checkmark &  & \checkmark & \checkmark \\ \hline

   \bottomrule
\end{tabular}
}\caption{Significant features identified by different methods: LOCO-MP, LOCO-Split, GCM, VIM, CPI, and Floodgate. A checkmark indicates that the feature was identified as significant by the corresponding method.} 
\label{table:impo}
\end{table}

Table\ref{table:impo} includes features identified as significant by at least one method, with a checkmark indicating that the feature is identified as significant by the corresponding method. In particular, for LOCO-MP, we declare that one feature is significant if its one-sided confidence lower bound is positive.

\subsection{Empirical Studies on J+MP Predictive Interval}\label{sec:j+mp}
We evaluate the performance of {J+MP} predictive inference approach on two regression and two classification tasks: a high-dimensional RNA-seq data set from the Religious Orders Study and Memory and Aging Project (ROSMAP) Study \citep{bennett2018religious} with 507 observations and 900 features, the communities and crime data set \citep{asuncion2007uci} with 1993 observations and 99 features, and the spambase classification data \citep{blake1998uci} with 4601 observations and 57 features, as well as one high dimensional classification PANCAN cancer RNA-seq data set \citep{weinstein2013cancer} with 761 observations and 13244 features. We transform every feature to have mean $0$ and variance $1$, and the response of the regression problems is also standardized. The results of {J+MP} predictive intervals are compared to existing conformal inference methods, including split conformal \citep{lei2018distribution}, cross conformal \citep{vovk2015cross}, {Jackknife+} \citep{barber2021predictive} and {J+aB} \citep{kim2020predictive}. For the base prediction algorithms, we select a linear model (logistic regression for classification) with ridge penalty and a non-linear decision tree as base prediction models. In terms of parameter tuning, {J+MP} sets the penalty hyperparameter of the (logistic) ridge model as $0.0001$ and selects the minipatch sizes $(m,n)$ which leads to the lowest mean squared error in regression or highest accuracy in classification via bootstrap validation. For other conformal methods, the penalty hyperparameter is selected via bootstrap validation. The number of folds in cross conformal is set to be $5$. In addition, we evaluate the {J+aB} and {J+MP} methods using a random $K$ drawn from $K\sim Binomial (\Tilde{K}, 1-\frac{n}{N+1})$, where $\Tilde{K} = 100$. And in {J+aB}, we apply sampling without replacement in the bootstrap step, and the optimal sampling size is selected via bootstrap validation. 

Table~\ref{table:j+mp} validates the performance of conformal intervals in terms of coverage (our target is $1-\alpha$), interval width (smaller is better), and computational time, under 100 train/test splits on each benchmark data  with the error rate $\alpha = 0.1$. Empirically, the various forms of {J+MP} and {J+aB} do not differ much from each other, respectively. Moreover, the results show that our {J+MP} and {J+aB} achieve valid and the most consistent coverage, and the other methods fail to provide $0.9$ coverage for the high-dimensional PANCAN classification data set with random forest model.  

Secondly, compared to methods with valid coverage, {J+MP} and {J+aB} have similar performance with relatively small widths. 
However, there is a trade-off between interval efficiency and computational efficiency. It makes sense that the {Jackknife+}-based methods are slower than split conformal and cross conformal, but both {J+MP} and {J+aB} are significantly faster than {Jackknife+}. In addition, {J+MP} further outruns {J+aB} with dramatically great computational savings, when dealing with large high dimensional data set or with the implementation of random forest. Specifically, in the case of PANCAN classification data set, {J+MP} even achieves similar computational efficiency with the {Conformal Split} method, and is dramatically faster than {Jackknife+}, depending on the size of data sets and the base prediction model.

\begin{table}

\centering
% \scalebox{0.72}{
\resizebox{1\textwidth}{!}{
\begin{tabular}{llrrrrrrrrrrrr}
\toprule
\multicolumn{2}{c}{ } & \multicolumn{4}{c}{\bf Coverage} & \multicolumn{4}{c}{\bf Width}& \multicolumn{4}{c}{\bf Time (s)} \\
\cmidrule(l{3pt}r{3pt}){3-6} \cmidrule(l{3pt}r{3pt}){7-10}  \cmidrule(l{3pt}r{3pt}){11-14}

 Base model & Conformal& RNA-seq  & Communities & PANCAN &Spambase & RNA-seq  & Communities & PANCAN  &Spambase & RNA-seq  & Communities & PANCAN  &Spambase\\
\midrule
(Logistic) Ridge 
&J+MP &  0.925 & 0.860 & 0.899 &0.897
&\bf 2.390 & 2.189 & \textbf{0.900} & 1.086
&  0.315 & 0.1097 &  17.166 & 1.472\\

&J+aB &  0.935 &  0.870 &  0.906 &0.897
& 2.447 &   2.017 & 0.906 &0.961
& 0.269 & 0.07 & 137.121  &40.065\\

&J+ &  0.907 &  0.890 &  0.906 & 0.897

&2.527 & \bf 1.908 & 0.906 & 0.955
& 3.818 & 0.349 & > 8 hrs  &14191.856 \\

&Split & 0.916 & 0.890 & 0.809 &0.816
&2.779 & 2.069 &  0.809 &\textbf{0.851}
& 0.00861 & 0.00531 & 235.043  &2.52\\

&Cross & 0.916 & 0.890 & 0.900 &0.906
& 2.598 &  1.919 & \bf 0.900 &0.964

& 0.0872 & 0.0115 & 1994.49 &18.411\\

\hline
Random Forest &J+MP &  0.925 &  0.910 &  0.903  &0.889
& \bf 2.262 & \bf 1.976 &  0.891 &1.025
& 2.659 & 0.118 & 1.833 &1.027\\

&J+aB & 0.934 &  0.900 &  0.890 &0.867
& 2.326 &2.167 & 0.906 & 0.914
& 4.844 &0.564 & 12.625 &38.106\\

&J+ &  0.972 & 0.890 &  0.886 & {0.851}
& 3.267 &  2.792 & 0.484 &0.870
& 90.517 & 3.748 & 2364.72 &3169.744\\

&Split & 0.925 & 0.930 & 0.793  &0.694   

&3.351& 3.774 &\textbf{0.835} &\textbf{0.701}
& 0.144& 0.0118 & 1.832 &0.830\\

&Cross & 0.981 & 0.910 & 0.801 &0.848  

& 3.103 & 2.785 & 0.804 &0.875
& 1.460 &0.096 & 16.795 &4.871\\

\bottomrule

\end{tabular}}
\caption{\textit{Comparative results for predictive intervals constructed via our Minipatch Jacknife+ ({J+MP}) method and existing methods in terms of coverage, interval width, and computational time on two regression (RNA-seq, Communities) and two classifications (PANCAN, Spambase) data sets.} }
\label{table:j+mp}
\end{table}

\bibliographystyle{plainnat}
\bibliography{ref.bib}
\end{document}